%% file: main.tex
\documentclass[10pt,twocolumn,letterpaper]{article}

\usepackage[pagenumbers]{cvpr} 

\usepackage{graphicx}
\usepackage{amsmath}
\usepackage{amssymb}
\usepackage{booktabs}
\usepackage{enumerate}
\usepackage{verse}
\usepackage{tikz}
\usepackage{verbatim}
\usepackage{algorithm}
\usepackage{algpseudocode}
\usepackage{xcolor}
\usepackage{adjustbox}
\usepackage{multirow}
\usepackage{wrapfig}
\usepackage{xcolor}
\usepackage{overpic}
\usepackage{tabularx}
\usepackage{array}

\usepackage{tikz}
\usetikzlibrary{shapes}
\usetikzlibrary{decorations.markings}
\usetikzlibrary{decorations.pathmorphing}
\usetikzlibrary{calc}
\usetikzlibrary{patterns}

\newcommand{\dee}{\mathrm{d}}

\newcommand{\ve}[1]{\mathbf{#1}}
\newcommand{\ves}[1]{\boldsymbol{#1}}
\newcommand{\mrm}[1]{\mathrm{#1}}

\newcommand{\mcal}[1]{\mathcal{#1}}

\newcommand{\Real}{\mathbb{R}}

\input{misc/noise_schedule_commands}

\definecolor{cvprblue}{rgb}{0.21,0.49,0.74}
\usepackage[pagebackref,breaklinks,colorlinks,allcolors=cvprblue]{hyperref}

\def\paperID{14} 
\def\confName{AI4CC}
\def\confYear{2025}

\title{Revisiting Diffusion Autoencoder Training for Image Reconstruction Quality}

\author{Pramook Khungurn\textsuperscript{1},
Sukit Seripanitkarn\textsuperscript{2},
Phonphrm Thawatdamrongkit\textsuperscript{2},
Supasorn Suwajanakorn\textsuperscript{2}\\
\textsuperscript{1}pixiv, Inc. \qquad \textsuperscript{2}VISTEC \\
{\tt\small pong@pixiv.co.jp} \qquad {\tt\small \{sukit.s\_s23,phonphrm.t\_s23,supasorn.s\}@vistec.ac.th}}

\begin{document}
\maketitle
\input{sec/0_abstract}    
\input{sec/1_intro}
\input{sec/2_related_works}

\input{sec/3_background}
\input{sec/4_method}
\input{sec/5_results}
\input{sec/6_conclusion}

{
    \small
    \bibliographystyle{ieeenat_fullname}
    \bibliography{main}
}

\include{sec/X_supplmenetary}


\end{document}

%% file: misc/noise_schedule_commands.tex
\DeclareRobustCommand{\shiftedcosineone}{\begin{tikzpicture} \definecolor{linecolor}{HTML}{2E8B57} \draw [linecolor] (0.0007460406930310566,0.19999847411527294) -- (0.01567143662372795,0.19932706801143163);  \draw [linecolor] (0.01567143662372795,0.19932706801143163) -- (0.030596832554424833,0.19743893194534823);  \draw [linecolor] (0.030596832554424833,0.19743893194534823) -- (0.04552222848512175,0.1943455914551612);  \draw [linecolor] (0.04552222848512175,0.1943455914551612) -- (0.060447624415818635,0.19006592887378893);  \draw [linecolor] (0.060447624415818635,0.19006592887378893) -- (0.07537302034651551,0.18462606806761328);  \draw [linecolor] (0.07537302034651551,0.18462606806761328) -- (0.09029841627721243,0.1780592149714797);  \draw [linecolor] (0.09029841627721243,0.1780592149714797) -- (0.10522381220790931,0.1704054548934169);  \draw [linecolor] (0.10522381220790931,0.1704054548934169) -- (0.12014920813860619,0.16171150782636767);  \draw [linecolor] (0.12014920813860619,0.16171150782636767) -- (0.1350746040693031,0.15203044326055304);  \draw [linecolor] (0.1350746040693031,0.15203044326055304) -- (0.15,0.14142135623730953);  \draw [linecolor] (0.15,0.14142135623730953) -- (0.16492539593069688,0.12994900662182754);  \draw [linecolor] (0.16492539593069688,0.12994900662182754) -- (0.17985079186139377,0.11768342379673796);  \draw [linecolor] (0.17985079186139377,0.11768342379673796) -- (0.1947761877920907,0.10469947918957219);  \draw [linecolor] (0.1947761877920907,0.10469947918957219) -- (0.20970158372278755,0.09107642924346776);  \draw [linecolor] (0.20970158372278755,0.09107642924346776) -- (0.22462697965348447,0.07689743162091327);  \draw [linecolor] (0.22462697965348447,0.07689743162091327) -- (0.23955237558418135,0.062249037593715695);  \draw [linecolor] (0.23955237558418135,0.062249037593715695) -- (0.2544777715148782,0.04722066371773674);  \draw [linecolor] (0.2544777715148782,0.04722066371773674) -- (0.26940316744557513,0.031904046017396596);  \draw [linecolor] (0.26940316744557513,0.031904046017396596) -- (0.28432856337627205,0.016392680011703877);  \draw [linecolor] (0.28432856337627205,0.016392680011703877) -- (0.2992539593069689,0.0007812500000000177); \end{tikzpicture}}

\DeclareRobustCommand{\shiftedcosinehalf}{\begin{tikzpicture} \definecolor{linecolor}{HTML}{B8860B} \draw [linecolor] (0.00037302176948789,0.19999847411527294) -- (0.01527976775004921,0.19747728207848847);  \draw [linecolor] (0.01527976775004921,0.19747728207848847) -- (0.03018651373061053,0.19055267089005923);  \draw [linecolor] (0.03018651373061053,0.19055267089005923) -- (0.04509325971117182,0.18022082343223925);  \draw [linecolor] (0.04509325971117182,0.18022082343223925) -- (0.06000000569173314,0.16770102444029555);  \draw [linecolor] (0.06000000569173314,0.16770102444029555) -- (0.07490675167229445,0.15410178755543064);  \draw [linecolor] (0.07490675167229445,0.15410178755543064) -- (0.08981349765285575,0.1402530283694359);  \draw [linecolor] (0.08981349765285575,0.1402530283694359) -- (0.10472024363341706,0.12668771172880097);  \draw [linecolor] (0.10472024363341706,0.12668771172880097) -- (0.11962698961397834,0.11369929425971592);  \draw [linecolor] (0.11962698961397834,0.11369929425971592) -- (0.13453373559453966,0.10141579196833533);  \draw [linecolor] (0.13453373559453966,0.10141579196833533) -- (0.14944048157510098,0.08986246563147417);  \draw [linecolor] (0.14944048157510098,0.08986246563147417) -- (0.1643472275556623,0.07900625796289329);  \draw [linecolor] (0.1643472275556623,0.07900625796289329) -- (0.1792539735362236,0.06878423894966054);  \draw [linecolor] (0.1792539735362236,0.06878423894966054) -- (0.19416071951678487,0.05912054948059743);  \draw [linecolor] (0.19416071951678487,0.05912054948059743) -- (0.2090674654973462,0.04993585481272025);  \draw [linecolor] (0.2090674654973462,0.04993585481272025) -- (0.22397421147790753,0.04115220141042675);  \draw [linecolor] (0.22397421147790753,0.04115220141042675) -- (0.23888095745846882,0.03269516556861958);  \draw [linecolor] (0.23888095745846882,0.03269516556861958) -- (0.2537877034390301,0.024494457992823362);  \draw [linecolor] (0.2537877034390301,0.024494457992823362) -- (0.2686944494195914,0.016483675401020442);  \draw [linecolor] (0.2686944494195914,0.016483675401020442) -- (0.2836011954001527,0.008599597197952258);  \draw [linecolor] (0.2836011954001527,0.008599597197952258) -- (0.29850794138071407,0.000781249999999999); \end{tikzpicture}}

\DeclareRobustCommand{\shiftedcosinefourth}{\begin{tikzpicture} \definecolor{linecolor}{HTML}{FF8C00} \draw [linecolor] (0.00037302176948789,0.19999847411527294) -- (0.01527976775004921,0.19747728207848847);  \draw [linecolor] (0.01527976775004921,0.19747728207848847) -- (0.03018651373061053,0.19055267089005923);  \draw [linecolor] (0.03018651373061053,0.19055267089005923) -- (0.04509325971117182,0.18022082343223925);  \draw [linecolor] (0.04509325971117182,0.18022082343223925) -- (0.06000000569173314,0.16770102444029555);  \draw [linecolor] (0.06000000569173314,0.16770102444029555) -- (0.07490675167229445,0.15410178755543064);  \draw [linecolor] (0.07490675167229445,0.15410178755543064) -- (0.08981349765285575,0.1402530283694359);  \draw [linecolor] (0.08981349765285575,0.1402530283694359) -- (0.10472024363341706,0.12668771172880097);  \draw [linecolor] (0.10472024363341706,0.12668771172880097) -- (0.11962698961397834,0.11369929425971592);  \draw [linecolor] (0.11962698961397834,0.11369929425971592) -- (0.13453373559453966,0.10141579196833533);  \draw [linecolor] (0.13453373559453966,0.10141579196833533) -- (0.14944048157510098,0.08986246563147417);  \draw [linecolor] (0.14944048157510098,0.08986246563147417) -- (0.1643472275556623,0.07900625796289329);  \draw [linecolor] (0.1643472275556623,0.07900625796289329) -- (0.1792539735362236,0.06878423894966054);  \draw [linecolor] (0.1792539735362236,0.06878423894966054) -- (0.19416071951678487,0.05912054948059743);  \draw [linecolor] (0.19416071951678487,0.05912054948059743) -- (0.2090674654973462,0.04993585481272025);  \draw [linecolor] (0.2090674654973462,0.04993585481272025) -- (0.22397421147790753,0.04115220141042675);  \draw [linecolor] (0.22397421147790753,0.04115220141042675) -- (0.23888095745846882,0.03269516556861958);  \draw [linecolor] (0.23888095745846882,0.03269516556861958) -- (0.2537877034390301,0.024494457992823362);  \draw [linecolor] (0.2537877034390301,0.024494457992823362) -- (0.2686944494195914,0.016483675401020442);  \draw [linecolor] (0.2686944494195914,0.016483675401020442) -- (0.2836011954001527,0.008599597197952258);  \draw [linecolor] (0.2836011954001527,0.008599597197952258) -- (0.29850794138071407,0.000781249999999999); \end{tikzpicture}}

\DeclareRobustCommand{\shiftedcosineeighth}{\begin{tikzpicture} \definecolor{linecolor}{HTML}{FF0000} \draw [linecolor] (0.00037302176948789,0.19999847411527294) -- (0.01527976775004921,0.19747728207848847);  \draw [linecolor] (0.01527976775004921,0.19747728207848847) -- (0.03018651373061053,0.19055267089005923);  \draw [linecolor] (0.03018651373061053,0.19055267089005923) -- (0.04509325971117182,0.18022082343223925);  \draw [linecolor] (0.04509325971117182,0.18022082343223925) -- (0.06000000569173314,0.16770102444029555);  \draw [linecolor] (0.06000000569173314,0.16770102444029555) -- (0.07490675167229445,0.15410178755543064);  \draw [linecolor] (0.07490675167229445,0.15410178755543064) -- (0.08981349765285575,0.1402530283694359);  \draw [linecolor] (0.08981349765285575,0.1402530283694359) -- (0.10472024363341706,0.12668771172880097);  \draw [linecolor] (0.10472024363341706,0.12668771172880097) -- (0.11962698961397834,0.11369929425971592);  \draw [linecolor] (0.11962698961397834,0.11369929425971592) -- (0.13453373559453966,0.10141579196833533);  \draw [linecolor] (0.13453373559453966,0.10141579196833533) -- (0.14944048157510098,0.08986246563147417);  \draw [linecolor] (0.14944048157510098,0.08986246563147417) -- (0.1643472275556623,0.07900625796289329);  \draw [linecolor] (0.1643472275556623,0.07900625796289329) -- (0.1792539735362236,0.06878423894966054);  \draw [linecolor] (0.1792539735362236,0.06878423894966054) -- (0.19416071951678487,0.05912054948059743);  \draw [linecolor] (0.19416071951678487,0.05912054948059743) -- (0.2090674654973462,0.04993585481272025);  \draw [linecolor] (0.2090674654973462,0.04993585481272025) -- (0.22397421147790753,0.04115220141042675);  \draw [linecolor] (0.22397421147790753,0.04115220141042675) -- (0.23888095745846882,0.03269516556861958);  \draw [linecolor] (0.23888095745846882,0.03269516556861958) -- (0.2537877034390301,0.024494457992823362);  \draw [linecolor] (0.2537877034390301,0.024494457992823362) -- (0.2686944494195914,0.016483675401020442);  \draw [linecolor] (0.2686944494195914,0.016483675401020442) -- (0.2836011954001527,0.008599597197952258);  \draw [linecolor] (0.2836011954001527,0.008599597197952258) -- (0.29850794138071407,0.000781249999999999); \end{tikzpicture}}

\DeclareRobustCommand{\shiftedcosinetwo}{\begin{tikzpicture} \definecolor{linecolor}{HTML}{008000} \draw [linecolor] (0.0014920586192859638,0.19999847411527294) -- (0.01639880459984726,0.1998150317869829);  \draw [linecolor] (0.01639880459984726,0.1998150317869829) -- (0.03130555058040858,0.19931956362904724);  \draw [linecolor] (0.03130555058040858,0.19931956362904724) -- (0.04621229656096987,0.19849438664012092);  \draw [linecolor] (0.04621229656096987,0.19849438664012092) -- (0.061119042541531186,0.19730946796451648);  \draw [linecolor] (0.061119042541531186,0.19730946796451648) -- (0.0760257885220925,0.195720454523986);  \draw [linecolor] (0.0760257885220925,0.195720454523986) -- (0.09093253450265379,0.1936657181953557);  \draw [linecolor] (0.09093253450265379,0.1936657181953557) -- (0.10583928048321511,0.19106219047501846);  \draw [linecolor] (0.10583928048321511,0.19106219047501846) -- (0.1207460264637764,0.18779970306716676);  \draw [linecolor] (0.1207460264637764,0.18779970306716676) -- (0.1356527724443377,0.18373353314705718);  \draw [linecolor] (0.1356527724443377,0.18373353314705718) -- (0.15055951842489904,0.1786749486375528);  \draw [linecolor] (0.15055951842489904,0.1786749486375528) -- (0.16546626440546033,0.1723799209288464);  \draw [linecolor] (0.16546626440546033,0.1723799209288464) -- (0.18037301038602163,0.16453714013815401);  \draw [linecolor] (0.18037301038602163,0.16453714013815401) -- (0.19527975636658296,0.15475859813567783);  \draw [linecolor] (0.19527975636658296,0.15475859813567783) -- (0.21018650234714426,0.14258011093137155);  \draw [linecolor] (0.21018650234714426,0.14258011093137155) -- (0.22509324832770555,0.1274858387124661);  \draw [linecolor] (0.22509324832770555,0.1274858387124661) -- (0.23999999430826685,0.1089787428890396);  \draw [linecolor] (0.23999999430826685,0.1089787428890396) -- (0.25490674028882815,0.08672055581813143);  \draw [linecolor] (0.25490674028882815,0.08672055581813143) -- (0.2698134862693895,0.0607427330358519);  \draw [linecolor] (0.2698134862693895,0.0607427330358519) -- (0.2847202322499508,0.03166580273565004);  \draw [linecolor] (0.2847202322499508,0.03166580273565004) -- (0.2996269782305121,0.0007812500000000302); \end{tikzpicture}}

\DeclareRobustCommand{\shiftedcosinefour}{\begin{tikzpicture} \definecolor{linecolor}{HTML}{0000FF} \draw [linecolor] (0.0029839351293708643,0.19999847411527294) -- (0.01782541281977147,0.1999452599225585);  \draw [linecolor] (0.01782541281977147,0.1999452599225585) -- (0.032666890510172075,0.19981378497243107);  \draw [linecolor] (0.032666890510172075,0.19981378497243107) -- (0.04750836820057268,0.19959794524942132);  \draw [linecolor] (0.04750836820057268,0.19959794524942132) -- (0.06234984589097328,0.19928735926110838);  \draw [linecolor] (0.06234984589097328,0.19928735926110838) -- (0.07719132358137389,0.19886630672692535);  \draw [linecolor] (0.07719132358137389,0.19886630672692535) -- (0.0920328012717745,0.19831201674704402);  \draw [linecolor] (0.0920328012717745,0.19831201674704402) -- (0.1068742789621751,0.19759198368765524);  \draw [linecolor] (0.1068742789621751,0.19759198368765524) -- (0.12171575665257571,0.1966597612057527);  \draw [linecolor] (0.12171575665257571,0.1966597612057527) -- (0.1365572343429763,0.19544828649066162);  \draw [linecolor] (0.1365572343429763,0.19544828649066162) -- (0.15139871203337693,0.19385906745785944);  \draw [linecolor] (0.15139871203337693,0.19385906745785944) -- (0.16624018972377752,0.19174423180816383);  \draw [linecolor] (0.16624018972377752,0.19174423180816383) -- (0.18108166741417814,0.1888759136749127);  \draw [linecolor] (0.18108166741417814,0.1888759136749127) -- (0.19592314510457873,0.18489264581116902);  \draw [linecolor] (0.19592314510457873,0.18489264581116902) -- (0.21076462279497934,0.17920347747298349);  \draw [linecolor] (0.21076462279497934,0.17920347747298349) -- (0.22560610048537993,0.17081564595755927);  \draw [linecolor] (0.22560610048537993,0.17081564595755927) -- (0.24044757817578055,0.15803702129136477);  \draw [linecolor] (0.24044757817578055,0.15803702129136477) -- (0.25528905586618117,0.13804736066688786);  \draw [linecolor] (0.25528905586618117,0.13804736066688786) -- (0.2701305335565818,0.10669659571232977);  \draw [linecolor] (0.2701305335565818,0.10669659571232977) -- (0.28497201124698235,0.06015813940413966);  \draw [linecolor] (0.28497201124698235,0.06015813940413966) -- (0.29981348893738297,0.0007812499999999657); \end{tikzpicture}}

\DeclareRobustCommand{\shiftedcosineeight}{\begin{tikzpicture} \definecolor{linecolor}{HTML}{8A2BE2} \draw [linecolor] (0.005966414184844298,0.19999847411527294) -- (0.020663430697924968,0.1999815685051226);  \draw [linecolor] (0.020663430697924968,0.1999815685051226) -- (0.035360447211005626,0.199945212770868);  \draw [linecolor] (0.035360447211005626,0.199945212770868) -- (0.05005746372408626,0.19988764208915863);  \draw [linecolor] (0.05005746372408626,0.19988764208915863) -- (0.06475448023716691,0.1998059423633662);  \draw [linecolor] (0.06475448023716691,0.1998059423633662) -- (0.07945149675024758,0.19969572702071695);  \draw [linecolor] (0.07945149675024758,0.19969572702071695) -- (0.09414851326332821,0.1995506121338342);  \draw [linecolor] (0.09414851326332821,0.1995506121338342) -- (0.10884552977640888,0.19936137669132176);  \draw [linecolor] (0.10884552977640888,0.19936137669132176) -- (0.1235425462894895,0.1991146061604564);  \draw [linecolor] (0.1235425462894895,0.1991146061604564) -- (0.13823956280257016,0.19879045009996663);  \draw [linecolor] (0.13823956280257016,0.19879045009996663) -- (0.15293657931565083,0.19835879283217078);  \draw [linecolor] (0.15293657931565083,0.19835879283217078) -- (0.1676335958287315,0.19777244374924338);  \draw [linecolor] (0.1676335958287315,0.19777244374924338) -- (0.1823306123418121,0.19695442231135818);  \draw [linecolor] (0.1823306123418121,0.19695442231135818) -- (0.19702762885489275,0.1957727926786017);  \draw [linecolor] (0.19702762885489275,0.1957727926786017) -- (0.21172464536797342,0.19398726243136669);  \draw [linecolor] (0.21172464536797342,0.19398726243136669) -- (0.22642166188105406,0.19112599033153432);  \draw [linecolor] (0.22642166188105406,0.19112599033153432) -- (0.24111867839413473,0.18617173377424667);  \draw [linecolor] (0.24111867839413473,0.18617173377424667) -- (0.25581569490721534,0.1766680407560809);  \draw [linecolor] (0.25581569490721534,0.1766680407560809) -- (0.27051271142029604,0.15593250585772148);  \draw [linecolor] (0.27051271142029604,0.15593250585772148) -- (0.2852097279333767,0.10548303325827174);  \draw [linecolor] (0.2852097279333767,0.10548303325827174) -- (0.2999067444464573,0.0007812499999999699); \end{tikzpicture}}

\DeclareRobustCommand{\shiftedcosinesixteen}{\begin{tikzpicture} \definecolor{linecolor}{HTML}{800080} \draw [linecolor] (0.011921205299194085,0.19999847411527294) -- (0.02632281364525685,0.1999924851238102);  \draw [linecolor] (0.02632281364525685,0.1999924851238102) -- (0.04072442199131963,0.19998168894621762);  \draw [linecolor] (0.04072442199131963,0.19998168894621762) -- (0.05512603033738237,0.19996556816174982);  \draw [linecolor] (0.05512603033738237,0.19996556816174982) -- (0.06952763868344515,0.19994331196224258);  \draw [linecolor] (0.06952763868344515,0.19994331196224258) -- (0.08392924702950792,0.19991372571305352);  \draw [linecolor] (0.08392924702950792,0.19991372571305352) -- (0.09833085537557067,0.19987508608609503);  \draw [linecolor] (0.09833085537557067,0.19987508608609503) -- (0.11273246372163344,0.1998249096832922);  \draw [linecolor] (0.11273246372163344,0.1998249096832922) -- (0.1271340720676962,0.1997595775645105);  \draw [linecolor] (0.1271340720676962,0.1997595775645105) -- (0.14153568041375897,0.19967370900059866);  \draw [linecolor] (0.14153568041375897,0.19967370900059866) -- (0.15593728875982174,0.1995590782912357);  \draw [linecolor] (0.15593728875982174,0.1995590782912357) -- (0.1703388971058845,0.19940265510227115);  \draw [linecolor] (0.1703388971058845,0.19940265510227115) -- (0.18474050545194726,0.19918285990375767);  \draw [linecolor] (0.18474050545194726,0.19918285990375767) -- (0.19914211379801,0.1988619148325343);  \draw [linecolor] (0.19914211379801,0.1988619148325343) -- (0.21354372214407277,0.19836886954210425);  \draw [linecolor] (0.21354372214407277,0.19836886954210425) -- (0.22794533049013554,0.1975577544728389);  \draw [linecolor] (0.22794533049013554,0.1975577544728389) -- (0.2423469388361983,0.19608913572648848);  \draw [linecolor] (0.2423469388361983,0.19608913572648848) -- (0.2567485471822611,0.1930250758881021);  \draw [linecolor] (0.2567485471822611,0.1930250758881021) -- (0.27115015552832383,0.18501151778944402);  \draw [linecolor] (0.27115015552832383,0.18501151778944402) -- (0.2855517638743866,0.15430586592129042);  \draw [linecolor] (0.2855517638743866,0.15430586592129042) -- (0.2999533722204494,0.0007812500000000767); \end{tikzpicture}}

\DeclareRobustCommand{\linearbeta}{\begin{tikzpicture} \definecolor{linecolor}{HTML}{000000} \draw [linecolor] (0.3,0.001270563617514006) -- (0.28409999999999996,0.002140056852008979);  \draw [linecolor] (0.28409999999999996,0.002140056852008979) -- (0.2682,0.0035032325378028667);  \draw [linecolor] (0.2682,0.0035032325378028667) -- (0.2526,0.005526497786623438);  \draw [linecolor] (0.2526,0.005526497786623438) -- (0.2367,0.008550591273109203);  \draw [linecolor] (0.2367,0.008550591273109203) -- (0.2208,0.012858688040131664);  \draw [linecolor] (0.2208,0.012858688040131664) -- (0.20520000000000002,0.01866672446437551);  \draw [linecolor] (0.20520000000000002,0.01866672446437551) -- (0.1893,0.026536848019192497);  \draw [linecolor] (0.1893,0.026536848019192497) -- (0.17339999999999997,0.036671128737504645);  \draw [linecolor] (0.17339999999999997,0.036671128737504645) -- (0.1578,0.0490005848520574);  \draw [linecolor] (0.1578,0.0490005848520574) -- (0.1419,0.06402279487110181);  \draw [linecolor] (0.1419,0.06402279487110181) -- (0.1263,0.0809756689001553);  \draw [linecolor] (0.1263,0.0809756689001553) -- (0.1104,0.10004553459093773);  \draw [linecolor] (0.1104,0.10004553459093773) -- (0.0945,0.12017096494977082);  \draw [linecolor] (0.0945,0.12017096494977082) -- (0.0789,0.13996355850394876);  \draw [linecolor] (0.0789,0.13996355850394876) -- (0.063,0.15900233932866242);  \draw [linecolor] (0.063,0.15900233932866242) -- (0.047099999999999996,0.17562624645753416);  \draw [linecolor] (0.047099999999999996,0.17562624645753416) -- (0.0315,0.1884141570852727);  \draw [linecolor] (0.0315,0.1884141570852727) -- (0.0156,0.1968622979938699);  \draw [linecolor] (0.0156,0.1968622979938699) -- (0.0,0.2); \end{tikzpicture}}

\DeclareRobustCommand{\shiftedcosinesixteensmall}{\begin{tikzpicture} \definecolor{linecolor}{HTML}{800080} \draw [linecolor] (0.008940903974395564,0.1499988555864547) -- (0.019742110233942636,0.14999436384285764);  \draw [linecolor] (0.019742110233942636,0.14999436384285764) -- (0.03054331649348972,0.1499862667096632);  \draw [linecolor] (0.03054331649348972,0.1499862667096632) -- (0.04134452275303678,0.14997417612131236);  \draw [linecolor] (0.04134452275303678,0.14997417612131236) -- (0.05214572901258386,0.1499574839716819);  \draw [linecolor] (0.05214572901258386,0.1499574839716819) -- (0.06294693527213095,0.14993529428479013);  \draw [linecolor] (0.06294693527213095,0.14993529428479013) -- (0.07374814153167801,0.14990631456457126);  \draw [linecolor] (0.07374814153167801,0.14990631456457126) -- (0.08454934779122508,0.14986868226246913);  \draw [linecolor] (0.08454934779122508,0.14986868226246913) -- (0.09535055405077214,0.14981968317338284);  \draw [linecolor] (0.09535055405077214,0.14981968317338284) -- (0.10615176031031923,0.14975528175044897);  \draw [linecolor] (0.10615176031031923,0.14975528175044897) -- (0.11695296656986631,0.14966930871842676);  \draw [linecolor] (0.11695296656986631,0.14966930871842676) -- (0.1277541728294134,0.14955199132670335);  \draw [linecolor] (0.1277541728294134,0.14955199132670335) -- (0.13855537908896046,0.14938714492781824);  \draw [linecolor] (0.13855537908896046,0.14938714492781824) -- (0.14935658534850751,0.1491464361244007);  \draw [linecolor] (0.14935658534850751,0.1491464361244007) -- (0.1601577916080546,0.14877665215657818);  \draw [linecolor] (0.1601577916080546,0.14877665215657818) -- (0.1709589978676017,0.14816831585462917);  \draw [linecolor] (0.1709589978676017,0.14816831585462917) -- (0.18176020412714874,0.14706685179486634);  \draw [linecolor] (0.18176020412714874,0.14706685179486634) -- (0.19256141038669583,0.14476880691607658);  \draw [linecolor] (0.19256141038669583,0.14476880691607658) -- (0.20336261664624292,0.138758638342083);  \draw [linecolor] (0.20336261664624292,0.138758638342083) -- (0.21416382290578997,0.1157293994409678);  \draw [linecolor] (0.21416382290578997,0.1157293994409678) -- (0.22496502916533703,0.0005859375000000574); \end{tikzpicture}}

\DeclareRobustCommand{\shiftedcosineeightsmall}{\begin{tikzpicture} \definecolor{linecolor}{HTML}{8A2BE2} \draw [linecolor] (0.004474810638633224,0.1499988555864547) -- (0.015497573023443725,0.14998617637884196);  \draw [linecolor] (0.015497573023443725,0.14998617637884196) -- (0.026520335408254225,0.14995890957815097);  \draw [linecolor] (0.026520335408254225,0.14995890957815097) -- (0.037543097793064695,0.14991573156686896);  \draw [linecolor] (0.037543097793064695,0.14991573156686896) -- (0.048565860177875196,0.14985445677252465);  \draw [linecolor] (0.048565860177875196,0.14985445677252465) -- (0.05958862256268569,0.1497717952655377);  \draw [linecolor] (0.05958862256268569,0.1497717952655377) -- (0.07061138494749616,0.14966295910037564);  \draw [linecolor] (0.07061138494749616,0.14966295910037564) -- (0.08163414733230666,0.1495210325184913);  \draw [linecolor] (0.08163414733230666,0.1495210325184913) -- (0.09265690971711714,0.14933595462034227);  \draw [linecolor] (0.09265690971711714,0.14933595462034227) -- (0.10367967210192763,0.14909283757497496);  \draw [linecolor] (0.10367967210192763,0.14909283757497496) -- (0.11470243448673813,0.14876909462412807);  \draw [linecolor] (0.11470243448673813,0.14876909462412807) -- (0.12572519687154862,0.14832933281193253);  \draw [linecolor] (0.12572519687154862,0.14832933281193253) -- (0.1367479592563591,0.1477158167335186);  \draw [linecolor] (0.1367479592563591,0.1477158167335186) -- (0.14777072164116958,0.14682959450895125);  \draw [linecolor] (0.14777072164116958,0.14682959450895125) -- (0.15879348402598006,0.145490446823525);  \draw [linecolor] (0.15879348402598006,0.145490446823525) -- (0.16981624641079057,0.14334449274865071);  \draw [linecolor] (0.16981624641079057,0.14334449274865071) -- (0.18083900879560105,0.13962880033068498);  \draw [linecolor] (0.18083900879560105,0.13962880033068498) -- (0.19186177118041153,0.13250103056706067);  \draw [linecolor] (0.19186177118041153,0.13250103056706067) -- (0.20288453356522204,0.1169493793932911);  \draw [linecolor] (0.20288453356522204,0.1169493793932911) -- (0.2139072959500325,0.07911227494370379);  \draw [linecolor] (0.2139072959500325,0.07911227494370379) -- (0.224930058334843,0.0005859374999999773); \end{tikzpicture}}

\DeclareRobustCommand{\shiftedcosinefoursmall}{\begin{tikzpicture} \definecolor{linecolor}{HTML}{0000FF} \draw [linecolor] (0.002237951347028148,0.1499988555864547) -- (0.013369059614828602,0.14995894494191886);  \draw [linecolor] (0.013369059614828602,0.14995894494191886) -- (0.024500167882629056,0.1498603387293233);  \draw [linecolor] (0.024500167882629056,0.1498603387293233) -- (0.03563127615042951,0.149698458937066);  \draw [linecolor] (0.03563127615042951,0.149698458937066) -- (0.04676238441822997,0.14946551944583128);  \draw [linecolor] (0.04676238441822997,0.14946551944583128) -- (0.057893492686030425,0.149149730045194);  \draw [linecolor] (0.057893492686030425,0.149149730045194) -- (0.06902460095383088,0.14873401256028299);  \draw [linecolor] (0.06902460095383088,0.14873401256028299) -- (0.08015570922163133,0.14819398776574141);  \draw [linecolor] (0.08015570922163133,0.14819398776574141) -- (0.09128681748943179,0.1474948209043145);  \draw [linecolor] (0.09128681748943179,0.1474948209043145) -- (0.10241792575723224,0.1465862148679962);  \draw [linecolor] (0.10241792575723224,0.1465862148679962) -- (0.11354903402503269,0.14539430059339456);  \draw [linecolor] (0.11354903402503269,0.14539430059339456) -- (0.12468014229283315,0.14380817385612285);  \draw [linecolor] (0.12468014229283315,0.14380817385612285) -- (0.13581125056063362,0.1416569352561845);  \draw [linecolor] (0.13581125056063362,0.1416569352561845) -- (0.14694235882843407,0.13866948435837675);  \draw [linecolor] (0.14694235882843407,0.13866948435837675) -- (0.15807346709623452,0.1344026081047376);  \draw [linecolor] (0.15807346709623452,0.1344026081047376) -- (0.16920457536403496,0.12811173446816945);  \draw [linecolor] (0.16920457536403496,0.12811173446816945) -- (0.18033568363183541,0.11852776596852357);  \draw [linecolor] (0.18033568363183541,0.11852776596852357) -- (0.1914667918996359,0.10353552050016589);  \draw [linecolor] (0.1914667918996359,0.10353552050016589) -- (0.20259790016743634,0.08002244678424732);  \draw [linecolor] (0.20259790016743634,0.08002244678424732) -- (0.2137290084352368,0.04511860455310474);  \draw [linecolor] (0.2137290084352368,0.04511860455310474) -- (0.22486011670303724,0.0005859374999999742); \end{tikzpicture}}

\DeclareRobustCommand{\shiftedcosinetwosmall}{\begin{tikzpicture} \definecolor{linecolor}{HTML}{008000} \draw [linecolor] (0.0011190439644644728,0.1499988555864547) -- (0.012299103449885445,0.14986127384023715);  \draw [linecolor] (0.012299103449885445,0.14986127384023715) -- (0.023479162935306438,0.1494896727217854);  \draw [linecolor] (0.023479162935306438,0.1494896727217854) -- (0.0346592224207274,0.14887078998009068);  \draw [linecolor] (0.0346592224207274,0.14887078998009068) -- (0.045839281906148395,0.14798210097338735);  \draw [linecolor] (0.045839281906148395,0.14798210097338735) -- (0.05701934139156938,0.1467903408929895);  \draw [linecolor] (0.05701934139156938,0.1467903408929895) -- (0.06819940087699035,0.14524928864651676);  \draw [linecolor] (0.06819940087699035,0.14524928864651676) -- (0.07937946036241134,0.14329664285626384);  \draw [linecolor] (0.07937946036241134,0.14329664285626384) -- (0.09055951984783231,0.14084977730037507);  \draw [linecolor] (0.09055951984783231,0.14084977730037507) -- (0.1017395793332533,0.13780014986029288);  \draw [linecolor] (0.1017395793332533,0.13780014986029288) -- (0.11291963881867428,0.13400621147816458);  \draw [linecolor] (0.11291963881867428,0.13400621147816458) -- (0.12409969830409524,0.1292849406966348);  \draw [linecolor] (0.12409969830409524,0.1292849406966348) -- (0.13527975778951623,0.12340285510361551);  \draw [linecolor] (0.13527975778951623,0.12340285510361551) -- (0.14645981727493723,0.11606894860175834);  \draw [linecolor] (0.14645981727493723,0.11606894860175834) -- (0.1576398767603582,0.10693508319852864);  \draw [linecolor] (0.1576398767603582,0.10693508319852864) -- (0.16881993624577918,0.09561437903434958);  \draw [linecolor] (0.16881993624577918,0.09561437903434958) -- (0.17999999573120015,0.08173405716677969);  \draw [linecolor] (0.17999999573120015,0.08173405716677969) -- (0.19118005521662115,0.06504041686359856);  \draw [linecolor] (0.19118005521662115,0.06504041686359856) -- (0.20236011470204213,0.04555704977688892);  \draw [linecolor] (0.20236011470204213,0.04555704977688892) -- (0.2135401741874631,0.02374935205173753);  \draw [linecolor] (0.2135401741874631,0.02374935205173753) -- (0.22472023367288407,0.0005859375000000226); \end{tikzpicture}}

\DeclareRobustCommand{\shiftedcosineonesmall}{\begin{tikzpicture} \definecolor{linecolor}{HTML}{2E8B57} \draw [linecolor] (0.0005595305197732925,0.1499988555864547) -- (0.011753577467795965,0.1494953010085737);  \draw [linecolor] (0.011753577467795965,0.1494953010085737) -- (0.022947624415818626,0.14807919895901114);  \draw [linecolor] (0.022947624415818626,0.14807919895901114) -- (0.03414167136384132,0.14575919359137088);  \draw [linecolor] (0.03414167136384132,0.14575919359137088) -- (0.045335718311863976,0.14254944665534167);  \draw [linecolor] (0.045335718311863976,0.14254944665534167) -- (0.05652976525988664,0.13846955105070996);  \draw [linecolor] (0.05652976525988664,0.13846955105070996) -- (0.06772381220790932,0.13354441122860974);  \draw [linecolor] (0.06772381220790932,0.13354441122860974) -- (0.078917859155932,0.12780409117006267);  \draw [linecolor] (0.078917859155932,0.12780409117006267) -- (0.09011190610395466,0.12128363086977575);  \draw [linecolor] (0.09011190610395466,0.12128363086977575) -- (0.10130595305197734,0.11402283244541475);  \draw [linecolor] (0.10130595305197734,0.11402283244541475) -- (0.1125,0.10606601717798213);  \draw [linecolor] (0.1125,0.10606601717798213) -- (0.12369404694802266,0.09746175496637065);  \draw [linecolor] (0.12369404694802266,0.09746175496637065) -- (0.13488809389604534,0.08826256784755346);  \draw [linecolor] (0.13488809389604534,0.08826256784755346) -- (0.146082140844068,0.07852460939217913);  \draw [linecolor] (0.146082140844068,0.07852460939217913) -- (0.15727618779209068,0.0683073219326008);  \draw [linecolor] (0.15727618779209068,0.0683073219326008) -- (0.16847023474011336,0.05767307371568494);  \draw [linecolor] (0.16847023474011336,0.05767307371568494) -- (0.17966428168813603,0.04668677819528677);  \draw [linecolor] (0.17966428168813603,0.04668677819528677) -- (0.1908583286361587,0.03541549778830255);  \draw [linecolor] (0.1908583286361587,0.03541549778830255) -- (0.20205237558418135,0.02392803451304745);  \draw [linecolor] (0.20205237558418135,0.02392803451304745) -- (0.21324642253220405,0.012294510008777905);  \draw [linecolor] (0.21324642253220405,0.012294510008777905) -- (0.2244404694802267,0.0005859375000000132); \end{tikzpicture}}

\DeclareRobustCommand{\shiftedcosinehalfsmall}{\begin{tikzpicture} \definecolor{linecolor}{HTML}{B8860B} \draw [linecolor] (0.0002797663271159175,0.1499988555864547) -- (0.01145982581253691,0.14810796155886635);  \draw [linecolor] (0.01145982581253691,0.14810796155886635) -- (0.0226398852979579,0.14291450316754442);  \draw [linecolor] (0.0226398852979579,0.14291450316754442) -- (0.033819944783378866,0.13516561757417941);  \draw [linecolor] (0.033819944783378866,0.13516561757417941) -- (0.045000004268799854,0.12577576833022164);  \draw [linecolor] (0.045000004268799854,0.12577576833022164) -- (0.05618006375422085,0.11557634066657296);  \draw [linecolor] (0.05618006375422085,0.11557634066657296) -- (0.06736012323964181,0.10518977127707692);  \draw [linecolor] (0.06736012323964181,0.10518977127707692) -- (0.0785401827250628,0.09501578379660074);  \draw [linecolor] (0.0785401827250628,0.09501578379660074) -- (0.08972024221048376,0.08527447069478694);  \draw [linecolor] (0.08972024221048376,0.08527447069478694) -- (0.10090030169590476,0.07606184397625149);  \draw [linecolor] (0.10090030169590476,0.07606184397625149) -- (0.11208036118132575,0.06739684922360561);  \draw [linecolor] (0.11208036118132575,0.06739684922360561) -- (0.12326042066674674,0.05925469347216997);  \draw [linecolor] (0.12326042066674674,0.05925469347216997) -- (0.1344404801521677,0.0515881792122454);  \draw [linecolor] (0.1344404801521677,0.0515881792122454) -- (0.14562053963758867,0.04434041211044807);  \draw [linecolor] (0.14562053963758867,0.04434041211044807) -- (0.15680059912300967,0.037451891109540184);  \draw [linecolor] (0.15680059912300967,0.037451891109540184) -- (0.16798065860843064,0.030864151057820054);  \draw [linecolor] (0.16798065860843064,0.030864151057820054) -- (0.17916071809385165,0.02452137417646468);  \draw [linecolor] (0.17916071809385165,0.02452137417646468) -- (0.1903407775792726,0.01837084349461752);  \draw [linecolor] (0.1903407775792726,0.01837084349461752) -- (0.2015208370646936,0.012362756550765332);  \draw [linecolor] (0.2015208370646936,0.012362756550765332) -- (0.21270089655011457,0.006449697898464192);  \draw [linecolor] (0.21270089655011457,0.006449697898464192) -- (0.22388095603553554,0.0005859374999999992); \end{tikzpicture}}

\DeclareRobustCommand{\shiftedcosinefourthsmall}{\begin{tikzpicture} \definecolor{linecolor}{HTML}{FF8C00} \draw [linecolor] (0.0002797663271159175,0.1499988555864547) -- (0.01145982581253691,0.14810796155886635);  \draw [linecolor] (0.01145982581253691,0.14810796155886635) -- (0.0226398852979579,0.14291450316754442);  \draw [linecolor] (0.0226398852979579,0.14291450316754442) -- (0.033819944783378866,0.13516561757417941);  \draw [linecolor] (0.033819944783378866,0.13516561757417941) -- (0.045000004268799854,0.12577576833022164);  \draw [linecolor] (0.045000004268799854,0.12577576833022164) -- (0.05618006375422085,0.11557634066657296);  \draw [linecolor] (0.05618006375422085,0.11557634066657296) -- (0.06736012323964181,0.10518977127707692);  \draw [linecolor] (0.06736012323964181,0.10518977127707692) -- (0.0785401827250628,0.09501578379660074);  \draw [linecolor] (0.0785401827250628,0.09501578379660074) -- (0.08972024221048376,0.08527447069478694);  \draw [linecolor] (0.08972024221048376,0.08527447069478694) -- (0.10090030169590476,0.07606184397625149);  \draw [linecolor] (0.10090030169590476,0.07606184397625149) -- (0.11208036118132575,0.06739684922360561);  \draw [linecolor] (0.11208036118132575,0.06739684922360561) -- (0.12326042066674674,0.05925469347216997);  \draw [linecolor] (0.12326042066674674,0.05925469347216997) -- (0.1344404801521677,0.0515881792122454);  \draw [linecolor] (0.1344404801521677,0.0515881792122454) -- (0.14562053963758867,0.04434041211044807);  \draw [linecolor] (0.14562053963758867,0.04434041211044807) -- (0.15680059912300967,0.037451891109540184);  \draw [linecolor] (0.15680059912300967,0.037451891109540184) -- (0.16798065860843064,0.030864151057820054);  \draw [linecolor] (0.16798065860843064,0.030864151057820054) -- (0.17916071809385165,0.02452137417646468);  \draw [linecolor] (0.17916071809385165,0.02452137417646468) -- (0.1903407775792726,0.01837084349461752);  \draw [linecolor] (0.1903407775792726,0.01837084349461752) -- (0.2015208370646936,0.012362756550765332);  \draw [linecolor] (0.2015208370646936,0.012362756550765332) -- (0.21270089655011457,0.006449697898464192);  \draw [linecolor] (0.21270089655011457,0.006449697898464192) -- (0.22388095603553554,0.0005859374999999992); \end{tikzpicture}}

\DeclareRobustCommand{\shiftedcosineeighthsmall}{\begin{tikzpicture} \definecolor{linecolor}{HTML}{FF0000} \draw [linecolor] (0.0002797663271159175,0.1499988555864547) -- (0.01145982581253691,0.14810796155886635);  \draw [linecolor] (0.01145982581253691,0.14810796155886635) -- (0.0226398852979579,0.14291450316754442);  \draw [linecolor] (0.0226398852979579,0.14291450316754442) -- (0.033819944783378866,0.13516561757417941);  \draw [linecolor] (0.033819944783378866,0.13516561757417941) -- (0.045000004268799854,0.12577576833022164);  \draw [linecolor] (0.045000004268799854,0.12577576833022164) -- (0.05618006375422085,0.11557634066657296);  \draw [linecolor] (0.05618006375422085,0.11557634066657296) -- (0.06736012323964181,0.10518977127707692);  \draw [linecolor] (0.06736012323964181,0.10518977127707692) -- (0.0785401827250628,0.09501578379660074);  \draw [linecolor] (0.0785401827250628,0.09501578379660074) -- (0.08972024221048376,0.08527447069478694);  \draw [linecolor] (0.08972024221048376,0.08527447069478694) -- (0.10090030169590476,0.07606184397625149);  \draw [linecolor] (0.10090030169590476,0.07606184397625149) -- (0.11208036118132575,0.06739684922360561);  \draw [linecolor] (0.11208036118132575,0.06739684922360561) -- (0.12326042066674674,0.05925469347216997);  \draw [linecolor] (0.12326042066674674,0.05925469347216997) -- (0.1344404801521677,0.0515881792122454);  \draw [linecolor] (0.1344404801521677,0.0515881792122454) -- (0.14562053963758867,0.04434041211044807);  \draw [linecolor] (0.14562053963758867,0.04434041211044807) -- (0.15680059912300967,0.037451891109540184);  \draw [linecolor] (0.15680059912300967,0.037451891109540184) -- (0.16798065860843064,0.030864151057820054);  \draw [linecolor] (0.16798065860843064,0.030864151057820054) -- (0.17916071809385165,0.02452137417646468);  \draw [linecolor] (0.17916071809385165,0.02452137417646468) -- (0.1903407775792726,0.01837084349461752);  \draw [linecolor] (0.1903407775792726,0.01837084349461752) -- (0.2015208370646936,0.012362756550765332);  \draw [linecolor] (0.2015208370646936,0.012362756550765332) -- (0.21270089655011457,0.006449697898464192);  \draw [linecolor] (0.21270089655011457,0.006449697898464192) -- (0.22388095603553554,0.0005859374999999992); \end{tikzpicture}}

\DeclareRobustCommand{\linearbetasmall}{\begin{tikzpicture} \definecolor{linecolor}{HTML}{000000} \draw [linecolor] (0.225,0.0009529227131355045) -- (0.213075,0.0016050426390067341);  \draw [linecolor] (0.213075,0.0016050426390067341) -- (0.20115,0.00262742440335215);  \draw [linecolor] (0.20115,0.00262742440335215) -- (0.18945,0.004144873339967579);  \draw [linecolor] (0.18945,0.004144873339967579) -- (0.17752500000000002,0.006412943454831903);  \draw [linecolor] (0.17752500000000002,0.006412943454831903) -- (0.1656,0.009644016030098746);  \draw [linecolor] (0.1656,0.009644016030098746) -- (0.1539,0.014000043348281631);  \draw [linecolor] (0.1539,0.014000043348281631) -- (0.14197500000000002,0.01990263601439437);  \draw [linecolor] (0.14197500000000002,0.01990263601439437) -- (0.13005,0.027503346553128482);  \draw [linecolor] (0.13005,0.027503346553128482) -- (0.11835000000000001,0.036750438639043044);  \draw [linecolor] (0.11835000000000001,0.036750438639043044) -- (0.10642499999999999,0.048017096153326354);  \draw [linecolor] (0.10642499999999999,0.048017096153326354) -- (0.094725,0.06073175167511647);  \draw [linecolor] (0.094725,0.06073175167511647) -- (0.0828,0.07503415094320329);  \draw [linecolor] (0.0828,0.07503415094320329) -- (0.07087500000000001,0.0901282237123281);  \draw [linecolor] (0.07087500000000001,0.0901282237123281) -- (0.059175000000000005,0.10497266887796156);  \draw [linecolor] (0.059175000000000005,0.10497266887796156) -- (0.04725,0.1192517544964968);  \draw [linecolor] (0.04725,0.1192517544964968) -- (0.035325,0.1317196848431506);  \draw [linecolor] (0.035325,0.1317196848431506) -- (0.023625,0.14131061781395451);  \draw [linecolor] (0.023625,0.14131061781395451) -- (0.0117,0.1476467234954024);  \draw [linecolor] (0.0117,0.1476467234954024) -- (0.0,0.15); \end{tikzpicture}}

\DeclareRobustCommand{\invertedschedule}{\begin{tikzpicture} \definecolor{linecolor}{HTML}{4169E1} \draw [linecolor] (0.29999999982765396,0.0007812500320081821) -- (0.2849999998448886,0.07578028606012077);  \draw [linecolor] (0.2849999998448886,0.07578028606012077) -- (0.2699999998621232,0.09051690785411814);  \draw [linecolor] (0.2699999998621232,0.09051690785411814) -- (0.2549999998793578,0.10063464840006582);  \draw [linecolor] (0.2549999998793578,0.10063464840006582) -- (0.23999999989659238,0.1086585911145849);  \draw [linecolor] (0.23999999989659238,0.1086585911145849) -- (0.224999999913827,0.11547005387449812);  \draw [linecolor] (0.224999999913827,0.11547005387449812) -- (0.20999999993106158,0.12149240630182001);  \draw [linecolor] (0.20999999993106158,0.12149240630182001) -- (0.1949999999482962,0.12696665957554973);  \draw [linecolor] (0.1949999999482962,0.12696665957554973) -- (0.1799999999655308,0.13204632266378027);  \draw [linecolor] (0.1799999999655308,0.13204632266378027) -- (0.1649999999827654,0.13683796543466983);  \draw [linecolor] (0.1649999999827654,0.13683796543466983) -- (0.14999999999999997,0.14142135623730953);  \draw [linecolor] (0.14999999999999997,0.14142135623730953) -- (0.13500000001723456,0.14586079396362858);  \draw [linecolor] (0.13500000001723456,0.14586079396362858) -- (0.12000000003446916,0.15021241184060935);  \draw [linecolor] (0.12000000003446916,0.15021241184060935) -- (0.10500000005170376,0.15452982675272264);  \draw [linecolor] (0.10500000005170376,0.15452982675272264) -- (0.09000000006893837,0.15886974290592118);  \draw [linecolor] (0.09000000006893837,0.15886974290592118) -- (0.07500000008617297,0.16329931615968427);  \draw [linecolor] (0.07500000008617297,0.16329931615968427) -- (0.060000000103407565,0.167908637588998);  \draw [linecolor] (0.060000000103407565,0.167908637588998) -- (0.04500000012064217,0.17283711274316968);  \draw [linecolor] (0.04500000012064217,0.17283711274316968) -- (0.030000000137876768,0.1783443001402879);  \draw [linecolor] (0.030000000137876768,0.1783443001402879) -- (0.01500000015511137,0.18508740703961002);  \draw [linecolor] (0.01500000015511137,0.18508740703961002) -- (1.7234597586591708e-10,0.19999847411527294); \end{tikzpicture}}

\DeclareRobustCommand{\invertedschedulesmall}{\begin{tikzpicture} \definecolor{linecolor}{HTML}{4169E1} \draw [linecolor] (0.2249999998707405,0.0005859375240061365) -- (0.21374999988366644,0.056835214545090565);  \draw [linecolor] (0.21374999988366644,0.056835214545090565) -- (0.2024999998965924,0.0678876808905886);  \draw [linecolor] (0.2024999998965924,0.0678876808905886) -- (0.19124999990951835,0.07547598630004936);  \draw [linecolor] (0.19124999990951835,0.07547598630004936) -- (0.1799999999224443,0.08149394333593867);  \draw [linecolor] (0.1799999999224443,0.08149394333593867) -- (0.16874999993537027,0.08660254040587358);  \draw [linecolor] (0.16874999993537027,0.08660254040587358) -- (0.1574999999482962,0.09111930472636501);  \draw [linecolor] (0.1574999999482962,0.09111930472636501) -- (0.14624999996122215,0.0952249946816623);  \draw [linecolor] (0.14624999996122215,0.0952249946816623) -- (0.1349999999741481,0.0990347419978352);  \draw [linecolor] (0.1349999999741481,0.0990347419978352) -- (0.12374999998707405,0.10262847407600235);  \draw [linecolor] (0.12374999998707405,0.10262847407600235) -- (0.11249999999999999,0.10606601717798213);  \draw [linecolor] (0.11249999999999999,0.10606601717798213) -- (0.10125000001292593,0.10939559547272144);  \draw [linecolor] (0.10125000001292593,0.10939559547272144) -- (0.09000000002585187,0.112659308880457);  \draw [linecolor] (0.09000000002585187,0.112659308880457) -- (0.07875000003877783,0.11589737006454195);  \draw [linecolor] (0.07875000003877783,0.11589737006454195) -- (0.06750000005170378,0.11915230717944088);  \draw [linecolor] (0.06750000005170378,0.11915230717944088) -- (0.05625000006462973,0.12247448711976318);  \draw [linecolor] (0.05625000006462973,0.12247448711976318) -- (0.045000000077555675,0.12593147819174846);  \draw [linecolor] (0.045000000077555675,0.12593147819174846) -- (0.03375000009048163,0.12962783455737725);  \draw [linecolor] (0.03375000009048163,0.12962783455737725) -- (0.022500000103407577,0.13375822510521593);  \draw [linecolor] (0.022500000103407577,0.13375822510521593) -- (0.011250000116333527,0.1388155552797075);  \draw [linecolor] (0.011250000116333527,0.1388155552797075) -- (1.2925948189943782e-10,0.1499988555864547); \end{tikzpicture}}

%% file: sec/0_abstract.tex
\begin{abstract}
   Diffusion autoencoders (DAEs) are typically formulated as a noise prediction model and trained with a linear-$\beta$ noise schedule \cite{Ho:2020} that spends much of its sampling steps at high noise levels. Because high noise levels are associated with recovering large-scale image structures and low noise levels with recovering details, this configuration can result in low-quality and blurry images. However, it should be possible to improve details while spending fewer steps recovering structures because the latent code should already contain structural information. Based on this insight, we propose a new DAE training method that improves the quality of reconstructed images. We divide training into two phases. In the first phase, the DAE is trained as a vanilla autoencoder by always setting the noise level to the highest, forcing the encoder and decoder to populate the latent code with structural information. In the second phase, we incorporate a noise schedule that spends more time in the low-noise region \cite{Hoogeboom:2023}, allowing the DAE to learn how to perfect the details. Our method results in images that have accurate high-level structures and low-level details while still preserving useful properties of the latent codes.

\end{abstract}

%% file: sec/1_intro.tex
\section{Introduction}
\label{sec:intro}

The diffusion autoencoder (DAE), proposed in 2022 by Preechakul \etal, is a type of latent generative models in which diffusion models are used to both (1) generate latent representations and (2) decode them to full-fledged data items \cite{Preechakul:2022}. More specifically, a diffusion autoencoder has three components. The \emph{encoder} is a feed-forward network that encodes an image $\ve{x}$ into a 1D vector $\ve{z}$ called a \emph{latent code}.\footnote{The original paper calls this the ``semantic subcode.''} The \emph{decoder} is a conditional diffusion model \cite{SohlDickstein:2015, Ho:2020} tasked with reconstructing $\ve{x}$ given $\ve{z}$ through the diffusion process. The \emph{latent code model}\footnote{The original paper calls this the ``latent DDIM.''} is an unconditional diffusion model trained to sample $\mathbf{z}$ from its distribution. Preechakul \etal show that DAEs can learn latent codes that can be manipulated to yield meaningful changes in the image space and interpolated to produce smooth transitions similar to those produced by GANs \cite{Goodfellow:2014} and VAEs \cite{Kingma:2014}. Because of these capabilities, DAEs have been used or extended for numerous tasks and domains, including facial image generation \cite{Su:2023,Kim:DCFace:2023}, facial image manipulation \cite{Cho:2024,Bounareli:2024,Ding:2023,Jia:DisControlFace:2023,Du:2023,Wang:S3Editor:2024,Cho:2023,Liu:DiffDub:2024,Kim:DVAE:2023}, medical imaging \cite{Ijishakin:2024,He:DMCVR:2023,Keicher:2023}, security \cite{Blasingame:2024,Blasingame:FastDim:2024,Blasingame:GreedyDim:2024,Long:2024}, visual counterfactual explanation generation \cite{Komanduri:2024,Sobieski:2024}, 3D content generation \cite{Zeng:2022,Wang:RODIN:2023,Hu:2023}, and image compression \cite{Yang:2023}.

Despite DAEs' popularity, we only found few studies that optimize their model architecture and hyperparameters \cite{Hudson:2023,Yang:2023}. On the contrary, there are many such studies on vanilla diffusion models \cite{Nichol:2021,Dhariwal:2021,Karras:2022,Hoogeboom:2023,Karras:2023,Choi:2022,Peebles:2023,Crowson:2024}. 
As a result, how to best implement a DAE is still not well understood. More specifically, DAE is still largely stuck in the way diffusion models were implemented in 2020: the decoder is trained to predict noise under the ``linear-$\beta$'' schedule \cite{Ho:2020,Preechakul:2022}. However, there are now two alternative prediction types \cite{Song:DDIM:2020,Salimans:2022} and several other noise schedules \cite{Nichol:2021,Hoogeboom:2023,Hudson:2023,Song:2023,Chen:NoiseSchedule:2023}.

We show that the original implementation above can be significantly improved in terms of image reconstruction quality. We propose a new method for training DAEs without changing the architecture. It improves image reconstruction quality over the baseline \cite{Preechakul:2022} while preserving the interpretability, interpolability, and manipulability of the latent codes.

Our method consists of three parts. First, 
we change the noise schedule from the linear-$\beta$ schedule to one that have more sampling steps in the low-noise region \cite{Hoogeboom:2023}, hoping that the DAE would produce sharper and more accurate details. Second, we change what the model predicts from noise to velocity \cite{Salimans:2022}, as we discovered that noise prediction does not work with the new noise schedule. These two changes together produce a model with improved details but worsen overall image similarity. Third, we split training into two phases. In the first phase, the input is always pure Gaussian noise, so the model is trained as a vanilla autoencoder. In the second phase, the model is trained as a DAE using the proposed noise schedule. The two-phase algorithm makes the model a much better autoencoder that yields more faithful reconstruction of both details and large-scale structures.

%% file: sec/2_related_works.tex
\section{Related Works}
\label{sec:related-works}

Many generative models are \emph{latent variable models} whose process of generating a data item $\ve{x}$ involves first generating a \emph{latent code} $\ve{z}$ in a separate space and then sampling $\ve{x}$ conditioned on $\ve{z}$. In image generation context, there are two main types of latent codes. An \emph{image-like latent code} is a smaller image whose pixels are feature vectors. As a result, the composition of $\ve{x}$ can be encoded by simply placing the right feature at the right pixel. Spatial latent codes are mainly used to speed up the generation process \cite{Esser:2020,Rombach:2022}. A \emph{vector latent code} is a 1D vector with no spatial extent. They are employed by GANs and VAEs, and researchers have found that these codes can be interpolated \cite{Goodfellow:2014} and manipulated \cite{Zhu:2018} to yield high-level changes to images. Moreover, they also provide image representations that can be used for other downstream tasks.

Diffusion models are a class of generative models introduced by Sohl-Dickstein \etal in 2015 \cite{SohlDickstein:2015} and popularized by Ho \etal\ in 2020 \cite{Ho:2020}. Compared to GANs, the SOTA generative model at the time, diffusion models are much easier to train and soon become better at image generation \cite{Dhariwal:2021}. Diffusion models do not have vector latent codes (though something functionally similar can be extracted \cite{Kwon:2023}). However, there are many ways vector latent codes can interact with diffusion models. Diffusion models can sample vector latent codes that are later decoded by other types of generative models \cite{Rombach:2022,Vahdat:2021,Li:ROUG:2024}, or they can enhance images generated by other latent code models \cite{Pandey:2022}. The lack of ``native'' vector latent codes makes it harder to perform semantic image manipulation and representation learning in the sense that existing latent code manipulation techniques \cite{Zhu:2018, Harkonen:2020, Yang:LFM-GAN:2021} would no longer apply. It is thus interesting to ask whether one can design a diffusion model with them.


DAE is perhaps the simplest of such a design. It is an autoencoder whose encoder produces a 1D vector and whose decoder is a diffusion model that takes the vector as a conditioning signal \cite{Preechakul:2022} through adaptive normalization layers \cite{Huang:2017:AdaIN}. It can perform many tasks mentioned in the introduction, and its design has been modified in multiple directions. These include creating a DAE from a pretrained vanilla diffusion model \cite{Zhang:PDAE:2023}, extending the encoder \cite{Mittal:2023,Leng:DiffGAE:2023}, adding more structures to vector latent codes \cite{Yue:DITI:2024,Lu:2023,Yang:DisDiff:2023}, introducing new losses \cite{Wang:InfoDiffusion:2023,Jin:2024,Hwa:2024} and so on \cite{Yang:CrossAttn:2024, Kim:DBAE:2024, Xu:DiscoDiff:2024,Wang:InfoDiffusion:2023}. Our work does not change DAE's design but gives a new training method that improves image reconstruction. 


The closest work to ours is by Yang and Mandt in which they study how DAEs can be used for image compression \cite{Yang:2023} and so focus on reconstruction quality. Nevertheless, their latent code produced by the encoder is actually image-like, and they do not evaluate whether it can still be interpolated or manipulated. Another closely related work is by Hudson \etal, which, like ours, proposes a new way to train DAEs \cite{Hudson:2023}. The authors introduce a new noise schedule and make the decoder perform view synthesis instead of reconstruction. In contrast, our work uses a different noise schedule and does not change what a DAE does. We evaluated Hudson \etal's training setup without incorporating view synthesis.

For brevity, we shall refer to ``vector latent code as'' as just ``latent code'' from now on.

%% file: sec/3_background.tex
\section{Background}
\label{sec:background}

We represent an image with a real vector $\ve{x} \in \Real^D$, and let $p_{\mrm{data}}$ denote the distribution of images. Following Kingma \etal~\cite{Kingma:2021}, we define a stochastic process $\{ \ve{x}_t : t \in [0,1] \}$ such that $\ve{x}_0 \sim p_{\mrm{data}}$ and $\ve{x}_t \sim \mcal{N}(\alpha_t \ve{x}_0, \sigma_t^2I)$ for all $0 \leq t \leq 1$. We may write $\ve{x}_t = \alpha_t \ve{x}_0 + \sigma_t \ves{\varepsilon}$, where $\ves{\varepsilon} \sim \mcal{N}(\ve{0},I)$. The scalar functions $\alpha_t$ and $\sigma_t$ are collectively known as the \emph{noise schedule}. In this paper, we use the \emph{variance preserving} formulation \cite{Song:2021}, where we require that $\alpha_0 = 1$, $\alpha_1 = 0$, and $\alpha_t^2 + \sigma_t^2 = 1$ for all $t$, so the noise schedule is defined by $\alpha_t$ alone. We can show that $E[\ve{x}_0 | \ve{x}_t] = (\ve{x}_t + \sigma_t^2 \nabla \log p_t(\ve{x}_t))/\alpha_t$, where $p_t(\cdot)$ denotes the probability density of $\ve{x}_t$ \cite{Efron:2011}. The quantity $\nabla \log p_t(\ve{x}_t)$ is called the \emph{score} \cite{Song:2019}.

{\bf Formulation and training.} A diffusion model is a neural network $\ve{f}_{\ves{\theta}}$ such that $\ve{f}_{\ves{\theta}}(\ve{x}_t, t)$ can be used to estimate the score. The original formulation \cite{SohlDickstein:2015, Ho:2020} predicts the Gaussian noise $\ves{\varepsilon}$ used to create $\ve{x}_t$ from $\ve{x}_0$. 
After the model is trained well, we have that $\nabla \log p_t(\ve{x}_t) \approx - \ve{f}_{\ves{\theta}}(\ve{x}_t, t)/\sigma_t$. We call this type of model a \emph{noise prediction} or a \emph{$\varepsilon$-prediction} model. Song \etal observe that one can also train the model to predict $\ve{x}_0$, which gives $\nabla \log p_t(\ve{x}_t) \approx (\alpha_t \ve{f}_{\ves{\theta}}(\ve{x}_t,t) - \ve{x}_t) /\sigma_t^2$ \cite{Song:DDIM:2020}. We call this type of model an \emph{$x$-predition model}. Lastly, Salimans and Ho propose the \emph{$v$-prediction model} where the model is trained to predict the \emph{velocity} $\ve{v}_t := \alpha_t \ves{\varepsilon} - \sigma_t \ve{x}_0$, which gives $\nabla \log p_t(\ve{x}_t) \approx -\ve{x}_t - \alpha_t \ve{f}_{\ves{\theta}}(\ve{x}_t, t) / \sigma_t$ \cite{Salimans:2022}.   



{\bf Sampling.} Using a trained diffusion model, we can generate a data sample by starting from a Gaussian noise sample and gradually transforming it. There are many algorithms to do so \cite{Ho:2020,Tachibana:2021,Song:2021,Dockhorn:2022,Karras:2022,Liu:2022,Lu:2022,Zhang:ExpInt:2022,Zhang:gDDIM:2022,Zhao:UniPC:2023}, but we rely on a simple algorithm called \emph{DDIM sampling} \cite{Song:DDIM:2020}. We fix $K+1$ timesteps $0 \lesssim t_0 < t_1 < \dotsb < t_K \lesssim 1$. Then, we start by sampling $\ve{x}_{t_K} \sim \mcal{N}(\ve{0},I)$. Once we have $\ve{x}_{t_k}$ for some $k$, we can compute $$\ve{x}_{t_{k-1}} := \frac{\alpha_{t_{k-1}} \ve{x}_t + (\alpha_{t_{k-1}} \sigma_{t_k}^2 -\sigma_{t_{k-1}}\sigma_{t_k})\nabla \log p_t(\ve{x}_t)}{\alpha_{t_k}},$$ and repeat this step until we reach $\ve{x}_{t_0}$, which we output. The score $\nabla \log p_t(\ve{x}_t)$ needs to be replaced with the appropriate approximation according to the model's prediction type.

{\bf Diffusion autoencoder.} A DAE has an encoder $E$, a feed-forward neural network which computes a latent code from an input image: $\ve{z} = E(\ve{x})$. 
The decoder is a conditional diffusion model $D(\ve{x}_t, t, \ve{z})$, which receives $\ve{z}$ as an extra input. Preechakul \etal formulate it as an $\varepsilon$-prediction model, but we can make it predict either $x$ or $v$ too. Applying DDIM sampling to $D$ with $\ve{z} = E(\ve{x})$ should result in a reconstruction of $\ve{x}$. However, it is not exact because the initial Gaussian noise $\ve{x}_{t_K}$ can introduce variation to the output. As a result, $\ve{x}_{t_K}$ is considered a part of the image's encoding and is called the \emph{stochastic subcode}.  A more accurate reconstruction can be obtained by computing an $x_{t_K}$ that is specific to the input image by a process called \emph{DDIM inversion} \cite{Song:DDIM:2020}. 
Though, we generally prefer a DAE where the reconstruction is as faithful as possible without DDIM inversion because it implies that the latent code contains more information and so can be more useful for other tasks.


{\bf Noise schedules.} Originally, $\alpha_t$ is defined as a discrete-time function. We fix timesteps $0 = t_0 < t_1 < \dotsb < t_K \lesssim 1$ where $K$ is typically 1000, and let $\alpha_{t_0} = \alpha_0 = 1$. We then define a sequence $(\beta_1, \beta_2, \dotsc, \beta_K)$ and set $\alpha_{t_k} = \sqrt{\prod_{j=1}^k (1 - \beta_j)}$. Ho \etal propose the \emph{linear-$\beta$ schedule} where $\beta_k$ varies linearly from $10^{-4}$ to $0.02$ \cite{Ho:2020}. Preechakul \etal use this schedule to train their encoders and decoders.

\begin{figure}
    \centering   
    \includegraphics[width=8cm]{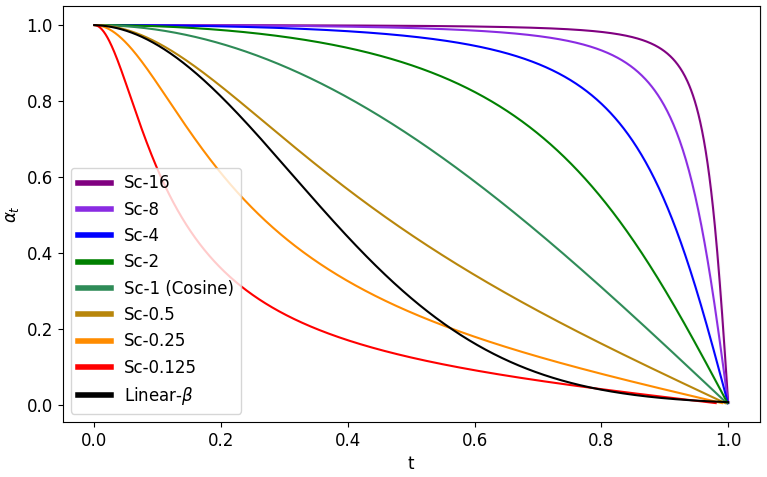}
    \caption{Some noise schedules in the shifted cosine family and the linear-$\beta$ noise schedule.}
    \label{fig:noise-schedules}
\end{figure}

Later works define $\alpha_t$ as a continuous function. Nichol and Dhariwal propose $\alpha_t \approx \cos(\pi t/2)$, the \emph{cosine schedule} \cite{Nichol:2021}. Hoogeboom \etal propose what we call the \emph{shifted cosine schedules} \cite{Hoogeboom:2023},
\begin{align*}
    \alpha_t = \textsc{Sc}(t;S) = \sqrt{\mathrm{sigmoid}( -2 \ln \tan (\pi t/2) + 2\ln S) },
\end{align*}
which form a family parameterized by a \emph{scaling factor} $S$. Observe that $\textsc{Sc}(t;1) = \cos (\pi t/2)$. We plot some shifted cosine schedules and the linear-$\beta$ schedule in Figure~\ref{fig:noise-schedules}. We can see that, as $S$ increases, $\texttt{Sc-}S$ spends less time in the high-noise region (i.e., the range of $t$ where $\alpha_t$ is low). 
Hoogeboom \etal found that, to train diffusion models for high-resolution images  ($256 \times 256$ or $512 \times 512$), it is better to use $\alpha_t = \textsc{Sc}(t; 32/d)$, where $d$ is the width of the images.
Chen \etal reports similar findings and also proposes another family of shifted noise schedules based on the sigmoid function \cite{Chen:NoiseSchedule:2023}, but we limit our experiments to the shifted cosine family. For brevity, we will refer to $\textsc{Sc}(t;S)$ as simply $\textsc{Sc-}S$ fron now on.

Hudson \etal propose a new ``inverted'' noise schedule that they claim to be beneficial to representation learning \cite{Hudson:2023}. Unfortunately, we could not find the exact formula in the paper, but we still evaluated Hudson \etal's training method ($\varepsilon$-prediction + inverted schedule) in our context and found that it did not work well. Details can be found in the supplementary material.

All the noise schedules we have discussed so far are variance preserving, but there are many that do not obey this constraint \cite{Song:2021,Karras:2022}. We do not use or evaluate them in this paper.

%% file: sec/4_method.tex
\section{Method}
\label{sec:method}

We propose a new method to train the encoder and decoder of a DAE~\cite{Preechakul:2022} (not the latent code model).
\begin{itemize}
    \item Make the decoder predict $v$ instead of $\varepsilon$.
    
    \item Divide training into two phases. First, jointly train the encoder and decoder as a vanilla autoencoder by always sampling $t = 1$ so that the input image is pure Gaussian noise. Second, train them normally as done in previous work \cite{Preechakul:2022}, incorporating a noise schedule. In our experiments, the first phase lasts one-fourth of the whole training session, but we have not optimized for the optimal division.

    \item The noise schedule used in the second phase should have more sampling steps in the low-noise region. We recommend $\textsc{Sc-}4$ (\shiftedcosinefour).
\end{itemize}
Pseudocode can be found in the supplementary material. We derived our method through a series of experiments on the CIFAR-10 dataset \cite{Krizhevsky:CIFAR10}. In this section, we discuss these experiments and how they informed our choices. 

\subsection{Motivation and Goal}
\label{sec:methods-motivation-and-goals}

Preechakul \etal report that a DAE can generate high-quality images with fewer sampling steps than what a typical diffusion model needs \cite{Preechakul:2022}.\footnote{In particular, see Section 5.5 and Figure 6 in the paper.} The reason is that the latent code $\ve{z}$, as an input, helps the decoder to denoise better. Preechakul \etal also show that the latent code captures high-level structures while the stochastic subcode captures low-level details.

Based on the above findings, the linear-$\beta$ noise schedule (\linearbeta) that Preechakul \etal use may be suboptimal. During the sampling process, high-level structures are recovered in timesteps with high noise levels, while low-level details are refined in those with low noise levels. The schedule allocates most of its timesteps in the high-noise region. As a result, the process would waste timesteps recovering information already present in the latent code instead of generating details. We thus hypothesize that \emph{training a DAE with a noise schedule that spends more time in the low-noise region, such as $\textsc{Sc-}S$ with $S \geq 1$ (i.e., \shiftedcosineone, \shiftedcosinetwo, \shiftedcosinefour, or \shiftedcosineeight), would yield better images.}

We now elaborate on what ``better images'' are. As DAE's main use is manipulating \emph{existing} images through semantic code manipulation and reconstruction,
we limit our work to improving its ability to reconstruct rather than to sample new random ones, as in unconditional sampling. 
There are two methods to reconstruct an image with a DAE.
In \emph{stochastic autoencoding}, we encode an input image to a latent code, then sample a random stochastic subcode, and feed both to the decoder. An \emph{inversion} is the same process, but the stochastic subcode is computed with DDIM inversion. 
We aim to improve both methods but first focus on stochastic autoencoding. Later, we will demonstrate that doing so also enhances inversion.

\begin{figure}
   \centering
   \scriptsize
   \begin{tabular}{c@{\hskip 0.1cm}c@{\hskip 0.1cm}c@{\hskip 0.1cm}c@{\hskip 0.1cm}c@{\hskip 0.1cm}c} 
      A & B & C & D & E & F \\
      \includegraphics*[width=1.25cm]{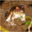} &
      \includegraphics*[width=1.25cm]{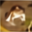} & 
      \includegraphics*[width=1.25cm]{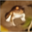} &
      \includegraphics*[width=1.25cm]{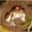} &
      \includegraphics*[width=1.25cm]{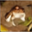} &
      \includegraphics*[width=1.25cm]{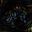} \\
      \includegraphics*[width=1.25cm]{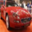} &
      \includegraphics*[width=1.25cm]{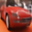} & 
      \includegraphics*[width=1.25cm]{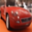} &
      \includegraphics*[width=1.25cm]{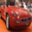} &
      \includegraphics*[width=1.25cm]{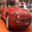} &
      \includegraphics*[width=1.25cm]{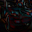} \\
      \includegraphics*[width=1.25cm]{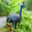} &
      \includegraphics*[width=1.25cm]{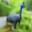} &
      \includegraphics*[width=1.25cm]{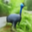} & 
      \includegraphics*[width=1.25cm]{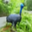} &
      \includegraphics*[width=1.25cm]{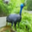} &
      \includegraphics*[width=1.25cm]{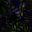} \\
      PSNR & \textcolor{red}{30.531} & 28.457 & 26.887 & 26.135 \\
      SSIM & \textcolor{red}{0.9517} & 0.9288 & 0.9048 & 0.8898 \\
      LPIPS & 0.0094 & 0.0087 & \textcolor{red}{0.0068} & 0.0072 \\
      FID (tf) & 37.245 & 17.744 & 5.245 & \textcolor{red}{1.723} \\
      FID (cf) & 46.392 & 21.252 & 6.251 & \textcolor{red}{2.157}
  \end{tabular}
  \caption{Some (A) CIFAR-10 images and their reconstructions by (B) a vanilla autoencoder trained to minimize the L2 loss, (C) stochastic autoencoding with a DAE with 10 DDIM sampling steps, and (D) stochastic autoencoding with the same DAE at 50 steps and (E) 1000 steps. We show (F) the absolute pixel differences between (D) and (E) to highlight the subtle differences between them.
  We also show the FID (computed by the \texttt{torch-fidelity} (tf) library \cite{Obukhov:2020} and the \texttt{clean-fid} (cf) library \cite{Parmar:2022} in that order) and averaged PSNR, SSIM, and LPIPS values computed from the reconstructions against the dataset. We can see that PNSR and SSIM overly award blurry images (Column B and C). LPIPS does not fall for the same trap but scores the sharpest (as measured by FID) reconstructions at 1000 steps worse than those at 50 steps. The FID, on the other hand, does not measure alignment between the reconstructed and the original, so it scores the most structurally accurate images in Column B the worst.}
  \label{fig:metrics-on-cifar-10}
\end{figure}

Among many choices for reconstruction quality metrics (at least 16 \cite{Yang:2023}), we 
use the following four, commonly used in previous works \cite{Preechakul:2022,Lu:2023,Hudson:2023}: Peak Signal-too-Noise Ratio (PSNR), Structural Similarity Index Measure (SSIM) \cite{Wang:2004}, Learned Perceptual Image Patch Similarity (LPIPS) \cite{Zhang:2018:LPIPS}, and Fr\'{e}chet Inception Distance (FID) \cite{Heusel:2017}.\footnote{FID values depend much on implementation details. We record those computed by two widely used libraries: \texttt{torch-fidelity} \cite{Obukhov:2020} and \texttt{clean-fid} \cite{Parmar:2022}.}  The first three metrics measure similarity between original and reconstructed images on a pair-by-pair basis but may fail in measuring sharpness and texture similarity. On the other hand, FID measures these aspects well by comparing the distributions of the original and reconstructed images \cite{Hudson:2023}.\footnote{Hudson \etal report that the FID captures sharpness and ``realism.'' However, we found the concept of realism to be vague and decided to replace it with ``texture similarity,'' which means ``similarity of details.''} However, it completely ignores the direct similarity between the original and the reconstruction (Figure~\ref{fig:metrics-on-cifar-10}). Thus, we found it necessary to use all four metrics together.

We generate images with DDIM sampling, and the number of sampling steps can affect the metrics: FID improves significantly when we increase the step count. Nevertheless, more steps mean a longer sampling time. We fix the number of steps to 50, based on the fact that it is the recommended number of steps for a popular diffusion model, Stable Diffusion \cite{StableDiffusion:DefaultNFEs}. Users in general would be familiar with both the time required and the quality afforded by this particular step count.


\subsection{Problems with Noise Prediction}

\begin{figure}
    \scriptsize
    \begin{tabular}{c@{\hskip 0.1cm}c@{\hskip 0.1cm}c@{\hskip 0.1cm}c@{\hskip 0.1cm}c@{\hskip 0.1cm}c@{\hskip 0.1cm}c@{\hskip 0.1cm}c} 
        A & B & C & D & A & B & C & D \\
        \includegraphics*[width=0.9cm]{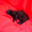} &
        \includegraphics*[width=0.9cm]{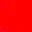} & 
        \includegraphics*[width=0.9cm]{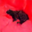} &
        \includegraphics*[width=0.9cm]{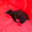} &
        \includegraphics*[width=0.9cm]{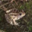} &
        \includegraphics*[width=0.9cm]{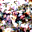} & 
        \includegraphics*[width=0.9cm]{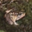} &
        \includegraphics*[width=0.9cm]{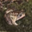} \\
        \includegraphics*[width=0.9cm]{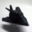} &
        \includegraphics*[width=0.9cm]{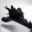} & 
        \includegraphics*[width=0.9cm]{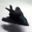} &
        \includegraphics*[width=0.9cm]{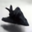} &
        \includegraphics*[width=0.9cm]{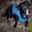} &
        \includegraphics*[width=0.9cm]{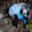} & 
        \includegraphics*[width=0.9cm]{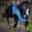} &
        \includegraphics*[width=0.9cm]{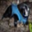} \\
    \end{tabular}
    \caption{Some (A) original CIFAR-10 images and their reconstruction by the (B) $\varepsilon$-predicting, (C) $x$-predicting, and (D) $v$-predicting DAEs in Table~\ref{table:prediction-type-ablation}. The $\varepsilon$-predicting DAE simply fails at being an accurate autoencoder.}
    \label{fig:prediction-type-ablation}
    \vspace{-1em}
\end{figure}

The simplest modification to Preechakul \etal's training method is to change the linear-$\beta$ schedule (\linearbeta) to one that spends more time in the low-noise region. Nevertheless, we learned through an experiment that an $\varepsilon$-prediction model does not work well with such a noise schedule. In the experiment, we chose $\textsc{Sc-4}$ (\shiftedcosinefour) as a representative and trained three DAEs, which predict $\varepsilon$, $x$, and $v$, for 32M examples (640 epochs). The size of the latent code is $512$, following the original paper \cite{Preechakul:2022}. Other training details can be found in the supplementary material.
We report metrics of reconstructions obtained by stochastic autoencoding in Table~\ref{table:prediction-type-ablation}, which shows that the $\varepsilon$-prediction model performs poorly on the three image similarity metrics.
It cannot autoencode many images accurately, as can be seen in Figure~\ref{fig:prediction-type-ablation}, and so is not usable. On the other hand, the $x$-prediction and $v$-prediction models are still viable.


\begin{table}
    \scriptsize
    \centering
    \resizebox{\columnwidth}{!}{%
    \setlength{\tabcolsep}{3pt}
    \begin{tabular}{c|ccccc} 
        \toprule
        \textbf{Prediction type} &
        PSNR$\uparrow$&
        SSIM$\uparrow$&
        LPIPS$\downarrow$& 
        FID$\downarrow$ (tf)&
        FID$\downarrow$ (cf)\\ 
        \midrule
        $\varepsilon$-prediction & 15.534 & 0.4954 & 0.0373 & 4.7651 & 5.1589 \\
        $x$-prediction & 27.147 & 0.9075 & 0.0065 & 3.9349 & 5.8098 \\
        $v$-prediction & 23.664 & 0.8237 & 0.0116 & 1.8133 & 2.1323 \\
        \bottomrule
      \end{tabular}      
      }
    \caption{Quality of CIFAR-10 images reconstructed by DAEs trained on the $\textsc{Sc-}4$ schedule.}
    \label{table:prediction-type-ablation}
    \vspace{-1em}
\end{table}

The DAE's encoder and decoder must learn to autoencode images together. We first note that using a noise schedule such as $\textsc{Sc-}4$ (\shiftedcosinefour) makes this task harder for all models. When the noise level is low, almost all the information about the original image is already present in the decoder's noised input, reducing the need to utilize the encoder's output. As a result, the encoder and decoder have fewer opportunities to learn to cooperate under a noise schedule that spends more time in the low-noise region. That fact the $\varepsilon$-prediction model becomes unusable suggests that it struggles to learn how to autoencode more than the other two. This might be due to the fact that, at high noise levels, an $\varepsilon$-predicting decoder is asked to output something already similar to the noised input, so there is little pressure for the encoder and decoder to learn to cooperate during training.



While an $\varepsilon$-prediction model can become usable if we use schedules that spend more time in the high-noise region, such as $\textsc{Sc-2}$ (\shiftedcosinetwo) or $\textsc{Sc-}1$ (\shiftedcosineone), we decided to rule out formulating a DAE with $\varepsilon$-prediction when we try other schedules from the shifted cosine family in order to limit resources we use in further experiments.

\subsection{Impact of Noise Schedules}
With the prediction types narrowed down, we next identify the best noise schedule. We trained 10 DAEs using prediction types $x$ and $v$, and 5 different schedules (linear-$\beta$ \linearbeta\ and $\textsc{Sc}$ with $S=$ 1\shiftedcosineone,2\shiftedcosinetwo,4\shiftedcosinefour,8\shiftedcosineeight). Training details are unchanged from the previous section. Table~\ref{table:one-phase-noise-schedule-ablation} shows the image quality metrics, revealing three trends.

First, when a noise schedule spends less time in the high-noise region (i.e., going from \linearbeta\ to \shiftedcosineeight), the three image similarity metrics worsen.

Second, the FID, on the contrary, improves until it peaks at $\textsc{Sc-}4$ (\shiftedcosinefour) and then worsens at $\textsc{Sc-}8$ (\shiftedcosineeight).

Third, for the same noise schedule, $x$-prediction generally yields better image similarity scores than $v$-prediction but worse FID. 

\begin{table}
    \scriptsize
    \centering
    \resizebox{0.98\columnwidth}{!}{%
    \setlength{\tabcolsep}{3pt}
    \begin{tabular}{cc|ccccc}
        \toprule
        \textbf{Pred} &
        \textbf{Schedule} &
        PSNR$\uparrow$&
        SSIM$\uparrow$&
        LPIPS$\downarrow$& 
        FID$\downarrow$ (tf)&
        FID$\downarrow$ (cf)\\ 
        \midrule
        $x$ & linear-$\beta$ \linearbetasmall & \textcolor{red}{30.296} & \textcolor{red}{0.9516} & \textcolor{red}{0.0038} & 6.8907 & 8.4820 \\
        $x$ & $\textsc{Sc-}1$ \shiftedcosineonesmall & 29.019 & 0.9369 & 0.0046 & 5.0481 & 7.0456 \\
        $x$ & $\textsc{Sc-}2$ \shiftedcosinetwosmall & 28.245 & 0.9265 & 0.0053 & 4.8375 & 6.9604 \\
        $x$ & $\textsc{Sc-}4$ \shiftedcosinefoursmall & 27.147 & 0.9075 & 0.0065 & 3.9349 & 5.8098 \\
        $x$ & $\textsc{Sc-}8$ \shiftedcosineeightsmall & 25.057 & 0.8596 & 0.0101 & 4.4343 & 6.2864 \\ 
        & \\[-2ex]
        \hline & \\[-2ex]
        $v$ & linear-$\beta$ \linearbetasmall & 29.438 & 0.9437 & 0.0046 & 7.7344 & 9.4366 \\
        $v$ & $\textsc{Sc-}1$ \shiftedcosineonesmall & 28.626 & 0.9333 & 0.0048 & 4.3175 & 5.2237 \\
        $v$ & $\textsc{Sc-}2$ \shiftedcosinetwosmall & 27.094 & 0.9082 & 0.0064 & 2.7339 & 3.3155 \\
        $v$ & $\textsc{Sc-}4$ \shiftedcosinefoursmall & 23.664 & 0.8237 & 0.0116 & \textcolor{red}{1.8133} & \textcolor{red}{2.1323} \\
        $v$ & $\textsc{Sc-}8$ \shiftedcosineeightsmall & 20.842 & 0.7176 & 0.0193 & 2.0235 & 2.3003 \\
        \bottomrule
      \end{tabular}    
      }
    \caption{Impact of noise schedule and model prediction type on image quality of DAEs trained on CIFAR-10.}
    \label{table:one-phase-noise-schedule-ablation}
\end{table}

\begin{figure}
    \scriptsize
    \centering
    \begin{tabular}{c@{\hskip 0.1cm}c@{\hskip 0.1cm}c@{\hskip 0.1cm}c@{\hskip 0.1cm}c@{\hskip 0.1cm}c}
        Original &
        $x;$ linear-$\beta$ &
        $x; \textsc{Sc-}1$ \shiftedcosineonesmall &
        $x; \textsc{Sc-}2$ \shiftedcosinetwosmall &
        $x; \textsc{Sc-}$4 \shiftedcosinefoursmall &
        $x; \textsc{Sc-}$8 \shiftedcosineeightsmall \\
        \includegraphics*[width=1.2cm]{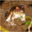} &
        \includegraphics*[width=1.2cm]{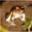} &
        \includegraphics*[width=1.2cm]{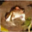} &
        \includegraphics*[width=1.2cm]{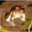} &
        \includegraphics*[width=1.2cm]{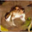} &
        \includegraphics*[width=1.2cm]{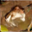} \\
        &
        $v;$ linear-$\beta$ &
        $v; \textsc{Sc-}1$ \shiftedcosineonesmall &
        $v; \textsc{Sc-}2$ \shiftedcosinetwosmall &
        $v; \textsc{Sc-}4$ \shiftedcosinefoursmall &
        $v; \textsc{Sc-}8$ \shiftedcosineeightsmall \\
        &
        \includegraphics*[width=1.2cm]{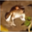} &
        \includegraphics*[width=1.2cm]{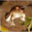} &
        \includegraphics*[width=1.2cm]{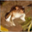} &
        \includegraphics*[width=1.2cm]{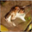} &
        \includegraphics*[width=1.2cm]{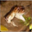}
    \end{tabular}
    \caption{Reconstructions of a frog image from CIFAR-10 by some of the DAEs from Table~\ref{table:one-phase-noise-schedule-ablation}. Notice that the images become sharper and noisier but less accurate as one moves to the right.}
    \label{fig:one-phase-noise-schedule-ablation}
\end{figure}

The first trend suggests that autoencoding is easier to learn at high noise levels. The second trend follows from the fact that the FID is greatly influenced by details, and more time in the low-noise region allows the decoder to perfect the details more. However, spending too much time in the low-noise region can impair the DAE's autoencoding ability, resulting in image distribution drift and poorer FID.
We surmise that this is the case for $\textsc{Sc-}8$ (\shiftedcosineeight). 
As evidence, we show reconstructions of a frog image from the data in Figure~\ref{fig:one-phase-noise-schedule-ablation}. 
Notice the reconstruction inaccuracies from $\textsc{Sc-}8$ (\shiftedcosineeight): the $x$-prediction model misses the green blob on the right, and the $v$-prediction model misses the frog's eyes.
The last trend is a direct result of model prediction type. 
Because an $x$-prediction model is always asked to recover the original image, it would naturally be better at autoencoding. 
However, it may not be as effective at recovering details compared to the $v$-prediction model due to their training differences. At low-noise levels, which are critical for detail recovery, the noised input already closely resembles the clean input that the $x$-prediction model has to predict. Because the prediction requires minimal effort, it provides fewer learning opportunities.



We thought that using a shifted cosine schedule would improve all scores, but instead we discovered a tradeoff. The combination with the best FID ($v$-prediction and $\textsc{Sc-}4$ \shiftedcosinefour) results in poor image similarity scores, while the combination with the best image similarity ($x$-prediction and linear-$\beta$ \linearbeta) results in a poor FID.

\subsection{Two-Phase Training Algorithm}
The tradeoff motivated us to seek a solution that improves both reconstruction accuracy and texture similarity simultaneously. Our idea is to start with the combination that yields the best FID score, $v$-prediction and $\textsc{Sc-}4$ (\shiftedcosinefour), and then modify it to improve reconstruction quality.
The main problem with the combo is that the DAE spends too little time in the high-noise region to learn to autoencode.

To fix this, we divide training into two phases, where the DAE is first trained to only autoencode and then trained normally to improve sharpness and texture similarity. The only difference between the two phases is how the time variable, $t$, is sampled during training: $t$ is always set to $1$ in the first phase but sampled from $[0,1]$ in the second. 

To see why the first phase teaches the model to autoencode, recall that when $t=1$, the noisy input $\ve{x}_1$ is a Gaussian noise. Because the decoder is a $v$-prediction model, it is asked to predict $\ve{v}_1 = \alpha_1 \ves{\epsilon} - \sigma_1 \ve{x}_0 = -\ve{x}_0$, which is (the negative of) a clean image. Because no information about $\ve{x}_0$ can be conveyed through $\ve{x}_1$, the encoder is forced to compress that information into the latent code $\ve{z}$ in order to help the decoder do its job. Hence, Phase 1 teaches the system to autoencode.

Unfortunately, because we add a training phase, it becomes unclear whether $\textsc{Sc-}4$ is still the optimal noise schedule. Those that spend even more time in the low-noise region---$\textsc{Sc-}8$ (\shiftedcosineeight) and $\textsc{Sc-}16$ (\shiftedcosinesixteen)---may perform better because the second phase starts with a trained autoencoder. 


To validate two-phase training and to find its best noise schedule, we trained a DAE in the first phase for 16M examples (80 epochs). We then trained four DAEs from this starting state for 48M examples (240 epochs) using shifted cosine schedules with $S = 2$\shiftedcosinetwo, $4$\shiftedcosinefour, $8$\shiftedcosineeight, and $16$\shiftedcosinesixteen. The metrics are available in Table~\ref{table:two-phase-noise-schedule-ablation} along with those of a DAE trained normally with $\textsc{Sc-}4$ (\shiftedcosinefour) for 64M examples (320 epochs) to match the total training length of other models.

\begin{table}
    \scriptsize
    \centering
    \resizebox{0.98\columnwidth}{!}{%
    \setlength{\tabcolsep}{2pt}
    \begin{tabular}{ll|ccccc}
        \toprule
        \textbf{Phase 1} &
        \textbf{Phase 2} & 
        PSNR$\uparrow$&
        SSIM$\uparrow$&
        LPIPS$\downarrow$& 
        FID$\downarrow$ (tf)&
        FID$\downarrow$ (cf)\\ 
        \midrule
        $\textsc{Sc-}4$ \shiftedcosinefoursmall & N/A & 25.955 & 0.8863 & 0.0072 & 1.2566 & 1.4634 \\
        \hline & \\ [-2ex]
        $t=1$ & N/A & 30.531 & 0.9517 & 0.0094 & 37.2447 & 46.3921 \\
        \hline & \\ [-2ex]
        $t=1$ & $\textsc{Sc-}2$ \shiftedcosinetwosmall & \textcolor{red}{30.711} & \textcolor{red}{0.9580} & \textcolor{red}{0.0030} & 3.2172 & 4.0042 \\
        $t=1$ & $\textsc{Sc-}4$ \shiftedcosinefoursmall & 30.351 & 0.9552 & 0.0032 & \textcolor{red}{2.1149} & 2.7153 \\
        $t=1$ & $\textsc{Sc-}8$ \shiftedcosineeightsmall & 28.647 & 0.9367 & 0.0050 & 2.1368 & \textcolor{red}{2.6579} \\
        $t=1$ & $\textsc{Sc-}16$ \shiftedcosinesixteensmall & 22.477 & 0.8176 & 0.0167 & 3.4845 & 4.0712 \\
        \bottomrule
    \end{tabular}
    }
    \caption{Performance of DAEs trained with 1-phase (first two rows) and 2-phase (last four rows) algorithms using different noise schedules
    on CIFAR-10. The term $t=1$ means that the input is always pure Gaussian noise.}
    \label{table:two-phase-noise-schedule-ablation}
\end{table}

Based on the results above, $\textsc{Sc-}4$ (\shiftedcosinefour) seems to yield the best balance between reconstruction accuracy and sharpness/texture similarity, and so it turns out to be the optimal noise schedule for two-phase training as well.

%% file: sec/5_results.tex
\section{Results}
\label{sec:results}

We evaluated the proposed two-phase training method against the baseline, the training method proposed by Preechakul \etal~\cite{Preechakul:2022}, which trains an $\varepsilon$-prediction model normally with the linear-$\beta$ schedule \cite{Preechakul:2022}. We also include results from training a $v$-prediction model with $\textsc{Sc-}4$ (\shiftedcosinefour) normally to serve as an ablation study. We also experimented with the training method from Hudson \etal's paper ($\varepsilon$-prediction model with the ``inverted'' schedule) \cite{Hudson:2023}, but we relegate the results to the supplementary material, as we found that it did not produce a usable autoencoder.

\subsection{Image Reconstruction}
\label{sec:image-reconstruction-results}

We trained DAEs on 4 datasets: CIFAR-10 at image resolution $32\times32$, CelebA \cite{Liu:CelebA:2015} at $64\times64$, LSUN Bedroom \cite{Yu:LSUN:2015} at $128\times128$, and ImageNet \cite{Deng:ImageNet:2009} at $32\times32$. As stated earlier, the first phase of the two-phase training algorithm takes one fourth of the total training length, and the training lengths for models on the same dataset are identical. Model and training details can be found in the supplementary material. 

For each datasets, we used the DAEs to reconstruct $50,000$ images with both stochastic autoencoding and inversion as discussed in Section~\ref{sec:methods-motivation-and-goals}. DDIM sampling and DDIM inversion both use 50 steps. Metrics are available in Table~\ref{table:image-reconstruction-exp}. We can see that our method always yield the best image similarity metrics and sometimes the best FID scores. Even when its FID scores are lower than those of the 1-phase training algorithm, the differences are minimal. Moreover, all of our metrics are better than those of Preechakul \etal's training method. The 1-phase training algorithm yielded the best FID scores most of the time, but its other metrics are worse than the baseline. This shows that the 2-phase training method yields both reconstruction accuracy and texture similarity. 

\begin{table*}
    \centering
    \resizebox{1.0\textwidth}{!}{%
    \setlength{\tabcolsep}{4pt}
    \scriptsize
    \begin{tabular}{ll|rrrrr|rrrrr}
        \toprule
        \multirow{2}{*}{\textbf{Dataset}}& \multirow{2}{*}{\textbf{Method}} & \multicolumn{5}{c|}{\textbf{Stochastic decoding}} & \multicolumn{5}{c}{\textbf{Inversion}} \\
        & & PSNR$\uparrow$ & SSIM$\uparrow$ & LPIPS$\downarrow$ & FID$\downarrow$ (tf) & FID$\downarrow$ (cf) & PSNR$\uparrow$ & SSIM$\uparrow$ & LPIPS$\downarrow$ & FID$\downarrow$ (tf) & FID$\downarrow$ (cf) \\
        \midrule
        & Preechakul \etal & 26.887 & 0.9048 & 0.006787 & 5.2448 & 6.2506 & 35.667 & 0.9834 & 0.002249 & 11.4542 & 14.8504 \\
        CIFAR-10 32  & 1-phase; $\textsc{Sc-}4$ \shiftedcosinefoursmall & 25.955 & 0.8863 & 0.007197 & \textcolor{red}{1.2566} & \textcolor{red}{1.4634} & 49.510 & 0.9995 & 0.000071 & \textcolor{red}{0.5434} & \textcolor{red}{0.6872} \\
        & 2-phase; $\textsc{Sc-}4$ \shiftedcosinefoursmall & \textcolor{red}{30.351} & \textcolor{red}{0.9552} & \textcolor{red}{0.003217} & 2.1149 & 2.7153 & \textcolor{red}{50.113} & \textcolor{red}{0.9996} & \textcolor{red}{0.000044} & 0.7560 & 0.9032 \\
        \hline & \\ [-2ex]
        & Preechakul \etal & 23.974 & 0.8186 & 0.030239 & 11.3050 & 14.3548 & 35.929 & 0.9744 & 0.008320 & 15.0237 & 19.5310 \\
        CelebA 64 & 1-phase; $\textsc{Sc-}4$ \shiftedcosinefoursmall & 23.943 & 0.8160 & 0.025720 & \textcolor{red}{1.8559} & \textcolor{red}{2.2449} & 45.969 & 0.9978 & 0.000766 & \textcolor{red}{1.2281} & \textcolor{red}{1.6972} \\
        & 2-phase; $\textsc{Sc-}4$ \shiftedcosinefoursmall & \textcolor{red}{26.122} & \textcolor{red}{0.8696} & \textcolor{red}{0.021513} & 2.3592 & 2.9030 & \textcolor{red}{47.646} & \textcolor{red}{0.9982} & \textcolor{red}{0.000700} & 1.5277 & 2.1107 \\
        \hline & \\ [-2ex]
        & Preechakul \etal & 16.120 & 0.4051 & 0.282071 & 28.8706 & 26.2747 & 29.039 & 0.8932 & 0.102119 & 16.8510 & 17.9375 \\
        Bedroom 128  & 1-phase; $\textsc{Sc-}4$ \shiftedcosinefoursmall & 17.964 & 0.4757 & 0.268361 & 3.3717 & \textcolor{red}{3.9362} & 30.412 & 0.9457 & 0.056554 & 1.6781 & 2.3033 \\
        & 2-phase; $\textsc{Sc-}4$ \shiftedcosinefoursmall & \textcolor{red}{20.309} & \textcolor{red}{0.5982} & \textcolor{red}{0.183325} & \textcolor{red}{3.2151} & 3.9847 & \textcolor{red}{37.268} & \textcolor{red}{0.9850} & \textcolor{red}{0.014716} & \textcolor{red}{1.1632} & \textcolor{red}{2.0454} \\
        \hline & \\ [-2ex]
        & Preechakul \etal & 21.168 & 0.7146 & 0.017876 & 12.4695 & 13.6909 & 34.794 & 0.9811 & 0.003229 & 8.8971 & 11.5388 \\
        ImageNet 32 & 1-phase; $\textsc{Sc-}4$ \shiftedcosinefoursmall & 20.708 & 0.6979 & 0.018849 & 8.5026 & 9.4456 & 45.444 & 0.9985 & 0.000183 & \textcolor{red}{0.5794} & \textcolor{red}{0.7365} \\
        & 2-phase; $\textsc{Sc-}4$ \shiftedcosinefoursmall & \textcolor{red}{27.811} & \textcolor{red}{0.9210} & \textcolor{red}{0.005147} & \textcolor{red}{4.1131} & \textcolor{red}{5.5029} & \textcolor{red}{48.892} & \textcolor{red}{0.9994} & \textcolor{red}{0.000045} & 0.6588 & 0.8226 \\
        \bottomrule
    \end{tabular}
    }
    \caption{Comparison of image reconstruction performance of the proposed training method with baselines.}
    \label{table:image-reconstruction-exp}
\end{table*}

\begin{figure}
    \centering
    \scriptsize
    \renewcommand{\arraystretch}{1.5}
    \begin{tabular}{@{\hskip 0.1cm}l@{\hskip 0.1cm}c@{\hskip 0.1cm}c@{\hskip 0.1cm}c@{\hskip 0.1cm}c@{\hskip 0.1cm}c@{\hskip 0.1cm}c@{\hskip 0.1cm}}        
        & \multicolumn{5}{@{\hskip 0.1cm}c@{\hskip 0.1cm}}{CelebA 64} \\
        &
        Original &
        Inversion &
        \multicolumn{3}{@{\hskip 0.1cm}c@{\hskip 0.1cm}}{Stochastic decodings} \\
        Preechakul \etal &
        \begin{minipage}{1.20cm}
        \centering
        \begin{tikzpicture}
            \node[anchor=south west,inner sep=0] (image) at (0,0) {
                \includegraphics*[width=1.15cm]{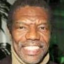} 
                };
            \begin{scope}[x={(image.south east)},y={(image.north west)}]
                \draw[red,thin] (.25, .7) rectangle (0.75, 0.90);
            \end{scope}
        \end{tikzpicture} 
        \end{minipage}& 
        \begin{minipage}{1.20cm}
        \centering
        \begin{tikzpicture}
            \node[anchor=south west,inner sep=0] (image) at (0,0) {
                \includegraphics*[width=1.15cm]{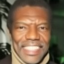} 
                };
            \begin{scope}[x={(image.south east)},y={(image.north west)}]
                \draw[red,thin] (.25, .7) rectangle (0.75, 0.90);
            \end{scope}
        \end{tikzpicture} 
        \end{minipage}&
        \begin{minipage}{1.20cm}
        \centering
        \begin{tikzpicture}
            \node[anchor=south west,inner sep=0] (image) at (0,0) {
                \includegraphics*[width=1.15cm]{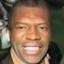} 
                };
            \begin{scope}[x={(image.south east)},y={(image.north west)}]
                \draw[red,thin] (.25, .7) rectangle (0.75, 0.90);
            \end{scope}
        \end{tikzpicture} 
        \end{minipage}&
        \begin{minipage}{1.20cm}
        \centering
        \begin{tikzpicture}
            \node[anchor=south west,inner sep=0] (image) at (0,0) {
                \includegraphics*[width=1.15cm]{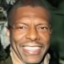} 
                };
            \begin{scope}[x={(image.south east)},y={(image.north west)}]
                \draw[red,thin] (.25, .7) rectangle (0.75, 0.90);
            \end{scope}
        \end{tikzpicture}
        \end{minipage}&
        \begin{minipage}{1.20cm}
        \centering
        \begin{tikzpicture}
            \node[anchor=south west,inner sep=0] (image) at (0,0) {
                \includegraphics*[width=1.15cm]{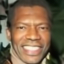} 
                };
            \begin{scope}[x={(image.south east)},y={(image.north west)}]
                \draw[red,thin] (.25, .7) rectangle (0.75, 0.90);
            \end{scope}
        \end{tikzpicture}
        \end{minipage}\\
        1-phase; $\textsc{Sc-}4$ &
        \begin{minipage}{1.20cm}
        \centering
        \begin{tikzpicture}
            \node[anchor=south west,inner sep=0] (image) at (0,0) {
                \includegraphics*[width=1.15cm]{images/reconstruction/celeba_64/originals/00000000.png} 
                };
            \begin{scope}[x={(image.south east)},y={(image.north west)}]
                \draw[red,thin] (.25, .7) rectangle (0.75, 0.90);
            \end{scope}
        \end{tikzpicture} 
        \end{minipage}&
        \begin{minipage}{1.20cm}
        \centering
        \begin{tikzpicture}
            \node[anchor=south west,inner sep=0] (image) at (0,0) {
                \includegraphics*[width=1.15cm]{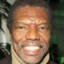} 
                };
            \begin{scope}[x={(image.south east)},y={(image.north west)}]
                \draw[red,thin] (.25, .7) rectangle (0.75, 0.90);
            \end{scope}
        \end{tikzpicture}
        \end{minipage}&
        \begin{minipage}{1.20cm}
        \centering
        \begin{tikzpicture}
            \node[anchor=south west,inner sep=0] (image) at (0,0) {
                \includegraphics*[width=1.15cm]{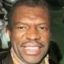} 
                };
            \begin{scope}[x={(image.south east)},y={(image.north west)}]
                \draw[red,thin] (.25, .7) rectangle (0.75, 0.90);
            \end{scope}
        \end{tikzpicture} 
        \end{minipage}&
        \begin{minipage}{1.20cm}
        \centering
        \begin{tikzpicture}
            \node[anchor=south west,inner sep=0] (image) at (0,0) {
                \includegraphics*[width=1.15cm]{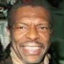} 
                };
            \begin{scope}[x={(image.south east)},y={(image.north west)}]
                \draw[red,thin] (.25, .7) rectangle (0.75, 0.90);
            \end{scope}
        \end{tikzpicture}
        \end{minipage}&
        \begin{minipage}{1.20cm}
        \centering
        \begin{tikzpicture}
            \node[anchor=south west,inner sep=0] (image) at (0,0) {
                \includegraphics*[width=1.15cm]{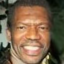} 
                };
            \begin{scope}[x={(image.south east)},y={(image.north west)}]
                \draw[red,thin] (.25, .7) rectangle (0.75, 0.90);
            \end{scope}
        \end{tikzpicture} 
        \end{minipage}\\
        2-phase; $\textsc{Sc-}4$ &
        \begin{minipage}{1.20cm}
        \centering
        \begin{tikzpicture}
            \node[anchor=south west,inner sep=0] (image) at (0,0) {
                \includegraphics*[width=1.15cm]{images/reconstruction/celeba_64/originals/00000000.png} 
                };
            \begin{scope}[x={(image.south east)},y={(image.north west)}]
                \draw[red,thin] (.25, .7) rectangle (0.75, 0.90);
            \end{scope}
        \end{tikzpicture} 
        \end{minipage}&
        \begin{minipage}{1.20cm}
        \centering
        \begin{tikzpicture}
            \node[anchor=south west,inner sep=0] (image) at (0,0) {
                \includegraphics*[width=1.15cm]{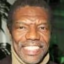} 
                };
            \begin{scope}[x={(image.south east)},y={(image.north west)}]
                \draw[red,thin] (.25, .7) rectangle (0.75, 0.90);
            \end{scope}
        \end{tikzpicture} 
        \end{minipage}&
        \begin{minipage}{1.20cm}
        \centering
        \begin{tikzpicture}
            \node[anchor=south west,inner sep=0] (image) at (0,0) {
                \includegraphics*[width=1.15cm]{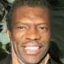} 
                };
            \begin{scope}[x={(image.south east)},y={(image.north west)}]
                \draw[red,thin] (.25, .7) rectangle (0.75, 0.90);
            \end{scope}
        \end{tikzpicture}
        \end{minipage}&
        \begin{minipage}{1.20cm}
        \centering
        \begin{tikzpicture}
            \node[anchor=south west,inner sep=0] (image) at (0,0) {
                \includegraphics*[width=1.15cm]{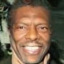} 
                };
            \begin{scope}[x={(image.south east)},y={(image.north west)}]
                \draw[red,thin] (.25, .7) rectangle (0.75, 0.90);
            \end{scope}
        \end{tikzpicture} 
        \end{minipage}&
        \begin{minipage}{1.20cm}
        \centering
        \begin{tikzpicture}
            \node[anchor=south west,inner sep=0] (image) at (0,0) {
                \includegraphics*[width=1.15cm]{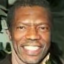} 
                };
            \begin{scope}[x={(image.south east)},y={(image.north west)}]
                \draw[red,thin] (.25, .7) rectangle (0.75, 0.90);
            \end{scope}
        \end{tikzpicture}
        \end{minipage}
        \\ [-1.5ex]
        \\[-1.5ex]
        Preechakul \etal &
        \begin{minipage}{1.20cm}         
        \centering
        \begin{tikzpicture}
            \node[anchor=south west,inner sep=0] (image) at (0,0) {
                \includegraphics[width=1.15cm, trim=12.8 44.8 12.8 6.4, clip]{images/reconstruction/celeba_64/originals/00000000.png}
            };
            \begin{scope}
                \draw[red,thick] (0.01,0.01) rectangle (1.14, 0.39);
            \end{scope}
        \end{tikzpicture}
        \end{minipage}&
        \begin{minipage}{1.20cm}         
        \centering
        \begin{tikzpicture}
            \node[anchor=south west,inner sep=0] (image) at (0,0) {
                \includegraphics[width=1.15cm, trim=12.8 44.8 12.8 6.4, clip]{images/reconstruction/celeba_64/diffae_000/0000_inversion.png}
            };
            \begin{scope}
                \draw[red,thick] (0.01,0.01) rectangle (1.14, 0.39);
            \end{scope}
        \end{tikzpicture}
        \end{minipage}&
        \begin{minipage}{1.20cm}         
        \centering
        \begin{tikzpicture}
            \node[anchor=south west,inner sep=0] (image) at (0,0) {
                \includegraphics[width=1.15cm, trim=12.8 44.8 12.8 6.4, clip]{images/reconstruction/celeba_64/diffae_000/0000_stochastic_0000.png}
            };
            \begin{scope}
                \draw[red,thick] (0.01,0.01) rectangle (1.14, 0.39);
            \end{scope}
        \end{tikzpicture} 
        \end{minipage}&
        \begin{minipage}{1.20cm}         
        \centering
        \begin{tikzpicture}
            \node[anchor=south west,inner sep=0] (image) at (0,0) {
                \includegraphics[width=1.15cm, trim=12.8 44.8 12.8 6.4, clip]{images/reconstruction/celeba_64/diffae_000/0000_stochastic_0001.png}
            };
            \begin{scope}
                \draw[red,thick] (0.01,0.01) rectangle (1.14, 0.39);
            \end{scope}
        \end{tikzpicture}
        \end{minipage}&
        \begin{minipage}{1.20cm}         
        \centering
        \begin{tikzpicture}
            \node[anchor=south west,inner sep=0] (image) at (0,0) {
                \includegraphics[width=1.15cm, trim=12.8 44.8 12.8 6.4, clip]{images/reconstruction/celeba_64/diffae_000/0000_stochastic_0002.png}
            };
            \begin{scope}
                \draw[red,thick] (0.01,0.01) rectangle (1.14, 0.39);
            \end{scope}
        \end{tikzpicture}
        \end{minipage}\\
        1-phase; $\textsc{Sc-}4$ &
        \begin{minipage}{1.20cm}         
        \centering
        \begin{tikzpicture}
            \node[anchor=south west,inner sep=0] (image) at (0,0) {
                \includegraphics[width=1.15cm, trim=12.8 44.8 12.8 6.4, clip]{images/reconstruction/celeba_64/originals/00000000.png}
            };
            \begin{scope}
                \draw[red,thick] (0.01,0.01) rectangle (1.14, 0.39);
            \end{scope}
        \end{tikzpicture}
        \end{minipage}&
        \begin{minipage}{1.20cm}         
        \centering
        \begin{tikzpicture}
            \node[anchor=south west,inner sep=0] (image) at (0,0) {
                \includegraphics[width=1.15cm, trim=12.8 44.8 12.8 6.4, clip]{images/reconstruction/celeba_64/diffae_001/0000_inversion.png}
            };
            \begin{scope}
                \draw[red,thick] (0.01,0.01) rectangle (1.14, 0.39);
            \end{scope}
        \end{tikzpicture}
        \end{minipage}&
        \begin{minipage}{1.20cm}         
        \centering
        \begin{tikzpicture}
            \node[anchor=south west,inner sep=0] (image) at (0,0) {
                \includegraphics[width=1.15cm, trim=12.8 44.8 12.8 6.4, clip]{images/reconstruction/celeba_64/diffae_001/0000_stochastic_0000.png}
            };
            \begin{scope}
                \draw[red,thick] (0.01,0.01) rectangle (1.14, 0.39);
            \end{scope}
        \end{tikzpicture}
        \end{minipage}&
        \begin{minipage}{1.20cm}         
        \centering
        \begin{tikzpicture}
            \node[anchor=south west,inner sep=0] (image) at (0,0) {
                \includegraphics[width=1.15cm, trim=12.8 44.8 12.8 6.4, clip]{images/reconstruction/celeba_64/diffae_001/0000_stochastic_0001.png}
            };
            \begin{scope}
                \draw[red,thick] (0.01,0.01) rectangle (1.14, 0.39);
            \end{scope}
        \end{tikzpicture}
        \end{minipage}&
        \begin{minipage}{1.20cm}         
        \centering
        \begin{tikzpicture}
            \node[anchor=south west,inner sep=0] (image) at (0,0) {
                \includegraphics[width=1.15cm, trim=12.8 44.8 12.8 6.4, clip]{images/reconstruction/celeba_64/diffae_001/0000_stochastic_0002.png}
            };
            \begin{scope}
                \draw[red,thick] (0.01,0.01) rectangle (1.14, 0.39);
            \end{scope}
        \end{tikzpicture}
        \end{minipage}\\
        2-phase; $\textsc{Sc-}4$ &
        \begin{minipage}{1.20cm}         
        \centering
        \begin{tikzpicture}
            \node[anchor=south west,inner sep=0] (image) at (0,0) {
                \includegraphics[width=1.15cm, trim=12.8 44.8 12.8 6.4, clip]{images/reconstruction/celeba_64/originals/00000000.png}
            };
            \begin{scope}
                \draw[red,thick] (0.01,0.01) rectangle (1.14, 0.39);
            \end{scope}
        \end{tikzpicture}
        \end{minipage}&
        \begin{minipage}{1.20cm}         
        \centering
        \begin{tikzpicture}
            \node[anchor=south west,inner sep=0] (image) at (0,0) {
                \includegraphics[width=1.15cm, trim=12.8 44.8 12.8 6.4, clip]{images/reconstruction/celeba_64/diffae_002/0000_inversion.png}
            };
            \begin{scope}
                \draw[red,thick] (0.01,0.01) rectangle (1.14, 0.39);
            \end{scope}
        \end{tikzpicture}
        \end{minipage}&
        \begin{minipage}{1.20cm}         
        \centering
        \begin{tikzpicture}
            \node[anchor=south west,inner sep=0] (image) at (0,0) {
                \includegraphics[width=1.15cm, trim=12.8 44.8 12.8 6.4, clip]{images/reconstruction/celeba_64/diffae_002/0000_stochastic_0000.png}
            };
            \begin{scope}
                \draw[red,thick] (0.01,0.01) rectangle (1.14, 0.39);
            \end{scope}
        \end{tikzpicture}
        \end{minipage}&
        \begin{minipage}{1.20cm}         
        \centering
        \begin{tikzpicture}
            \node[anchor=south west,inner sep=0] (image) at (0,0) {
                \includegraphics[width=1.15cm, trim=12.8 44.8 12.8 6.4, clip]{images/reconstruction/celeba_64/diffae_002/0000_stochastic_0001.png}
            };
            \begin{scope}
                \draw[red,thick] (0.01,0.01) rectangle (1.14, 0.39);
            \end{scope}
        \end{tikzpicture}
        \end{minipage}&
        \begin{minipage}{1.20cm}         
        \centering
        \begin{tikzpicture}
            \node[anchor=south west,inner sep=0] (image) at (0,0) {
                \includegraphics[width=1.15cm, trim=12.8 44.8 12.8 6.4, clip]{images/reconstruction/celeba_64/diffae_002/0000_stochastic_0002.png}
            };
            \begin{scope}
                \draw[red,thick] (0.01,0.01) rectangle (1.14, 0.39);
            \end{scope}
        \end{tikzpicture}
        \end{minipage}
        \\
        & & & & & & \\
        & \multicolumn{5}{@{\hskip 0.1cm}c@{\hskip 0.1cm}}{ImageNet 32} \\
        &
        Original &
        Inversion &
        \multicolumn{3}{@{\hskip 0.1cm}c@{\hskip 0.1cm}}{Stochastic decodings} \\
        Preechakul \etal &
        \parbox[c]{1.15cm}{\includegraphics*[width=1.15cm]{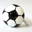}} & 
        \parbox[c]{1.15cm}{\includegraphics*[width=1.15cm]{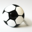}} & 
        \parbox[c]{1.15cm}{\includegraphics*[width=1.15cm]{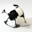}} & 
        \parbox[c]{1.15cm}{\includegraphics*[width=1.15cm]{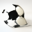}} &
        \parbox[c]{1.15cm}{\includegraphics*[width=1.15cm]{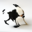}} \\
        1-phase; $\textsc{Sc-}4$ &
        \parbox[c]{1.15cm}{\includegraphics*[width=1.15cm]{images/reconstruction/imagenet_32/originals/00000000.png}} & 
        \parbox[c]{1.15cm}{\includegraphics*[width=1.15cm]{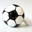}} & 
        \parbox[c]{1.15cm}{\includegraphics*[width=1.15cm]{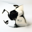}} & 
        \parbox[c]{1.15cm}{\includegraphics*[width=1.15cm]{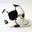}} &
        \parbox[c]{1.15cm}{\includegraphics*[width=1.15cm]{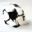}} \\
        2-phase; $\textsc{Sc-}4$ &
        \parbox[c]{1.15cm}{\includegraphics*[width=1.15cm]{images/reconstruction/imagenet_32/originals/00000000.png}} & 
        \parbox[c]{1.15cm}{\includegraphics*[width=1.15cm]{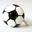}} & 
        \parbox[c]{1.15cm}{\includegraphics*[width=1.15cm]{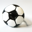}} & 
        \parbox[c]{1.15cm}{\includegraphics*[width=1.15cm]{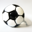}} &
        \parbox[c]{1.15cm}{\includegraphics*[width=1.15cm]{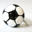}} 
    \end{tabular}
    \renewcommand{\arraystretch}{1}
    \caption{Image reconstruction on CelebA and ImageNet datasets.}
    \label{figure:image-reconstruction-exp}
    \vspace{-1em}
\end{figure}

\begin{figure}
    \scriptsize
    \centering
    \begin{tabular}{c@{\hskip 0.1cm}c@{\hskip 0.1cm}c@{\hskip 0.1cm}c}
        
        Original &
        Preechakul \etal &
        1-phase; $\textsc{Sc-}4$ (\shiftedcosinefoursmall) &
        2-phase; $\textsc{Sc-}4$ (\shiftedcosinefoursmall) \\
        
        \begin{tikzpicture}
        \node[anchor=south west,inner sep=0] (image) at (0,0) {
            \includegraphics*[width=1.9cm]{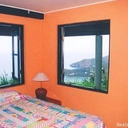} 
            };
        \begin{scope}[x={(image.south east)},y={(image.north west)}]
            \draw[red,thick] (.20, .5) rectangle (0.45, 0.25);
        \end{scope}
        \end{tikzpicture} &

        \begin{tikzpicture}
            \node[anchor=south west,inner sep=0] (image) at (0,0) {
                \includegraphics*[width=1.9cm]{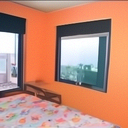} 
                };
            \begin{scope}[x={(image.south east)},y={(image.north west)}]
                \draw[red,thick] (.20, .5) rectangle (0.45, 0.25);
            \end{scope}
        \end{tikzpicture} &
        
        \begin{tikzpicture}
            \node[anchor=south west,inner sep=0] (image) at (0,0) {
                \includegraphics*[width=1.9cm]{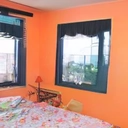} 
                };
            \begin{scope}[x={(image.south east)},y={(image.north west)}]
                \draw[red,thick] (.20, .5) rectangle (0.45, 0.25);
            \end{scope}
        \end{tikzpicture} &

        \begin{tikzpicture}
            \node[anchor=south west,inner sep=0] (image) at (0,0) {
                \includegraphics*[width=1.9cm]{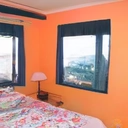} 
                };
            \begin{scope}[x={(image.south east)},y={(image.north west)}]
                \draw[red,thick] (.20, .5) rectangle (0.45, 0.25);
            \end{scope}
        \end{tikzpicture} \\

        \begin{tikzpicture}
            \node[anchor=south west,inner sep=0] (image) at (0,0) {
                \includegraphics[width=1.9cm, trim=26 32 70 64, clip]{images/reconstruction/bedroom_128/originals/00000002.png}
                };
            \begin{scope}
                \draw[red,thick] (0,0) rectangle (1.9, 1.9);
            \end{scope}
        \end{tikzpicture} &

        \begin{tikzpicture}
            \node[anchor=south west,inner sep=0] (image) at (0,0) {
                \includegraphics[width=1.9cm, trim=26 32 70 64, clip]{images/reconstruction/bedroom_128/diffae_000/0002_stochastic_0001.png}
                };
            \begin{scope}
                \draw[red,thick] (0,0) rectangle (1.9, 1.9);
            \end{scope}
        \end{tikzpicture} &

        \begin{tikzpicture}
            \node[anchor=south west,inner sep=0] (image) at (0,0) {
                \includegraphics[width=1.9cm, trim=26 32 70 64, clip]{images/reconstruction/bedroom_128/diffae_001/0002_stochastic_0001.png}
                };

            \begin{scope}
                \draw[red,thick] (0,0) rectangle (1.9, 1.9);
            \end{scope}
        \end{tikzpicture} &

        \begin{tikzpicture}
            \node[anchor=south west,inner sep=0] (image) at (0,0) {
                \includegraphics[width=1.9cm, trim=26 32 70 64, clip]{images/reconstruction/bedroom_128/diffae_002/0002_stochastic_0001.png}
                };
            \begin{scope}
                \draw[red,thick] (0,0) rectangle (1.9, 1.9);
            \end{scope}
        \end{tikzpicture} \\

    \end{tabular}
    
    \caption{Stochastic decodings of an image from the LSUN Bedroom dataset. The red bounding boxes indicate the most obvious differences between the three models.}
    \label{figure:image-reconstruction-zoomin}

\end{figure}

We show some of the reconstructions from the CelebA and ImageNet datasets in Figure~\ref{figure:image-reconstruction-exp}. Notice that the faces reconstructed by the DAE trained with the baseline method are blurrier than those reconstructed by other methods. 
The soccer ball images confirm that our two-phase training method yields much more accurate stochastic autoencoding than the comparisons. Additionally, we show images reconstructed by stochastic autoencoding from the LSUN Bedroom dataset in Figure~\ref{figure:image-reconstruction-zoomin}, and they further demonstrate that our method preserves image structures better than others. More results are available in the supplementary material. 

\subsection{Properties of Latent Codes}

We now show that the latent codes produced by our training method still retain the useful properties typically associated with DAE latent codes. Due to space constraints, more results are in the supplementary material.

\begin{figure}
    \scriptsize
    \centering
    \begin{tabular}{@{\hskip 0.0cm}c@{\hskip 0.07cm}c@{\hskip 0.07cm}c@{\hskip 0.07cm}c@{\hskip 0.07cm}c@{\hskip 0.07cm}c@{\hskip 0.07cm}c@{\hskip 0.0cm}}
        \multicolumn{7}{c}{CelebA 64 images, Preechakul \etal's training method} \\
        \includegraphics*[width=1.1cm]{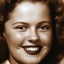} &
        \includegraphics*[width=1.1cm]{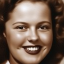} &
        \includegraphics*[width=1.1cm]{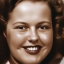} &
        \includegraphics*[width=1.1cm]{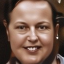} &
        \includegraphics*[width=1.1cm]{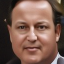} &
        \includegraphics*[width=1.1cm]{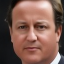} &
        \includegraphics*[width=1.1cm]{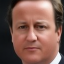} \\
        \multicolumn{7}{c}{CelebA 64 images, 1-phase training with $\textsc{Sc-}4$ schedule (\shiftedcosinefoursmall)} \\
        \includegraphics*[width=1.1cm]{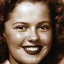} &
        \includegraphics*[width=1.1cm]{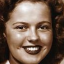} &
        \includegraphics*[width=1.1cm]{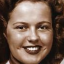} &
        \includegraphics*[width=1.1cm]{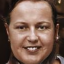} &
        \includegraphics*[width=1.1cm]{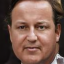} &
        \includegraphics*[width=1.1cm]{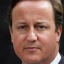} &
        \includegraphics*[width=1.1cm]{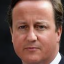} \\
        \multicolumn{7}{c}{CelebA 64 images, 2-phase training with $\textsc{Sc-}4$ schedule (\shiftedcosinefoursmall)} \\
        \includegraphics*[width=1.1cm]{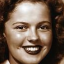} &
        \includegraphics*[width=1.1cm]{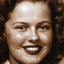} &
        \includegraphics*[width=1.1cm]{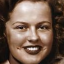} &
        \includegraphics*[width=1.1cm]{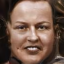} &
        \includegraphics*[width=1.1cm]{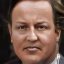} &
        \includegraphics*[width=1.1cm]{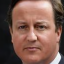} &
        \includegraphics*[width=1.1cm]{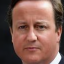} \\
        \multicolumn{7}{c}{Bedroom 128 images, Preechakul \etal's training method} \\
        \includegraphics*[width=1.1cm]{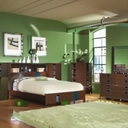} &
        \includegraphics*[width=1.1cm]{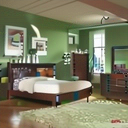} &
        \includegraphics*[width=1.1cm]{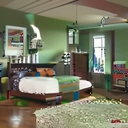} &
        \includegraphics*[width=1.1cm]{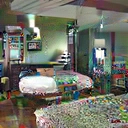} &
        \includegraphics*[width=1.1cm]{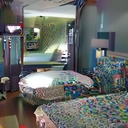} &
        \includegraphics*[width=1.1cm]{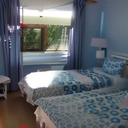} &
        \includegraphics*[width=1.1cm]{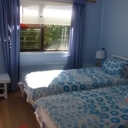} \\
        \multicolumn{7}{c}{Bedroom 128 images, 1-phase training with $\textsc{Sc-}4$ schedule (\shiftedcosinefoursmall)} \\
        \includegraphics*[width=1.1cm]{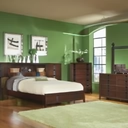} &
        \includegraphics*[width=1.1cm]{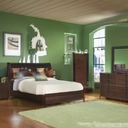} &
        \includegraphics*[width=1.1cm]{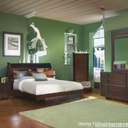} &
        \includegraphics*[width=1.1cm]{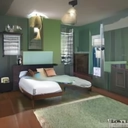} &
        \includegraphics*[width=1.1cm]{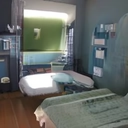} &
        \includegraphics*[width=1.1cm]{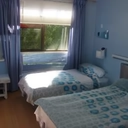} &
        \includegraphics*[width=1.1cm]{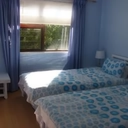} \\
        \multicolumn{7}{c}{Bedroom 128 images, 2-phase training with $\textsc{Sc-}4$ schedule (\shiftedcosinefoursmall)} \\
        \includegraphics*[width=1.1cm]{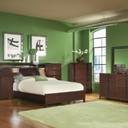} &
        \includegraphics*[width=1.1cm]{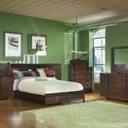} &
        \includegraphics*[width=1.1cm]{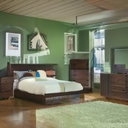} &
        \includegraphics*[width=1.1cm]{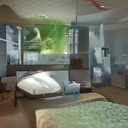} &
        \includegraphics*[width=1.1cm]{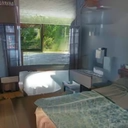} &
        \includegraphics*[width=1.1cm]{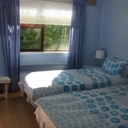} &
        \includegraphics*[width=1.1cm]{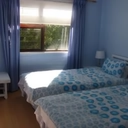}
    \end{tabular}
    \caption{Images generated by interpolating the latent codes and stochastic subcodes of pairs of images and then decoding with some of the DAEs in Section~\ref{sec:image-reconstruction-results}.}
    \label{figure:image-interpolation}
\end{figure}

\textbf{Interpolability.} In Figure~\ref{figure:image-interpolation}, we show images generated by spherical linear interpolation (slerp) of the semantic and stochastic subcodes of image pairs.  We can see that the in-betweens produced by the three training methods are similar in terms of difference between adjacent images, showing that our latent codes still retain interpolability. To quantify how smooth the interpolation was, we computed the perceptual path length (PPL) metrics \cite{Karras:StyleGan:2019} of 200 pairs of images from the four datasets and report their values in Table~\ref{table:ppl}. Our method achieves the lowest PPLs, indicating the smoothest interpolation.

\begin{table}
    \scriptsize
    \centering
    \resizebox{0.98\columnwidth}{!}{%
    \setlength{\tabcolsep}{4pt}
    \begin{tabular}{l|cccc}
        \toprule
        Method & CIFAR-10 & CelebA & Bedroom & ImageNet \\
        \midrule
        Preechakul \etal & 159.52 & 383.98 & 770.44 & 151.19 \\
        1-phase; $\textsc{Sc-}4$ \shiftedcosinefoursmall & 179.24 & 378.77 & 484.17 & 164.54 \\
        2-phase; $\textsc{Sc-}4$ \shiftedcosinefoursmall & \textcolor{red}{155.39} & \textcolor{red}{362.61} & \textcolor{red}{438.73} & \textcolor{red}{146.87} \\
        \bottomrule
    \end{tabular}
    }
    \caption{Average perceptual path length (PPL)$\downarrow$ between 200 image pairs. We created 101 in-betweens for each pair with DAEs trained using different methods. The PPLs were computed using the Euler method. See the supplementary for more details.}
    \label{table:ppl}
\end{table}

\textbf{Information within latent codes.} 
We evaluated how much information latent codes contain with \emph{linear classifier probes} \cite{Alain:2018}: we train a linear classifier on the codes and measure its performance.
We trained individual classifiers to predict each of the 40 face attributes that come with CelebA and used the area under receiver operating characteristic curve (AUROC) as metric. The macro averages over the attributes of the AUROC for Preechakul \etal's method, the 1-phase training method, and the proposed 2-phase training method are $0.9263$, $0.9254$, and $0.9195$, respectively. The AUROC values for individual properties are available in the supplementary material. While our proposed method came last, our score is only $0.0068$ or $0.73\%$ below the best baseline,
suggesting that much information is still retained in our latent codes.

We also trained multi-class linear classifiers to predict CIFAR-10 class membership using the negative log-likelihood loss on softmax logits computed from the latent codes. We tested them on latent codes of the CIFAR-10 test set.
The accuracies of the three methods (Preechakul \etal, 1-phase training with $\textsc{Sc-}4$ \shiftedcosinefour, and 2-phase training) are $72.71\%$, $70.61\%$, and $72.27\%$, and the macro averages of the AUROC values over 10 classes are $0.9624$, $0.9579$, and $0.9612$. While our proposed 2-phase training method did not achieve the best performance, the numbers were close to those of the baselines.

\begin{figure}
    \scriptsize
    \centering
    \def\imagewidth{1.45cm}
    \begin{tabular}{l@{\hskip 0.1cm}c@{\hskip 0.1cm}c@{\hskip 0.1cm}c@{\hskip 0.1cm}c@{\hskip 0.1cm}c}
        & Original & - Male & + Male & - Smile & + Smile\\
        A &
        \begin{minipage}{\imagewidth}\includegraphics*[width=\imagewidth]{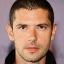}\end{minipage} &
        \begin{minipage}{\imagewidth}\includegraphics*[width=\imagewidth]{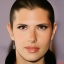}\end{minipage} &    
        \begin{minipage}{\imagewidth}\includegraphics*[width=\imagewidth]{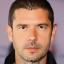}\end{minipage} &
        \begin{minipage}{\imagewidth}\includegraphics*[width=\imagewidth]{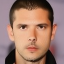}\end{minipage} &
        \begin{minipage}{\imagewidth}\includegraphics*[width=\imagewidth]{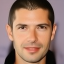}\end{minipage} \\   
        B &
        \begin{minipage}{\imagewidth}\includegraphics*[width=\imagewidth]{images/attr_manip/celeba_64/originals/00000000.png}\end{minipage} &
        \begin{minipage}{\imagewidth}\includegraphics*[width=\imagewidth]{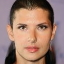}\end{minipage} &    
        \begin{minipage}{\imagewidth}\includegraphics*[width=\imagewidth]{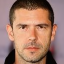}\end{minipage} &
        \begin{minipage}{\imagewidth}\includegraphics*[width=\imagewidth]{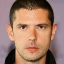}\end{minipage} &
        \begin{minipage}{\imagewidth}\includegraphics*[width=\imagewidth]{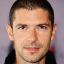}\end{minipage} \\
        C &
        \begin{minipage}{\imagewidth}\includegraphics*[width=\imagewidth]{images/attr_manip/celeba_64/originals/00000000.png}\end{minipage} &
        \begin{minipage}{\imagewidth}\includegraphics*[width=\imagewidth]{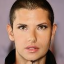}\end{minipage} &    
        \begin{minipage}{\imagewidth}\includegraphics*[width=\imagewidth]{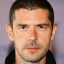}\end{minipage} &
        \begin{minipage}{\imagewidth}\includegraphics*[width=\imagewidth]{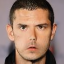}\end{minipage} &
        \begin{minipage}{\imagewidth}\includegraphics*[width=\imagewidth]{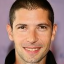}\end{minipage} \\
        A &
        \begin{minipage}{\imagewidth}\includegraphics*[width=\imagewidth]{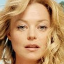}\end{minipage} &
        \begin{minipage}{\imagewidth}\includegraphics*[width=\imagewidth]{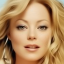}\end{minipage} &    
        \begin{minipage}{\imagewidth}\includegraphics*[width=\imagewidth]{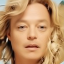}\end{minipage} &
        \begin{minipage}{\imagewidth}\includegraphics*[width=\imagewidth]{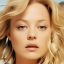}\end{minipage} &
        \begin{minipage}{\imagewidth}\includegraphics*[width=\imagewidth]{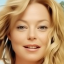}\end{minipage} \\
        B &
        \begin{minipage}{\imagewidth}\includegraphics*[width=\imagewidth]{images/attr_manip/celeba_64/originals/00000001.png}\end{minipage} &
        \begin{minipage}{\imagewidth}\includegraphics*[width=\imagewidth]{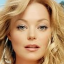}\end{minipage} &    
        \begin{minipage}{\imagewidth}\includegraphics*[width=\imagewidth]{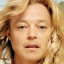}\end{minipage} &
        \begin{minipage}{\imagewidth}\includegraphics*[width=\imagewidth]{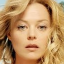}\end{minipage} &
        \begin{minipage}{\imagewidth}\includegraphics*[width=\imagewidth]{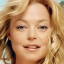}\end{minipage} \\
        C &
        \begin{minipage}{\imagewidth}\includegraphics*[width=\imagewidth]{images/attr_manip/celeba_64/originals/00000001.png}\end{minipage} &
        \begin{minipage}{\imagewidth}\includegraphics*[width=\imagewidth]{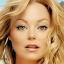}\end{minipage} &    
        \begin{minipage}{\imagewidth}\includegraphics*[width=\imagewidth]{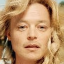}\end{minipage} &
        \begin{minipage}{\imagewidth}\includegraphics*[width=\imagewidth]{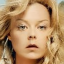}\end{minipage} &
        \begin{minipage}{\imagewidth}\includegraphics*[width=\imagewidth]{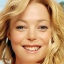}\end{minipage} \\
    \end{tabular}
    \caption{Attribute manipulation of CelebA images using DAEs trained by (A) Preechakul \etal's training method \cite{Preechakul:2022}, (B) the 1-phase training with $\textsc{Sc-}4$ (\shiftedcosinefour), and (C) our proposed 2-phase training with $\textsc{Sc-}4$ (\shiftedcosinefour).}
    \label{fig:attribute-manipulation}
    \vspace{-1em}
\end{figure}

\textbf{Attribute manipulation.} Lastly, we show that we can manipulate the latent codes to manipulate semantic attributes of the corresponding images using the same process as in the original DAE paper \cite{Preechakul:2022}. Figure~\ref{fig:attribute-manipulation} shows some of the manipulated images, which demonstrate the desired high-level changes, similar to the results from the original DAE.
Details on the manipulation process and more results can be found in the supplementary material.

%% file: sec/6_conclusion.tex
\section{Conclusion}
\label{sec:conclusion}

We argue that the training method for DAEs proposed in the original paper \cite{Preechakul:2022} and used by many subsequent studies is suboptimal: the noise schedule spends too many steps in the high-noise region, trying to recover high-level image structures. We surmised that the latent code already contains such information, so the decoder should allocate more steps in the low-noise region to improve details. However, we found that choosing a noise schedule involves a trade-off between details and structures.
We thus propose a two-phase training method that makes the DAE learn to accurately autoencode in the first and then perfect the details in the second, improving both aspects simultaneously over the baseline training method. 
We also show that the resulting latent codes retain the abilities to be interpolated and manipulated, while maintaining similar levels of semantic content to those learned by the original training method.

There are several directions for future work. While we show that $\textsc{Sc-}4$ (\shiftedcosinefour) leads to improvements on all metrics over the baseline, the shifted cosine family offers an infinite number of noise schedules due to the scaling factor $S$'s being a real number, and so $\textsc{Sc-}4$ may be suboptimal. Finding the optimal schedule can depend on the dataset, and image characteristics, such as resolution, may affect the optimal choice, similar to a finding by Hoogeboom \etal \cite{Hoogeboom:2023}.
Moreover, for the proposed two-phase training method, we have not yet optimized the relative duration of the two phases. How to determine this hyperparameter is an important open question.

%% file: sec/X_supplmenetary.tex
\newpage
\setcounter{page}{1}
\onecolumn
\begin{centering}
\Large
\textbf{\thetitle}\\
\vspace{0.5em}Supplementary Material \\
\vspace{1.0em}
\end{centering}

\section{Training Algorithms}

In this section, we show pseudocode for training diffusion autoencoders. Algorithm~\ref{alg:phase-1-training} trains a diffusion autoencoer as a vanilla autoencoder and is used in the first phase of our proposed method. Algorithm~\ref{alg:dae-training} is a generic for training a DAE that works with all types of model prediction: $\varepsilon$, $x$, and $v$. It used in the second phase of our method and also used to train baseline models that we compare against. Note that Algorithm~\ref{alg:phase-1-training} is a special case of Algorithm~\ref{alg:dae-training} where the decoder is a $v$-prediction model, and $t$ is always set to $1$.

\begin{algorithm}    
    \begin{algorithmic}[1]
\State Initialize $\theta$, the parameters of the encoder $E_\theta$ and the decoder $D_\theta$.
\While{target training length is not reached}
\State Sample a data item $\mathbf{x}_0 \sim p_{\mrm{data}}$.
\State $\ve{z} \gets E_{\theta}(\mathbf{x}_0)$
\State Sample $\ves{\varepsilon} \sim \mcal{N}(0,I)$.
\State $\ve{y}^* \gets -\ve{x}_0$
\State $\ve{y} \gets D_{\theta}(\ves{\varepsilon}, 1, \ve{z})$
\State $\mcal{L} \gets \| \ve{y} - \ve{y}_* \|^2$
\State Update $\theta$ according to the gradient $\nabla_{\theta} \mcal{L}$.
\EndWhile
    \end{algorithmic}
    \caption{Train a diffusion autoencoder as a vanilla encoder}
    \label{alg:phase-1-training}
\end{algorithm}

\begin{algorithm}    
    \begin{algorithmic}[1]
\State Initialize $\theta$, the parameters of the encoder $E_\theta$ and the decoder $D_\theta$.
\While{target training length is not reached}
\State Sample a data item $\mathbf{x}_0 \sim p_{\mrm{data}}$.
\State $\ve{z} \gets E_{\theta}(\mathbf{x}_0)$
\State Sample $t \sim \mcal{U}([0,1])$.
\State Sample $\ves{\varepsilon} \sim \mcal{N}(0,I)$.
\State $\ve{x}_t \gets \alpha_t \ve{x}_0 + \sigma_t \ves{\varepsilon}$
\If {$D_\theta$ is a $\varepsilon$-prediction model}
    \State $\ve{y}_* \gets \ves{\varepsilon}$
\ElsIf{$D_\theta$ is a $x$-prediction model}
    \State $\ve{y}_* \gets \ve{x}_0$
\ElsIf{$D_\theta$ is a $v$-prediction model}
    \State $\ve{y}_* \gets \alpha_t \ves{\varepsilon} - \sigma_t \ve{x}_0$
\EndIf
\State $\ve{y} \gets D_{\theta}(\ve{x}_t, t, \ve{z})$
\State $\mcal{L} \gets \| \ve{y} - \ve{y}_* \|^2$
\State Update $\theta$ according to the gradient $\nabla_{\theta} \mcal{L}$.
\EndWhile
    \end{algorithmic}
    \caption{Train a diffusion autoencoder normally}
    \label{alg:dae-training}
\end{algorithm}

\section{Network Architectures and Training Setups}

The DAE decoders and encoders are of the architectures used by Preechakul \etal~\cite{Preechakul:2022}, which in turn are based on the popular ADM architecture by Dhariwal and Nichol \cite{Dhariwal:2021}. The size of the latent code is fixed to $512$. Each decoder is a U-Net \cite{Ronneberger:2015} with attention layers \cite{Vaswani:2017}. The decoder is conditioned by two pieces of information: the time and the latent code. Each is transformed first by an MLP and then used to modulate, through adaptive group normalization (AdaGN) \cite{Dhariwal:2021} in ResNet blocks \cite{He:2016:ResNet}, the feature tensor that comes from the input image. Each encoder is the downsampling part of a U-Net with attention, and it has been configured so that the last output is a $512$-component 1D vector. More details can be found in the supplementary material of Preechakul \etal's paper \cite{Preechakul:2022}.  The DAEs that were trained on the same dataset had the same configuration, and these configurations are shown in Table~\ref{table:decoder-settings} and Table~\ref{table:encoder-settings}.

\begin{table*}[h]
    \centering        
    \begin{tabular}{l|cccc}
        \toprule
        Configuration & CIFAR-10 32 & CelebA 64 & Bedroom 128 & ImageNet 32 \\
        \midrule
        Base \#\,channel & 256 & 64 & 128 & 128 \\
        Channel multipliers & [1,1,1] & [1,2,4,8] & [1,1,2,3,4] & [1,2,2,2] \\
        \# Resnet blocks per resolution & 3 & 2 & 2 & 3 \\
        Attention resolutions & [16,8] & [32, 16, 8] & [16] & [16,8] \\
        Dropout probability & 0.2 & 0.1 & 0.1 & 0.3 \\
        Size (MB) & 265 & 316 & 427 & 296 \\
        \bottomrule
    \end{tabular}
    \caption{ADM configurations of the decoders.}
    \label{table:decoder-settings}    
\end{table*}

\begin{table*}[h]
    \centering
    \begin{tabular}{l|cccc}
        \toprule
        Configuration & CIFAR-10 32 & CelebA 64 & Bedroom 128 & ImageNet 32 \\
        \midrule
        Base \#\,channel & 128 & 128 & 128 & 128 \\
        Channel multipliers & [1,2,4,8] & [1,2,4,8] & [1,1,2,3,4,4] & [1,2,4,8] \\
        \#\,Resnet blocks per resolution & 2 & 2 & 2 & 2 \\
        Attention resolutions & [16,8,4] & [16,8,4] & [16] & [16, 8, 4] \\
        Dropout probability & 0.0 & 0.0 & 0.0 & 0.0 \\
        Size (MB) & 224 & 224 & 96 & 224 \\
        \bottomrule
    \end{tabular}
    \caption{ADM configurations of the encoders.}
    \label{table:encoder-settings}
\end{table*}

\begin{table*}[h]
    \centering
    \begin{tabular}{l|cccc}
        \toprule
        Hyperparameter & CIFAR-10 32 & CelebA 64 & Bedroom 128 & ImageNet 32 \\
        \midrule
        Batch size & 256 & 192 & 96 & 128 \\
        Training length (\#\,examples) & 64M & 32M & 32M & 64M \\
        Warm-up length (\#\,examples) & 128K & 128K & 5,120 & 128K \\
        \bottomrule
    \end{tabular}
    \caption{Hyperparameters used for training of DAEs in Section~\ref{sec:results} of the main paper.}
    \label{table:training-settings}
\end{table*}

All DAEs in the main paper were trained with the Adam optimizer \cite{Kingma:2015} with $\beta_1 = 0.9$ and $\beta_2 = 0.999$. The learning rate had a warm-up period where it increased linearly until it reached the maximum value of $10^{-4}$ and then remained unchanged. The warm-up period is $128,000$ training examples for all datasets except LSUN Bedroom \cite{Yu:LSUN:2015}, for which it was $5\,120$ training examples. During training, running averages of the models parameters are computed with the exponential moving average (EMA) algorithm with decay of $0.9999$, the same as what was done by Ho \etal\,\cite{Ho:2020}. The time step $t$ is sampled uniformly from the interval $[0,1]$. We always used models with EMA parameters for evaluation. Other hyperparameters such as the batch size and the training length vary from dataset to dataset. We list those used in Section~\ref{sec:results} in Table~\ref{table:training-settings}. For the two-phase training algorithm, the first phase took $1/4$ of the training length, and the second-phase took the rest. This means that all DAEs were shown the same number of training examples during training. The DAEs in Section 4 uses mostly the same hyperparmeters as those in Section~\ref{sec:results}, but some DAEs were trained for 32M examples instead of 64M examples.

\section{Evaluation of Hudson \etal's Training Method}

\begin{figure}
\centering
\begin{tabular}{@{\hskip 0.1cm}c@{\hskip 0.1cm}c@{\hskip 0.1cm}}
\includegraphics*[height=6cm]{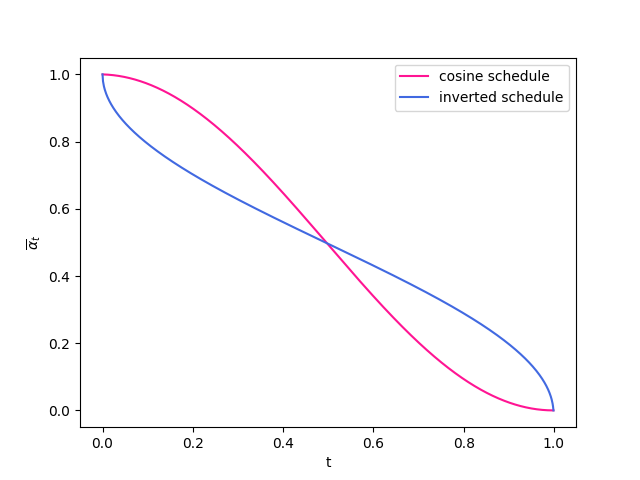} &
\includegraphics*[height=6cm]{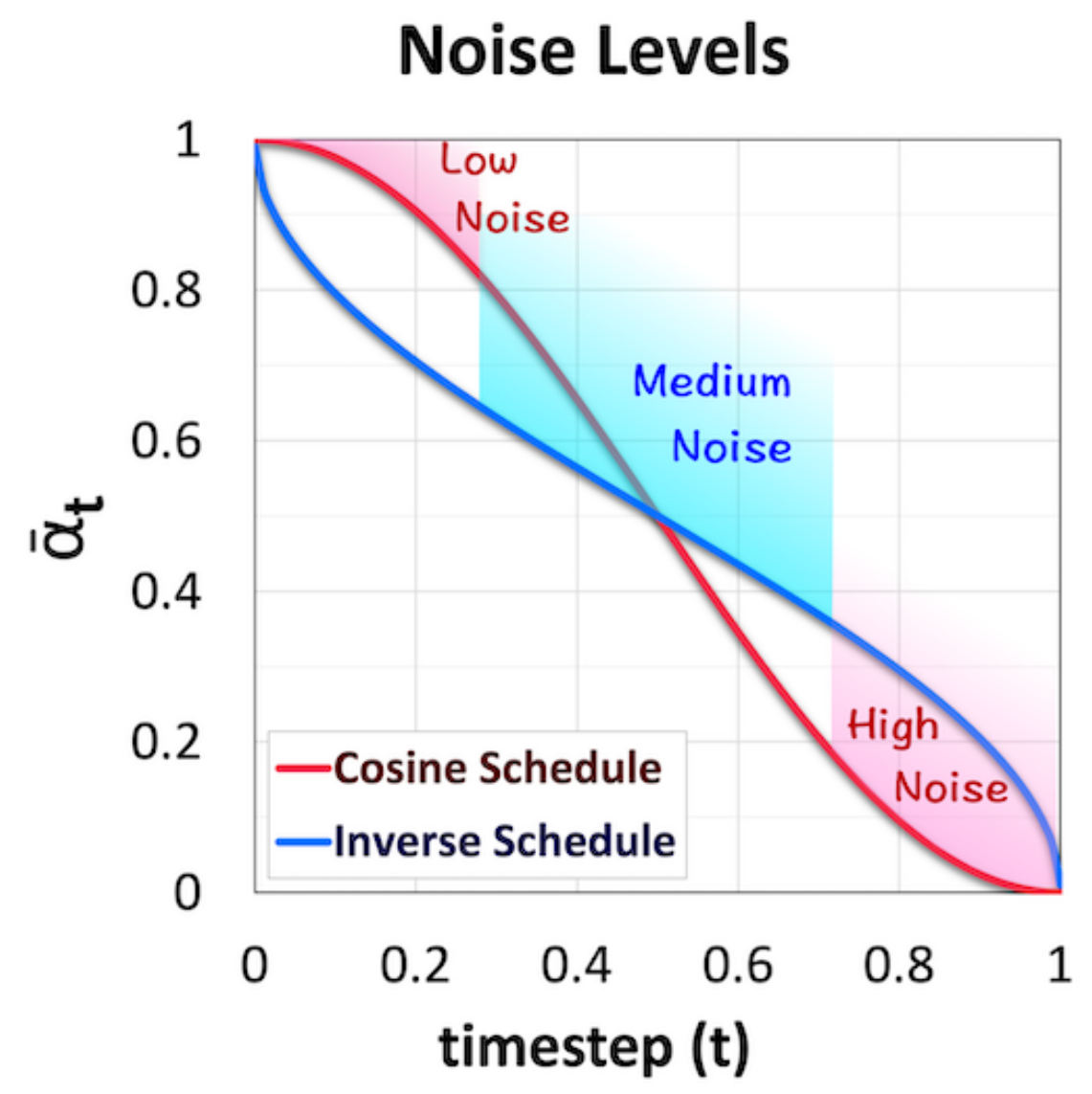}
\end{tabular}
\caption{To the left is our reconstruction of Hudson \etal's inverted noise schedule. We show it with with the cosine schedule, $\overline{\alpha}_t = \alpha_t^2 = \cos^2 (\pi t/2)$, so that the plot has the same composition as left plot in Figure 3 of Hudson \etal's paper (reproduced on the right). We found good agreement between the plots, and so we think the formula that we extracted from Xiang's code \cite{Xiang:SODA:2023} is right.}
\label{fig:inverted-schedule}
\end{figure}

Hudson \etal\ proposes a new noise schedule they call the ``inverted'' schedule \cite{Hudson:2023}. However, the schedule is only depicted in a figure. We could not find any mathematical expression for it inside the paper or the supplementary material, and there is no official code release at the time of writing of our paper. To make progress, we relied on an unofficial implementation by Xiang \cite{Xiang:SODA:2023}. Inspecting the code, we deduced that the formula for the inverted schedule was
\begin{align*}
\overline{\alpha}_t = \alpha_t^2 = \frac{2}{\pi} \cos^{-1}(\sqrt{t}),
\end{align*}
which gives
\begin{align*}
\alpha_t = \sqrt{\frac{2}{\pi} \cos^{-1}(\sqrt{t})}.
\end{align*}
To verify the formula, we plot the schedule with the cosine schedule and show the result in Figure~\ref{fig:inverted-schedule}. The plot looks like the relevant plot in Hudson \etal's paper, so the formula seems to be correct.

\begin{figure}
    \centering
    \includegraphics*[width=8.5cm]{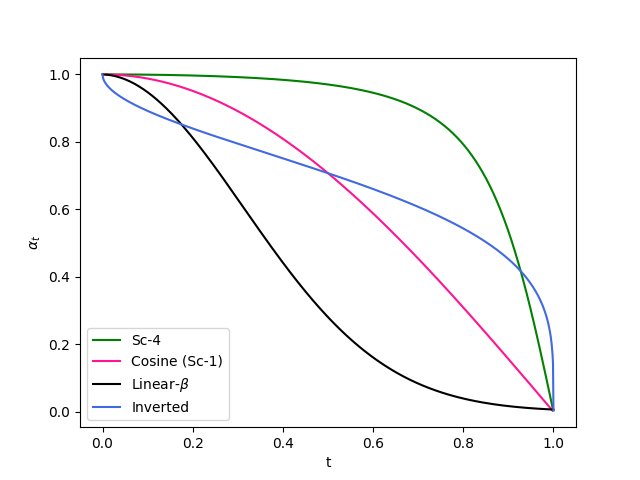}
    \caption{Comparison of Hudson \etal's inverted noise schedule against the linear-$\beta$, the cosine, and the $\textsc{Sc-}4$ schedules.}
    \label{fig:inverted-schedule-comparison}
\end{figure}

To put the inverted noise schedule in context, we plot its $\alpha_t$ function (not the $\overline{\alpha}_t$ function that is depicted in Figure~\ref{fig:inverted-schedule}) together with those of the linear-$\beta$ schedule, the cosine schedule (which is $\textsc{Sc-}1$), and the $\textsc{Sc-}4$ schedule in Figure~\ref{fig:inverted-schedule-comparison}. We can see that it indeed spends most of its time in the medium noise level, and it even spends less time in the high noise region than $\textsc{Sc-}4$.

Now that we have identified Hudson \etal's noise schedule, we also need to identify the prediction type that they use. This is not stated explicitly anywhere in the paper. However, on Page 4, the authors wrote ``$[\cdots]$ we extrapolated the model's output in the conditional direction: $\ves{\varepsilon}_{\theta}(\ve{x}_t|\ve{z}) - \ves{\varepsilon}_{\theta}(\ve{x}_t|\ve{0})$.'' The use of the symbol ``$\ves{\varepsilon}$'' led us to believe that Hudson \etal\ used the $\varepsilon$-prediction model.

\begin{figure*}[h]
    \centering
    \begin{tabular}{@{\hskip 0.1cm}l@{\hskip 0.2cm}c@{\hskip 0.1cm}c@{\hskip 0.1cm}c@{\hskip 0.1cm}c@{\hskip 0.1cm}c@{\hskip 0.1cm}}        
        & \multicolumn{4}{@{\hskip 0.1cm}c@{\hskip 0.1cm}}{CelebA 64} \\
        &
        Original &
        \multicolumn{3}{@{\hskip 0.1cm}c@{\hskip 0.1cm}}{Stochastic decodings} \\
        Preechakul \etal\ ($\varepsilon +$ \linearbeta) &
        \parbox[c]{2cm}{\includegraphics*[width=2cm]{images/reconstruction/celeba_64/originals/00000000.png}} & 
        \parbox[c]{2cm}{\includegraphics*[width=2cm]{images/reconstruction/celeba_64/diffae_000/0000_stochastic_0000.png}} & 
        \parbox[c]{2cm}{\includegraphics*[width=2cm]{images/reconstruction/celeba_64/diffae_000/0000_stochastic_0001.png}} &
        \parbox[c]{2cm}{\includegraphics*[width=2cm]{images/reconstruction/celeba_64/diffae_000/0000_stochastic_0002.png}} \\
        2-phase; $\textsc{Sc-}4$ ($v +$ \shiftedcosinefour) &
        \parbox[c]{2cm}{\includegraphics*[width=2cm]{images/reconstruction/celeba_64/originals/00000000.png}} & 
        \parbox[c]{2cm}{\includegraphics*[width=2cm]{images/reconstruction/celeba_64/diffae_002/0000_stochastic_0000.png}} & 
        \parbox[c]{2cm}{\includegraphics*[width=2cm]{images/reconstruction/celeba_64/diffae_002/0000_stochastic_0001.png}} &
        \parbox[c]{2cm}{\includegraphics*[width=2cm]{images/reconstruction/celeba_64/diffae_002/0000_stochastic_0002.png}} \\
        Hudson \etal\ ($\varepsilon +$ \invertedschedule) &
        \parbox[c]{2cm}{\includegraphics*[width=2cm]{images/reconstruction/celeba_64/originals/00000000.png}} & 
        \parbox[c]{2cm}{\includegraphics*[width=2cm]{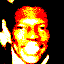}} & 
        \parbox[c]{2cm}{\includegraphics*[width=2cm]{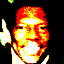}} &
        \parbox[c]{2cm}{\includegraphics*[width=2cm]{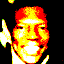}} \\
        & & & & & \\
        & \multicolumn{4}{@{\hskip 0.1cm}c@{\hskip 0.1cm}}{ImageNet 32} \\
        &
        Original &
        \multicolumn{3}{@{\hskip 0.1cm}c@{\hskip 0.1cm}}{Stochastic decodings} \\
        Preechakul \etal\ ($\varepsilon + $ \linearbeta) &
        \parbox[c]{2cm}{\includegraphics*[width=2cm]{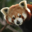}} & 
        \parbox[c]{2cm}{\includegraphics*[width=2cm]{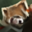}} & 
        \parbox[c]{2cm}{\includegraphics*[width=2cm]{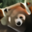}} &
        \parbox[c]{2cm}{\includegraphics*[width=2cm]{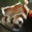}} \\
        2-phase; $\textsc{Sc-}4$ ($v +$ \shiftedcosinefour) &
        \parbox[c]{2cm}{\includegraphics*[width=2cm]{images/reconstruction/imagenet_32/originals/00000002.png}} & 
        \parbox[c]{2cm}{\includegraphics*[width=2cm]{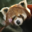}} & 
        \parbox[c]{2cm}{\includegraphics*[width=2cm]{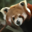}} &
        \parbox[c]{2cm}{\includegraphics*[width=2cm]{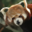}} \\
        Hudson \etal\ ($\varepsilon +$ \invertedschedule) &
        \parbox[c]{2cm}{\includegraphics*[width=2cm]{images/reconstruction/imagenet_32/originals/00000002.png}} & 
        \parbox[c]{2cm}{\includegraphics*[width=2cm]{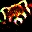}} & 
        \parbox[c]{2cm}{\includegraphics*[width=2cm]{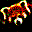}} &
        \parbox[c]{2cm}{\includegraphics*[width=2cm]{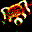}} 
    \end{tabular}
    \caption{Some reconstructions of images from the CelebA and ImageNet datasets by models trained with the baseline training method, our proposed method, and Hudson \etal's noise schedule.}
    \label{fig:hudson-reconstruction}
\end{figure*}

We evaluated Hudson \etal's training method ($\varepsilon$-prediction with the inverted noise schedule) by training a DAE for each of the 4 datasets we used. We reported metrics for quality of images reconstructed by stochastic autoencoding (no inversion) in Table~\ref{table:hudson-image-quality-metrics}. We can see that the numbers are particular bad. The PSNR values are very low, which means that the reconstructions are not accurate at all. We can see when we look at some of the reconstructed images in Figure~\ref{fig:hudson-reconstruction}. Observe that the colors of the images reconstructed by the DAEs trained with Hudson \etal's method are wrong.

\begin{table*}
    \centering
    \begin{tabular}{ll|rrrrr}
        \toprule
        \multirow{2}{*}{\textbf{Dataset}} &
        \multirow{2}{*}{\textbf{Method}} &
        \multicolumn{5}{c}{\textbf{Stochastic autoencoding}} \\
        &
        &
        \textbf{PSNR}$\uparrow$ &
        \textbf{SSIM}$\uparrow$ &
        \textbf{LPIPS}$\downarrow$ &
        \textbf{FID$\downarrow$ (tf)} &
        \textbf{FID$\downarrow$ (cf)} \\
        \midrule         
        \multirow{3}{*}{CIFAR-10 32} & Preechakul \etal\ ($\varepsilon +$ \linearbeta) & 26.887 & 0.9048 & 0.006787 & 5.2448 & 6.2506 \\
        & 2-phase; $\textsc{Sc-}4$ ($v +$ \shiftedcosinefour) & 30.351 & 0.9552 & 0.003217 & 2.1149 & 2.7153 \\
        & Hudson \etal\ ($\varepsilon +$ \invertedschedule) & \textcolor{blue}{3.834} & \textcolor{blue}{-0.1038} & \textcolor{blue}{0.235412} & \textcolor{blue}{117.0716} & \textcolor{blue}{130.0356} \\        
        \midrule        
        \multirow{3}{*}{CelebA 64} & Preechakul \etal\ ($\varepsilon +$ \linearbeta) & 23.974 & 0.8186 & 0.030239 & 11.3050 & 14.3548 \\
        & 2-phase; $\textsc{Sc-}4$ ($v +$ \shiftedcosinefour) & 26.122 & 0.8696 & 0.021513 & 2.3592 & 2.9030 \\
        & Hudson \etal\ ($\varepsilon +$ \invertedschedule) & \textcolor{blue}{10.985} & \textcolor{blue}{0.3069} & \textcolor{blue}{0.328190} & \textcolor{blue}{163.6320} & \textcolor{blue}{190.6296} \\        
        \midrule        
        \multirow{3}{*}{Bedroom 128} & Preechakul \etal\ ($\varepsilon +$ \linearbeta) & 16.120 & 0.4051 & 0.282071 & 28.8706 & 26.2747 \\
        & 2-phase; $\textsc{Sc-}4$ ($v +$ \shiftedcosinefour) & 20.309 & 0.5982 & 0.183325 & 3.2151 & 3.9847 \\
        & Hudson \etal\ ($\varepsilon +$ \invertedschedule) & \textcolor{blue}{7.709} & \textcolor{blue}{0.1615} & \textcolor{blue}{0.652557} & \textcolor{blue}{276.7475} & \textcolor{blue}{280.5181} \\        
        \midrule        
        \multirow{3}{*}{ImageNet 32} & Preechakul \etal\ ($\varepsilon +$ \linearbeta) & 21.168 & 0.7146 & 0.017876 & 12.4695 & 13.6909 \\
        & 2-phase; $\textsc{Sc-}4$ ($v +$ \shiftedcosinefour) & 27.811 & 0.9210 & 0.005147 & 4.1131 & 5.5029 \\
        & Hudson \etal\ ($\varepsilon +$ \invertedschedule) & \textcolor{blue}{12.983} & \textcolor{blue}{0.4536} & \textcolor{blue}{0.070556} & \textcolor{blue}{71.7630} & \textcolor{blue}{91.9553} \\        
        \bottomrule
    \end{tabular}
    \caption{Comparison of stochastic autoencoding performance of the baseline, our proposed method, and Hudson \etal's training method.}
    \label{table:hudson-image-quality-metrics}
\end{table*}

We were very surprised by the results because we expected that at least the colors would be correct. We checked our code for bugs but, to the best of our ability, could not find any. (We used the same code to train other DAEs, and we changed only the noise schedule to accommodate Hudson \etal's setting.) We surmise that the DAE's inability to autoencoder accurately is caused by the problems encountered by the $\varepsilon$-predictin DAE trained with $\textsc{Sc-}4$ in Section 4.2 of the main paper.

\newpage
\section{Computation of Perceptual Path Length}

The perceptual path length (PPL) metric was introduced by Karras \etal~\cite{Karras:StyleGan:2019} to measure smoothness of image changes due to interpolation of latent codes. It is defined as

\begin{align} 
\mrm{PPL} &:= \frac{1}{\varepsilon^2} E_{t \sim \mathcal{U}([0,1-\varepsilon))}\bigg[ d\Big(G\big(\mrm{Slerp}(\ve{z}_1, \ve{z}_2; t)\big), G\big(\mrm{Slerp}(\ve{z}_1, \ve{z}_2, t+\varepsilon)\big)\Big) \bigg]. \label{eq:ppl-monte-carlo}
\end{align}

Here, $\ve{z}_1$ and $\ve{z}_2$ are the latent codes being interpolated with spherical linear interpolation, which is denoted by the $\mrm{Slerp}(\cdot, \cdot; \cdot)$ function in the equation above. $G$ denotes a generator, which is a function that is capable of turning a latent code into an image. The function $d(\cdot, \cdot)$ is a distance function between two images, and Karras \etal\ implemented it with the LPIPS metrics \cite{Zhang:2018:LPIPS} computed with a VGG network \cite{Simonyan:2015}. Lastly, $\varepsilon$ is a small constant, which Karras \etal\ uses $10^{-4}$. The expectation is taken over the random variable $t$, the interpolation ``time,'' which is sampled uniformly from the interval $[0, 1-\varepsilon)$.

Equation~\eqref{eq:ppl-monte-carlo} is an approximation of the integral
\begin{align*}
\frac{1}{\varepsilon^2} \int_{t=0}^1 \Bigg[ \lim_{\Delta t \rightarrow 0} \frac{d\big(G\big(\mrm{Slerp}(\ve{z}_1, \ve{z}_2; t)\big), G\big(\mrm{Slerp}(\ve{z}_1, \ve{z}_2, t+\Delta t)\big)\big)}{\Delta t}\Bigg]\, \dee t
\end{align*}
by Monte Carlo integration. We made the calculation deterministic by dividing $[0,1]$ into $N$ subintervals, each of length $1/N$. Then, we compute the integral for each interval and add the results up. In other words,
\begin{align*}
\mrm{PPL} &= \frac{1}{\varepsilon^2} \sum_{i=0}^{N-1} \int_{t=i/N}^{(i+1)/N} \bigg[ \lim_{\Delta t \rightarrow 0} \frac{d\big(G\big(\mrm{Slerp}(\ve{z}_1, \ve{z}_2; t)\big), G\big(\mrm{Slerp}(\ve{z}_1, \ve{z}_2, t+\Delta t)\big)\big)}{\Delta t}\Bigg]\, \dee t \\
&\approx \frac{1}{\varepsilon^2} \sum_{i=0}^{N-1} \frac{d\Big(G\big(\mrm{Slerp}(\ve{z}_1, \ve{z}_2; i/N)\big), G\big(\mrm{Slerp}(\ve{z}_1, \ve{z}_2, (i+1)/N\big)\Big)}{N}.
\end{align*}
Lastly, we set $\varepsilon = 1/N$, yielding
\begin{align*}
\mrm{PPL} &:= N \sum_{i=0}^{N-1} d\Big(G\big(\mrm{Slerp}(\ve{z}_1, \ve{z}_2; i/N)\big), G\big(\mrm{Slerp}(\ve{z}_1, \ve{z}_2, (i+1)/N\big)\Big).
\end{align*}
In Section 5.2, we use $N = 100$.

\onecolumn

\section{Additional Results on Information with Latent Codes}

We show AUROC values of CelebA attributes in Table~\ref{tab:celeba-auroc} and CIFAR-10 classes in Table~\ref{tab:cifar10-auroc}

\begin{table*}[h]
\centering
\begin{tabular}{l|rrr}
\toprule
\textbf{Atttribute} & \textbf{Preechakul \etal}  & \textbf{1-phase; $\textsc{Sc-}4$} (\shiftedcosinefour) & \textbf{2-phase; $\textsc{Sc-}4$} (\shiftedcosinefour) \\
\midrule
5\_o\_Clock\_Shadow & 0.9416 (2) & 0.9419 (1) & 0.9356 (3) \\
Arched\_Eyebrows & 0.8986 (1) & 0.8948 (3) & 0.8957 (2) \\
Attractive & 0.9060 (2) & 0.9070 (1) & 0.9019 (3) \\
Bags\_Under\_Eyes & 0.8758 (2) & 0.8759 (1) & 0.8732 (3) \\
Bald & 0.9881 (2) & 0.9884 (1) & 0.9834 (3) \\
Bangs & 0.9824 (2) & 0.9829 (1) & 0.9743 (3) \\
Big\_Lips & 0.7661 (1) & 0.7634 (2) & 0.7619 (3) \\
Big\_Nose & 0.8747 (1) & 0.8739 (2) & 0.8723 (3) \\
Black\_Hair & 0.9354 (1) & 0.9320 (2) & 0.9196 (3) \\
Blond\_Hair & 0.9794 (1) & 0.9784 (2) & 0.9727 (3) \\
Blurry & 0.9659 (3) & 0.9683 (1) & 0.9660 (2) \\
Brown\_Hair & 0.8572 (1) & 0.8538 (2) & 0.8216 (3) \\
Bushy\_Eyebrows & 0.9288 (1) & 0.9272 (2) & 0.9235 (3) \\
Chubby & 0.9464 (2) & 0.9465 (1) & 0.9403 (3) \\
Double\_Chin & 0.9577 (1) & 0.9565 (2) & 0.9528 (3) \\
Eyeglasses & 0.9965 (2) & 0.9969 (1) & 0.9959 (3) \\
Goatee & 0.9745 (1) & 0.9740 (2) & 0.9707 (3) \\
Gray\_Hair & 0.9869 (1) & 0.9860 (2) & 0.9835 (3) \\
Heavy\_Makeup & 0.9762 (1) & 0.9758 (2) & 0.9734 (3) \\
High\_Cheekbones & 0.9434 (1) & 0.9433 (2) & 0.9400 (3) \\
Male & 0.9967 (1) & 0.9965 (2) & 0.9963 (3) \\
Mouth\_Slightly\_Open & 0.9816 (1) & 0.9788 (3) & 0.9797 (2) \\
Mustache & 0.9656 (2) & 0.9667 (1) & 0.9644 (3) \\
Narrow\_Eyes & 0.8795 (1) & 0.8778 (2) & 0.8766 (3) \\
No\_Beard & 0.9728 (2) & 0.9729 (1) & 0.9695 (3) \\
Oval\_Face & 0.7463 (2) & 0.7484 (1) & 0.7411 (3) \\
Pale\_Skin & 0.9703 (2) & 0.9704 (1) & 0.9670 (3) \\
Pointy\_Nose & 0.7856 (1) & 0.7843 (2) & 0.7792 (3) \\
Receding\_Hairline & 0.9350 (1) & 0.9331 (2) & 0.9262 (3) \\
Rosy\_Cheeks & 0.9624 (2) & 0.9624 (1) & 0.9576 (3) \\
Sideburns & 0.9797 (1) & 0.9797 (2) & 0.9745 (3) \\
Smiling & 0.9810 (1) & 0.9803 (3) & 0.9804 (2) \\
Straight\_Hair & 0.7966 (1) & 0.7921 (2) & 0.7643 (3) \\
Wavy\_Hair & 0.8802 (1) & 0.8782 (2) & 0.8548 (3) \\
Wearing\_Earrings & 0.9014 (1) & 0.8980 (2) & 0.8816 (3) \\
Wearing\_Hat & 0.9935 (1) & 0.9929 (2) & 0.9865 (3) \\
Wearing\_Lipstick & 0.9835 (1) & 0.9829 (2) & 0.9816 (3) \\
Wearing\_Necklace & 0.7984 (1) & 0.7971 (2) & 0.7928 (3) \\
Wearing\_Necktie & 0.9441 (1) & 0.9409 (2) & 0.9372 (3) \\
Young & 0.9150 (2) & 0.9151 (1) & 0.9107 (3) \\
\midrule
\textbf{Average} & 0.9263 (1) & 0.9254 (2) & 0.9195 (3) \\
\bottomrule
\end{tabular}
\caption{AUROC Values for CelebA attributes. The numbers in parentheses are the ranks of individual AUROC values when they are compared to corresponding numbers of other training methods.}
\label{tab:celeba-auroc}
\end{table*}

\begin{table*}[h]
\centering
\begin{tabular}{l|rrr}
\toprule
\textbf{Class} & \textbf{Preechakul \etal} & \textbf{1-phase; $\textsc{Sc-}4$} (\shiftedcosinefour) & \textbf{2-phase; $\textsc{Sc-}4$} (\shiftedcosinefour) \\
\midrule
airplane & 0.9720 (2) & 0.9701 (3) & 0.9724 (1) \\
automobile & 0.9830 (2) & 0.9813 (3) & 0.9837 (1) \\
bird & 0.9347 (1) & 0.9298 (2) & 0.9281 (3) \\
cat & 0.9192 (2) & 0.9119 (3) & 0.9205 (1) \\
deer & 0.9609 (1) & 0.9494 (3) & 0.9515 (2) \\
dog & 0.9428 (2) & 0.9360 (3) & 0.9437 (1) \\
frog & 0.9785 (2) & 0.9781 (3) & 0.9812 (1) \\
horse & 0.9714 (2) & 0.9664 (3) & 0.9722 (1) \\
ship & 0.9839 (1) & 0.9795 (3) & 0.9799 (2) \\
truck & 0.9774 (2) & 0.9766 (3) & 0.9786 (1) \\
\midrule
\textbf{Average} & 0.9624 (1) & 0.9579 (3) & 0.9612 (2) \\
\bottomrule
\end{tabular}
\caption{AUROC Values for CIFAR-10 classes. The numbers in parentheses are the ranks of individual AUROC values when they are compared to corresponding numbers of other training methods.}
\label{tab:cifar10-auroc}
\end{table*}

\input{sec/E_reconstruction}

%% file: sec/E_reconstruction.tex
\section{Reconstruction Results}

\subsection{CIFAR-10 32}

\begin{center}
    \begin{scriptsize}
        \begin{tabular}{
            @{\hskip 0.1cm}
            c@{\hskip 0.1cm}
            c@{\hskip 0.1cm}c@{\hskip 0.1cm}c@{\hskip 0.1cm}c@{\hskip 0.1cm}c@{\hskip 0.1cm}c@{\hskip 0.1cm}        
            c@{\hskip 0.1cm}c@{\hskip 0.1cm}c@{\hskip 0.1cm}c@{\hskip 0.1cm}c@{\hskip 0.1cm}c@{\hskip 0.1cm}
            }        
            &
            Original &
            Inversion &
            \multicolumn{3}{@{\hskip 0.1cm}c@{\hskip 0.1cm}}{Stochastic decodings} &
            Original &
            Inversion &
            \multicolumn{3}{@{\hskip 0.1cm}c@{\hskip 0.1cm}}{Stochastic decodings} \\
            Preechakul \etal &
            \parbox[c]{1.15cm}{\includegraphics*[width=1.15cm]{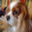}} & 
            \parbox[c]{1.15cm}{\includegraphics*[width=1.15cm]{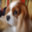}} & 
            \parbox[c]{1.15cm}{\includegraphics*[width=1.15cm]{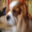}} & 
            \parbox[c]{1.15cm}{\includegraphics*[width=1.15cm]{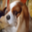}} &
            \parbox[c]{1.15cm}{\includegraphics*[width=1.15cm]{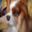}} &
            \parbox[c]{1.15cm}{\includegraphics*[width=1.15cm]{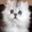}} & 
            \parbox[c]{1.15cm}{\includegraphics*[width=1.15cm]{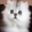}} & 
            \parbox[c]{1.15cm}{\includegraphics*[width=1.15cm]{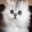}} & 
            \parbox[c]{1.15cm}{\includegraphics*[width=1.15cm]{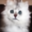}} &
            \parbox[c]{1.15cm}{\includegraphics*[width=1.15cm]{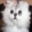}} \\
            1-phase; $\textsc{Sc-}4$ \shiftedcosinefoursmall &
            \parbox[c]{1.15cm}{\includegraphics*[width=1.15cm]{images/reconstruction/cifar_10/originals/00000000.png}} & 
            \parbox[c]{1.15cm}{\includegraphics*[width=1.15cm]{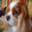}} & 
            \parbox[c]{1.15cm}{\includegraphics*[width=1.15cm]{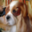}} & 
            \parbox[c]{1.15cm}{\includegraphics*[width=1.15cm]{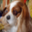}} &
            \parbox[c]{1.15cm}{\includegraphics*[width=1.15cm]{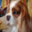}} &
            \parbox[c]{1.15cm}{\includegraphics*[width=1.15cm]{images/reconstruction/cifar_10/originals/00000001.png}} & 
            \parbox[c]{1.15cm}{\includegraphics*[width=1.15cm]{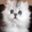}} & 
            \parbox[c]{1.15cm}{\includegraphics*[width=1.15cm]{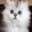}} & 
            \parbox[c]{1.15cm}{\includegraphics*[width=1.15cm]{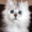}} &
            \parbox[c]{1.15cm}{\includegraphics*[width=1.15cm]{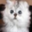}} \\
            2-phase; $\textsc{Sc-}4$ \shiftedcosinefoursmall &
            \parbox[c]{1.15cm}{\includegraphics*[width=1.15cm]{images/reconstruction/cifar_10/originals/00000000.png}} & 
            \parbox[c]{1.15cm}{\includegraphics*[width=1.15cm]{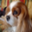}} & 
            \parbox[c]{1.15cm}{\includegraphics*[width=1.15cm]{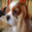}} & 
            \parbox[c]{1.15cm}{\includegraphics*[width=1.15cm]{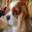}} &
            \parbox[c]{1.15cm}{\includegraphics*[width=1.15cm]{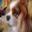}} &
            \parbox[c]{1.15cm}{\includegraphics*[width=1.15cm]{images/reconstruction/cifar_10/originals/00000001.png}} & 
            \parbox[c]{1.15cm}{\includegraphics*[width=1.15cm]{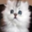}} & 
            \parbox[c]{1.15cm}{\includegraphics*[width=1.15cm]{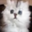}} & 
            \parbox[c]{1.15cm}{\includegraphics*[width=1.15cm]{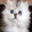}} &
            \parbox[c]{1.15cm}{\includegraphics*[width=1.15cm]{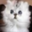}} \\
            & 
            & 
            & 
            & 
            & 
            & 
            & 
            & 
            & 
            & 
            \\ 
            Preechakul \etal &
            \parbox[c]{1.15cm}{\includegraphics*[width=1.15cm]{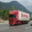}} & 
            \parbox[c]{1.15cm}{\includegraphics*[width=1.15cm]{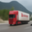}} & 
            \parbox[c]{1.15cm}{\includegraphics*[width=1.15cm]{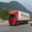}} & 
            \parbox[c]{1.15cm}{\includegraphics*[width=1.15cm]{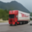}} &
            \parbox[c]{1.15cm}{\includegraphics*[width=1.15cm]{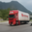}} &
            \parbox[c]{1.15cm}{\includegraphics*[width=1.15cm]{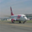}} & 
            \parbox[c]{1.15cm}{\includegraphics*[width=1.15cm]{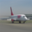}} & 
            \parbox[c]{1.15cm}{\includegraphics*[width=1.15cm]{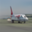}} & 
            \parbox[c]{1.15cm}{\includegraphics*[width=1.15cm]{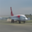}} &
            \parbox[c]{1.15cm}{\includegraphics*[width=1.15cm]{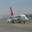}} \\
            1-phase; $\textsc{Sc-}4$ \shiftedcosinefoursmall &
            \parbox[c]{1.15cm}{\includegraphics*[width=1.15cm]{images/reconstruction/cifar_10/originals/00000002.png}} & 
            \parbox[c]{1.15cm}{\includegraphics*[width=1.15cm]{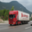}} & 
            \parbox[c]{1.15cm}{\includegraphics*[width=1.15cm]{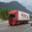}} & 
            \parbox[c]{1.15cm}{\includegraphics*[width=1.15cm]{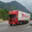}} &
            \parbox[c]{1.15cm}{\includegraphics*[width=1.15cm]{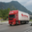}} &
            \parbox[c]{1.15cm}{\includegraphics*[width=1.15cm]{images/reconstruction/cifar_10/originals/00000003.png}} & 
            \parbox[c]{1.15cm}{\includegraphics*[width=1.15cm]{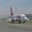}} & 
            \parbox[c]{1.15cm}{\includegraphics*[width=1.15cm]{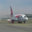}} & 
            \parbox[c]{1.15cm}{\includegraphics*[width=1.15cm]{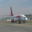}} &
            \parbox[c]{1.15cm}{\includegraphics*[width=1.15cm]{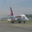}} \\
            2-phase; $\textsc{Sc-}4$ \shiftedcosinefoursmall &
            \parbox[c]{1.15cm}{\includegraphics*[width=1.15cm]{images/reconstruction/cifar_10/originals/00000002.png}} & 
            \parbox[c]{1.15cm}{\includegraphics*[width=1.15cm]{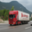}} & 
            \parbox[c]{1.15cm}{\includegraphics*[width=1.15cm]{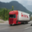}} & 
            \parbox[c]{1.15cm}{\includegraphics*[width=1.15cm]{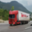}} &
            \parbox[c]{1.15cm}{\includegraphics*[width=1.15cm]{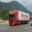}} &
            \parbox[c]{1.15cm}{\includegraphics*[width=1.15cm]{images/reconstruction/cifar_10/originals/00000003.png}} & 
            \parbox[c]{1.15cm}{\includegraphics*[width=1.15cm]{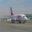}} & 
            \parbox[c]{1.15cm}{\includegraphics*[width=1.15cm]{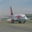}} & 
            \parbox[c]{1.15cm}{\includegraphics*[width=1.15cm]{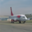}} &
            \parbox[c]{1.15cm}{\includegraphics*[width=1.15cm]{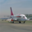}} \\
        \end{tabular}        
    \end{scriptsize}
\end{center}

CIFAR-10 is a small dataset (50K examples) of small images ($32 \times 32$), and so the 512-dimensional latent codes seem to have captured almost all of the image information, save for extremely fine details. Hence, reconstructions by all DAEs are already very close to the originals. We can observe minor differences between them only if we look carefully enough.

\subsection{CelebA 64}

\begin{center}
\begin{scriptsize}
    \begin{tabular}{
        @{\hskip 0.1cm}
        c@{\hskip 0.1cm}
        c@{\hskip 0.1cm}c@{\hskip 0.1cm}c@{\hskip 0.1cm}c@{\hskip 0.1cm}c@{\hskip 0.1cm}c@{\hskip 0.1cm}        
        c@{\hskip 0.1cm}c@{\hskip 0.1cm}c@{\hskip 0.1cm}c@{\hskip 0.1cm}c@{\hskip 0.1cm}c@{\hskip 0.1cm}
        }        
        &
        Original &
        Inversion &
        \multicolumn{3}{@{\hskip 0.1cm}c@{\hskip 0.1cm}}{Stochastic decodings} &
        Original &
        Inversion &
        \multicolumn{3}{@{\hskip 0.1cm}c@{\hskip 0.1cm}}{Stochastic decodings} \\
        Preechakul \etal &
        \parbox[c]{1.15cm}{\includegraphics*[width=1.15cm]{images/reconstruction/celeba_64/originals/00000000.png}} & 
        \parbox[c]{1.15cm}{\includegraphics*[width=1.15cm]{images/reconstruction/celeba_64/diffae_000/0000_inversion.png}} & 
        \parbox[c]{1.15cm}{\includegraphics*[width=1.15cm]{images/reconstruction/celeba_64/diffae_000/0000_stochastic_0000.png}} & 
        \parbox[c]{1.15cm}{\includegraphics*[width=1.15cm]{images/reconstruction/celeba_64/diffae_000/0000_stochastic_0001.png}} &
        \parbox[c]{1.15cm}{\includegraphics*[width=1.15cm]{images/reconstruction/celeba_64/diffae_000/0000_stochastic_0002.png}} &
        \parbox[c]{1.15cm}{\includegraphics*[width=1.15cm]{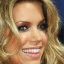}} & 
        \parbox[c]{1.15cm}{\includegraphics*[width=1.15cm]{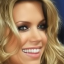}} & 
        \parbox[c]{1.15cm}{\includegraphics*[width=1.15cm]{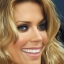}} & 
        \parbox[c]{1.15cm}{\includegraphics*[width=1.15cm]{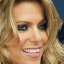}} &
        \parbox[c]{1.15cm}{\includegraphics*[width=1.15cm]{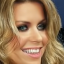}} \\
        1-phase; $\textsc{Sc-}4$ \shiftedcosinefoursmall &
        \parbox[c]{1.15cm}{\includegraphics*[width=1.15cm]{images/reconstruction/celeba_64/originals/00000000.png}} & 
        \parbox[c]{1.15cm}{\includegraphics*[width=1.15cm]{images/reconstruction/celeba_64/diffae_001/0000_inversion.png}} & 
        \parbox[c]{1.15cm}{\includegraphics*[width=1.15cm]{images/reconstruction/celeba_64/diffae_001/0000_stochastic_0000.png}} & 
        \parbox[c]{1.15cm}{\includegraphics*[width=1.15cm]{images/reconstruction/celeba_64/diffae_001/0000_stochastic_0001.png}} &
        \parbox[c]{1.15cm}{\includegraphics*[width=1.15cm]{images/reconstruction/celeba_64/diffae_001/0000_stochastic_0002.png}} &
        \parbox[c]{1.15cm}{\includegraphics*[width=1.15cm]{images/reconstruction/celeba_64/originals/00000001.png}} & 
        \parbox[c]{1.15cm}{\includegraphics*[width=1.15cm]{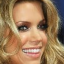}} & 
        \parbox[c]{1.15cm}{\includegraphics*[width=1.15cm]{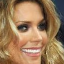}} & 
        \parbox[c]{1.15cm}{\includegraphics*[width=1.15cm]{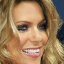}} &
        \parbox[c]{1.15cm}{\includegraphics*[width=1.15cm]{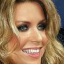}} \\
        2-phase; $\textsc{Sc-}4$ \shiftedcosinefoursmall &
        \parbox[c]{1.15cm}{\includegraphics*[width=1.15cm]{images/reconstruction/celeba_64/originals/00000000.png}} & 
        \parbox[c]{1.15cm}{\includegraphics*[width=1.15cm]{images/reconstruction/celeba_64/diffae_002/0000_inversion.png}} & 
        \parbox[c]{1.15cm}{\includegraphics*[width=1.15cm]{images/reconstruction/celeba_64/diffae_002/0000_stochastic_0000.png}} & 
        \parbox[c]{1.15cm}{\includegraphics*[width=1.15cm]{images/reconstruction/celeba_64/diffae_002/0000_stochastic_0001.png}} &
        \parbox[c]{1.15cm}{\includegraphics*[width=1.15cm]{images/reconstruction/celeba_64/diffae_002/0000_stochastic_0002.png}} &
        \parbox[c]{1.15cm}{\includegraphics*[width=1.15cm]{images/reconstruction/celeba_64/originals/00000001.png}} & 
        \parbox[c]{1.15cm}{\includegraphics*[width=1.15cm]{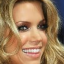}} & 
        \parbox[c]{1.15cm}{\includegraphics*[width=1.15cm]{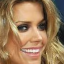}} & 
        \parbox[c]{1.15cm}{\includegraphics*[width=1.15cm]{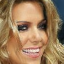}} &
        \parbox[c]{1.15cm}{\includegraphics*[width=1.15cm]{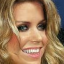}} \\
        & 
        & 
        & 
        & 
        & 
        & 
        & 
        & 
        & 
        & 
        \\ 
        Preechakul \etal &
        \parbox[c]{1.15cm}{\includegraphics*[width=1.15cm]{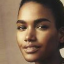}} & 
        \parbox[c]{1.15cm}{\includegraphics*[width=1.15cm]{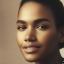}} & 
        \parbox[c]{1.15cm}{\includegraphics*[width=1.15cm]{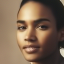}} & 
        \parbox[c]{1.15cm}{\includegraphics*[width=1.15cm]{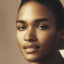}} &
        \parbox[c]{1.15cm}{\includegraphics*[width=1.15cm]{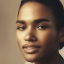}} &
        \parbox[c]{1.15cm}{\includegraphics*[width=1.15cm]{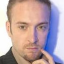}} & 
        \parbox[c]{1.15cm}{\includegraphics*[width=1.15cm]{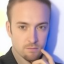}} & 
        \parbox[c]{1.15cm}{\includegraphics*[width=1.15cm]{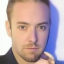}} & 
        \parbox[c]{1.15cm}{\includegraphics*[width=1.15cm]{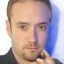}} &
        \parbox[c]{1.15cm}{\includegraphics*[width=1.15cm]{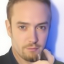}} \\
        1-phase; $\textsc{Sc-}4$ \shiftedcosinefoursmall &
        \parbox[c]{1.15cm}{\includegraphics*[width=1.15cm]{images/reconstruction/celeba_64/originals/00000002.png}} & 
        \parbox[c]{1.15cm}{\includegraphics*[width=1.15cm]{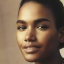}} & 
        \parbox[c]{1.15cm}{\includegraphics*[width=1.15cm]{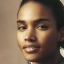}} & 
        \parbox[c]{1.15cm}{\includegraphics*[width=1.15cm]{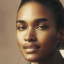}} &
        \parbox[c]{1.15cm}{\includegraphics*[width=1.15cm]{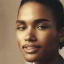}} &
        \parbox[c]{1.15cm}{\includegraphics*[width=1.15cm]{images/reconstruction/celeba_64/originals/00000003.png}} & 
        \parbox[c]{1.15cm}{\includegraphics*[width=1.15cm]{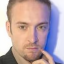}} & 
        \parbox[c]{1.15cm}{\includegraphics*[width=1.15cm]{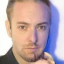}} & 
        \parbox[c]{1.15cm}{\includegraphics*[width=1.15cm]{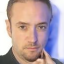}} &
        \parbox[c]{1.15cm}{\includegraphics*[width=1.15cm]{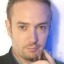}} \\
        2-phase; $\textsc{Sc-}4$ \shiftedcosinefoursmall &
        \parbox[c]{1.15cm}{\includegraphics*[width=1.15cm]{images/reconstruction/celeba_64/originals/00000002.png}} & 
        \parbox[c]{1.15cm}{\includegraphics*[width=1.15cm]{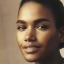}} & 
        \parbox[c]{1.15cm}{\includegraphics*[width=1.15cm]{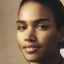}} & 
        \parbox[c]{1.15cm}{\includegraphics*[width=1.15cm]{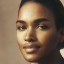}} &
        \parbox[c]{1.15cm}{\includegraphics*[width=1.15cm]{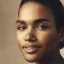}} &
        \parbox[c]{1.15cm}{\includegraphics*[width=1.15cm]{images/reconstruction/celeba_64/originals/00000003.png}} & 
        \parbox[c]{1.15cm}{\includegraphics*[width=1.15cm]{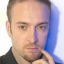}} & 
        \parbox[c]{1.15cm}{\includegraphics*[width=1.15cm]{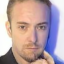}} & 
        \parbox[c]{1.15cm}{\includegraphics*[width=1.15cm]{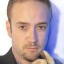}} &
        \parbox[c]{1.15cm}{\includegraphics*[width=1.15cm]{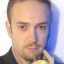}} \\
    \end{tabular}        
\end{scriptsize}
\end{center}

For the CelebA dataset, the 1-phase training method with $\textsc{Sc-}$4 (\shiftedcosinefour) and our proposed 2-phase training method produced sharper images compared to those of Preechakul \etal's method. We show examples of stochastic decodings that highlight differences in sharpness below.

\begin{center}
    \scriptsize
    \centering
    \def\imagewidth{2.5cm}
\begin{tabular}{c@{\hskip 0.1cm}c@{\hskip 0.1cm}c@{\hskip 0.1cm}c}
        
        Original &
        Preechakul \etal &
        1-phase; $\textsc{Sc-}4$ (\shiftedcosinefoursmall) &
        2-phase; $\textsc{Sc-}4$ (\shiftedcosinefoursmall) \\
\end{tabular}
\end{center}

\begin{center}
    \scriptsize
    \centering
    \def\imagewidth{2.5cm}
    \begin{tabular}{c@{\hskip 0.1cm}c@{\hskip 0.1cm}c@{\hskip 0.1cm}c}
        
        Original &
        Preechakul \etal &
        1-phase; $\textsc{Sc-}4$ (\shiftedcosinefoursmall) &
        2-phase; $\textsc{Sc-}4$ (\shiftedcosinefoursmall) \\
\end{tabular}
\end{center}

\begin{center}
    \scriptsize
    \centering
    \def\imagewidth{2.5cm}
    \begin{tabular}{c@{\hskip 0.1cm}c@{\hskip 0.1cm}c@{\hskip 0.1cm}c}
        
        Original &
        Preechakul \etal &
        1-phase; $\textsc{Sc-}4$ (\shiftedcosinefoursmall) &
        2-phase; $\textsc{Sc-}4$ (\shiftedcosinefoursmall) \\
\end{tabular}
\end{center}

\pagebreak
\subsection{LSUN Bedroom 128}

\begin{center}
\begin{scriptsize}
    \begin{tabular}{
        @{\hskip 0.1cm}
        c@{\hskip 0.1cm}
        c@{\hskip 0.1cm}c@{\hskip 0.1cm}c@{\hskip 0.1cm}c@{\hskip 0.1cm}c@{\hskip 0.1cm}c@{\hskip 0.1cm}        
        c@{\hskip 0.1cm}c@{\hskip 0.1cm}c@{\hskip 0.1cm}c@{\hskip 0.1cm}c@{\hskip 0.1cm}c@{\hskip 0.1cm}
        }        
        &
        Original &
        Inversion &
        \multicolumn{3}{@{\hskip 0.1cm}c@{\hskip 0.1cm}}{Stochastic decodings} &
        Original &
        Inversion &
        \multicolumn{3}{@{\hskip 0.1cm}c@{\hskip 0.1cm}}{Stochastic decodings} \\
        Preechakul \etal &
        \parbox[c]{1.15cm}{\includegraphics*[width=1.15cm]{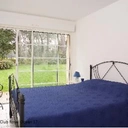}} & 
        \parbox[c]{1.15cm}{\includegraphics*[width=1.15cm]{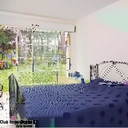}} & 
        \parbox[c]{1.15cm}{\includegraphics*[width=1.15cm]{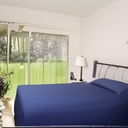}} & 
        \parbox[c]{1.15cm}{\includegraphics*[width=1.15cm]{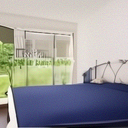}} &
        \parbox[c]{1.15cm}{\includegraphics*[width=1.15cm]{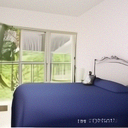}} &
        \parbox[c]{1.15cm}{\includegraphics*[width=1.15cm]{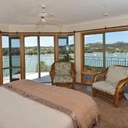}} & 
        \parbox[c]{1.15cm}{\includegraphics*[width=1.15cm]{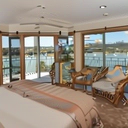}} & 
        \parbox[c]{1.15cm}{\includegraphics*[width=1.15cm]{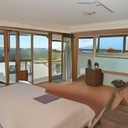}} & 
        \parbox[c]{1.15cm}{\includegraphics*[width=1.15cm]{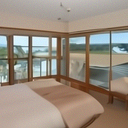}} &
        \parbox[c]{1.15cm}{\includegraphics*[width=1.15cm]{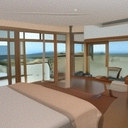}} \\
\end{tabular}        
\end{scriptsize}
\end{center}

LSUN Bedroom was the largest dataset we evaluated in terms of amount (3M images) and image size ($128 \times 128$). We found that the proposed 2-phase training method yielded the most accurate reconstructions. In the examples of stochastic decodings below, observe how our method preserved the shapes of the lamp, pillow, bed, and window while other methods sometimes distorted their shapes or made them disappear completely.

\begin{center}
    \scriptsize    
    \begin{tabular}{c@{\hskip 0.1cm}c@{\hskip 0.1cm}c@{\hskip 0.1cm}c@{\hskip 0.1cm}c}        
        \\
        \multicolumn{5}{c}{Original} \\
        \multicolumn{5}{c}{
            \begin{tikzpicture}
            \node[anchor=south west,inner sep=0] (original1) at (0,0) {
                \includegraphics*[width=2.0cm]{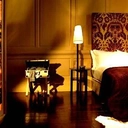} 
                };
            \begin{scope}[x={(original1.south east)},y={(original1.north west)}]
                \draw[red,thick] (.5, .4) rectangle (.7, .9);
            \end{scope}
            \end{tikzpicture}
        } \\
    \end{tabular}
    \\
    \begin{tabular}{c@{\hskip 0.1cm}c@{\hskip 0.1cm}c@{\hskip 0.1cm}c@{\hskip 0.1cm}c}
         Preechakul \etal &
        \parbox[c]{2cm}{
        \begin{tikzpicture}
            \node[anchor=south west,inner sep=0] (image) at (0,0) {
                \includegraphics*[width=2.0cm]{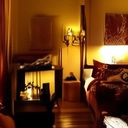} 
                };
            \begin{scope}[x={(image.south east)},y={(image.north west)}]
                \draw[red,thick] (.5, .4) rectangle (.7, .9);
            \end{scope}
            \end{tikzpicture}
        } &
        \parbox[c]{2cm}{
        \begin{tikzpicture}
            \node[anchor=south west,inner sep=0] (image) at (0,0) {
                \includegraphics*[width=2.0cm]{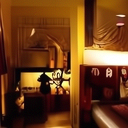} 
                };
            \begin{scope}[x={(image.south east)},y={(image.north west)}]
                \draw[red,thick] (.5, .4) rectangle (.7, .9);
            \end{scope}
            \end{tikzpicture}
        } &
        \parbox[c]{2cm}{
        \begin{tikzpicture}
            \node[anchor=south west,inner sep=0] (image) at (0,0) {
                \includegraphics*[width=2.0cm]{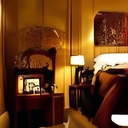} 
                };
            \begin{scope}[x={(image.south east)},y={(image.north west)}]
                \draw[red,thick] (.5, .4) rectangle (.7, .9);
            \end{scope}
            \end{tikzpicture}
        } &
        \parbox[c]{2cm}{
            \begin{tikzpicture}
            \node[anchor=south west,inner sep=0] (image) at (0,0) {
                \includegraphics*[width=2.0cm]{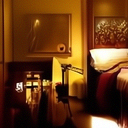} 
                };
            \begin{scope}[x={(image.south east)},y={(image.north west)}]
                \draw[red,thick] (.5, .4) rectangle (.7, .9);
            \end{scope}
            \end{tikzpicture}
        } \\
\end{tabular}
\end{center}

\pagebreak
\begin{center}
    \scriptsize
    \begin{tabular}{c@{\hskip 0.1cm}c@{\hskip 0.1cm}c@{\hskip 0.1cm}c@{\hskip 0.1cm}c}
        \\
        \multicolumn{5}{c}{Original} \\
        \multicolumn{5}{c}{
            \begin{tikzpicture}
            \node[anchor=south west,inner sep=0] (original1) at (0,0) {
                \includegraphics*[width=2.0cm]{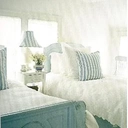} 
                };
            \begin{scope}[x={(original1.south east)},y={(original1.north west)}]
                \draw[red,thick] (.55, .3) rectangle (.83, .68);
                \draw[red,thick] (.0, .0) rectangle (0.7, 0.25);
            \end{scope}
            \end{tikzpicture}
        } \\
    \end{tabular}
    \\
    \begin{tabular}{c@{\hskip 0.1cm}c@{\hskip 0.1cm}c@{\hskip 0.1cm}c@{\hskip 0.1cm}c}
         Preechakul \etal &
        \parbox[c]{2cm}{
        \begin{tikzpicture}
            \node[anchor=south west,inner sep=0] (image) at (0,0) {
                \includegraphics*[width=2.0cm]{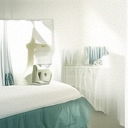} 
                };
            \begin{scope}[x={(image.south east)},y={(image.north west)}]
                \draw[red,thick] (.55, .3) rectangle (.83, .68);
                \draw[red,thick] (.0, .0) rectangle (0.7, 0.25);
            \end{scope}
            \end{tikzpicture}
        } &
        \parbox[c]{2cm}{
        \begin{tikzpicture}
            \node[anchor=south west,inner sep=0] (image) at (0,0) {
                \includegraphics*[width=2.0cm]{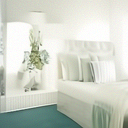} 
                };
            \begin{scope}[x={(image.south east)},y={(image.north west)}]
                \draw[red,thick] (.55, .3) rectangle (.83, .68);
                \draw[red,thick] (.0, .0) rectangle (0.7, 0.25);
            \end{scope}
            \end{tikzpicture}
        } &
        \parbox[c]{2cm}{
        \begin{tikzpicture}
            \node[anchor=south west,inner sep=0] (image) at (0,0) {
                \includegraphics*[width=2.0cm]{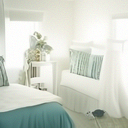} 
                };
            \begin{scope}[x={(image.south east)},y={(image.north west)}]
                \draw[red,thick] (.55, .3) rectangle (.83, .68);
                \draw[red,thick] (.0, .0) rectangle (0.7, 0.25);
            \end{scope}
            \end{tikzpicture}
        } &
        \parbox[c]{2cm}{
            \begin{tikzpicture}
            \node[anchor=south west,inner sep=0] (image) at (0,0) {
                \includegraphics*[width=2.0cm]{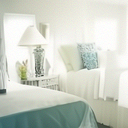} 
                };
            \begin{scope}[x={(image.south east)},y={(image.north west)}]
                \draw[red,thick] (.55, .3) rectangle (.83, .68);
                \draw[red,thick] (.0, .0) rectangle (0.7, 0.25);
            \end{scope}
            \end{tikzpicture}
        } \\
\end{tabular}
\end{center}

\begin{center}
    \begin{tabular}{c@{\hskip 0.1cm}c@{\hskip 0.1cm}c@{\hskip 0.1cm}c@{\hskip 0.1cm}c}
        \\
        \multicolumn{5}{c}{Original} \\
        \multicolumn{5}{c}{
            \begin{tikzpicture}
            \node[anchor=south west,inner sep=0] (original1) at (0,0) {
                \includegraphics*[width=2.0cm]{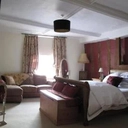} 
                };
            \begin{scope}[x={(original1.south east)},y={(original1.north west)}]
                \draw[red,thick] (.3, .025) rectangle (.7, .45);
            \end{scope}
            \end{tikzpicture}
        } \\
    \end{tabular}
    \\
    \begin{tabular}{c@{\hskip 0.1cm}c@{\hskip 0.1cm}c@{\hskip 0.1cm}c@{\hskip 0.1cm}c}
         Preechakul \etal &
        \parbox[c]{2cm}{
        \begin{tikzpicture}
            \node[anchor=south west,inner sep=0] (image) at (0,0) {
                \includegraphics*[width=2.0cm]{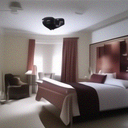} 
                };
            \begin{scope}[x={(image.south east)},y={(image.north west)}]
                \draw[red,thick] (.3, .025) rectangle (.7, .45);
            \end{scope}
            \end{tikzpicture}
        } &
        \parbox[c]{2cm}{
        \begin{tikzpicture}
            \node[anchor=south west,inner sep=0] (image) at (0,0) {
                \includegraphics*[width=2.0cm]{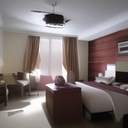} 
                };
            \begin{scope}[x={(image.south east)},y={(image.north west)}]
                \draw[red,thick] (.3, .025) rectangle (.7, .45);
            \end{scope}
            \end{tikzpicture}
        } &
        \parbox[c]{2cm}{
        \begin{tikzpicture}
            \node[anchor=south west,inner sep=0] (image) at (0,0) {
                \includegraphics*[width=2.0cm]{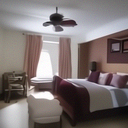} 
                };
            \begin{scope}[x={(image.south east)},y={(image.north west)}]
                \draw[red,thick] (.3, .025) rectangle (.7, .45);
            \end{scope}
            \end{tikzpicture}
        } &
        \parbox[c]{2cm}{
            \begin{tikzpicture}
            \node[anchor=south west,inner sep=0] (image) at (0,0) {
                \includegraphics*[width=2.0cm]{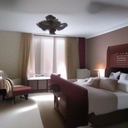} 
                };
            \begin{scope}[x={(image.south east)},y={(image.north west)}]
                \draw[red,thick] (.3, .025) rectangle (.7, .45);
            \end{scope}
            \end{tikzpicture}
        } \\
\end{tabular}
\end{center}

\pagebreak
\begin{center}
    \begin{tabular}{c@{\hskip 0.1cm}c@{\hskip 0.1cm}c@{\hskip 0.1cm}c@{\hskip 0.1cm}c}
        \\
        \multicolumn{5}{c}{Original} \\
        \multicolumn{5}{c}{
            \begin{tikzpicture}
            \node[anchor=south west,inner sep=0] (original1) at (0,0) {
                \includegraphics*[width=2.0cm]{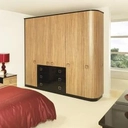} 
                };
            \begin{scope}[x={(original1.south east)},y={(original1.north west)}]
                \draw[red,thick] (.8, .2) rectangle (1., .9);
                \draw[red,thick] (.26, .23) rectangle (.55, .55);
            \end{scope}
            \end{tikzpicture}
        } \\
    \end{tabular}
    \\
    \begin{tabular}{c@{\hskip 0.1cm}c@{\hskip 0.1cm}c@{\hskip 0.1cm}c@{\hskip 0.1cm}c}
         Preechakul \etal &
        \parbox[c]{2cm}{
        \begin{tikzpicture}
            \node[anchor=south west,inner sep=0] (image) at (0,0) {
                \includegraphics*[width=2.0cm]{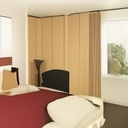} 
                };
            \begin{scope}[x={(image.south east)},y={(image.north west)}]
                \draw[red,thick] (.8, .2) rectangle (1., .9);
                \draw[red,thick] (.26, .23) rectangle (.55, .55);
            \end{scope}
            \end{tikzpicture}
        } &
        \parbox[c]{2cm}{
        \begin{tikzpicture}
            \node[anchor=south west,inner sep=0] (image) at (0,0) {
                \includegraphics*[width=2.0cm]{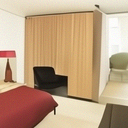} 
                };
            \begin{scope}[x={(image.south east)},y={(image.north west)}]
                \draw[red,thick] (.8, .2) rectangle (1., .9);
                \draw[red,thick] (.26, .23) rectangle (.55, .55);
            \end{scope}
            \end{tikzpicture}
        } &
        \parbox[c]{2cm}{
        \begin{tikzpicture}
            \node[anchor=south west,inner sep=0] (image) at (0,0) {
                \includegraphics*[width=2.0cm]{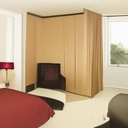} 
                };
            \begin{scope}[x={(image.south east)},y={(image.north west)}]
                \draw[red,thick] (.8, .2) rectangle (1., .9);
                \draw[red,thick] (.26, .23) rectangle (.55, .55);
            \end{scope}
            \end{tikzpicture}
        } &
        \parbox[c]{2cm}{
            \begin{tikzpicture}
            \node[anchor=south west,inner sep=0] (image) at (0,0) {
                \includegraphics*[width=2.0cm]{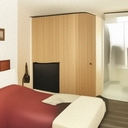} 
                };
            \begin{scope}[x={(image.south east)},y={(image.north west)}]
                \draw[red,thick] (.8, .2) rectangle (1., .9);
                \draw[red,thick] (.26, .23) rectangle (.55, .55);
            \end{scope}
            \end{tikzpicture}
        } \\
\end{tabular}
\end{center}

\subsection{ImageNet 32}

\begin{center}
\begin{scriptsize}
    \begin{tabular}{
        @{\hskip 0.1cm}
        c@{\hskip 0.1cm}
        c@{\hskip 0.1cm}c@{\hskip 0.1cm}c@{\hskip 0.1cm}c@{\hskip 0.1cm}c@{\hskip 0.1cm}c@{\hskip 0.1cm}        
        c@{\hskip 0.1cm}c@{\hskip 0.1cm}c@{\hskip 0.1cm}c@{\hskip 0.1cm}c@{\hskip 0.1cm}c@{\hskip 0.1cm}
        }        
        &
        Original &
        Inversion &
        \multicolumn{3}{@{\hskip 0.1cm}c@{\hskip 0.1cm}}{Stochastic decodings} &
        Original &
        Inversion &
        \multicolumn{3}{@{\hskip 0.1cm}c@{\hskip 0.1cm}}{Stochastic decodings} \\
        Preechakul \etal &
        \parbox[c]{1.15cm}{\includegraphics*[width=1.15cm]{images/reconstruction/imagenet_32/originals/00000000.png}} & 
        \parbox[c]{1.15cm}{\includegraphics*[width=1.15cm]{images/reconstruction/imagenet_32/diffae_000/0000_inversion.png}} & 
        \parbox[c]{1.15cm}{\includegraphics*[width=1.15cm]{images/reconstruction/imagenet_32/diffae_000/0000_stochastic_0000.png}} & 
        \parbox[c]{1.15cm}{\includegraphics*[width=1.15cm]{images/reconstruction/imagenet_32/diffae_000/0000_stochastic_0001.png}} &
        \parbox[c]{1.15cm}{\includegraphics*[width=1.15cm]{images/reconstruction/imagenet_32/diffae_000/0000_stochastic_0002.png}} &
        \parbox[c]{1.15cm}{\includegraphics*[width=1.15cm]{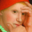}} & 
        \parbox[c]{1.15cm}{\includegraphics*[width=1.15cm]{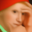}} & 
        \parbox[c]{1.15cm}{\includegraphics*[width=1.15cm]{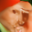}} & 
        \parbox[c]{1.15cm}{\includegraphics*[width=1.15cm]{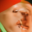}} &
        \parbox[c]{1.15cm}{\includegraphics*[width=1.15cm]{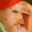}} \\
\end{tabular}        
\end{scriptsize}
\end{center}

For ImageNet 32, our proposed method yielded much more accurate stochastic decodings than other methods, and the difference in accuracy is very noticeable. In the examples below, while other methods distort the lens of the camera, the pattern on the vase, and the faces of animals, our method yielded images that are readable and free from deformation.


\pagebreak
\begin{center}
    \scriptsize    
\begin{tabular}{c@{\hskip 0.1cm}c@{\hskip 0.1cm}c@{\hskip 0.1cm}c@{\hskip 0.1cm}c}
        \\
        \multicolumn{5}{c}{Original} \\
        \multicolumn{5}{c}{
            \begin{tikzpicture}
            \node[anchor=south west,inner sep=0] (original1) at (0,0) {
                \includegraphics*[width=2.0cm]{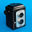} 
                };
            \begin{scope}[x={(original1.south east)},y={(original1.north west)}]
                \draw[red,thick] (.4, .1) rectangle (.8, .9);
            \end{scope}
            \end{tikzpicture}
        } \\
    \end{tabular}
    \\
    \begin{tabular}{c@{\hskip 0.1cm}c@{\hskip 0.1cm}c@{\hskip 0.1cm}c@{\hskip 0.1cm}c}
         Preechakul \etal &
        \parbox[c]{2cm}{
        \begin{tikzpicture}
            \node[anchor=south west,inner sep=0] (image) at (0,0) {
                \includegraphics*[width=2.0cm]{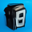} 
                };
            \begin{scope}[x={(image.south east)},y={(image.north west)}]
                \draw[red,thick] (.4, .1) rectangle (.8, .9);
            \end{scope}
            \end{tikzpicture}
        } &
        \parbox[c]{2cm}{
        \begin{tikzpicture}
            \node[anchor=south west,inner sep=0] (image) at (0,0) {
                \includegraphics*[width=2.0cm]{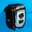} 
                };
            \begin{scope}[x={(image.south east)},y={(image.north west)}]
                \draw[red,thick] (.4, .1) rectangle (.8, .9);
            \end{scope}
            \end{tikzpicture}
        } &
        \parbox[c]{2cm}{
        \begin{tikzpicture}
            \node[anchor=south west,inner sep=0] (image) at (0,0) {
                \includegraphics*[width=2.0cm]{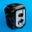} 
                };
            \begin{scope}[x={(image.south east)},y={(image.north west)}]
                \draw[red,thick] (.4, .1) rectangle (.8, .9);
            \end{scope}
            \end{tikzpicture}
        } &
        \parbox[c]{2cm}{
            \begin{tikzpicture}
            \node[anchor=south west,inner sep=0] (image) at (0,0) {
                \includegraphics*[width=2.0cm]{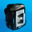} 
                };
            \begin{scope}[x={(image.south east)},y={(image.north west)}]
                \draw[red,thick] (.4, .1) rectangle (.8, .9);
            \end{scope}
            \end{tikzpicture}
        } \\
\end{tabular}
\end{center}

\begin{center}
    \begin{tabular}{c@{\hskip 0.1cm}c@{\hskip 0.1cm}c@{\hskip 0.1cm}c@{\hskip 0.1cm}c}
        \\
        \multicolumn{5}{c}{Original} \\
        \multicolumn{5}{c}{
            \begin{tikzpicture}
            \node[anchor=south west,inner sep=0] (original1) at (0,0) {
                \includegraphics*[width=2.0cm]{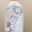} 
                };
            \begin{scope}[x={(original1.south east)},y={(original1.north west)}]
                \draw[red,thick] (.16, .0) rectangle (.76, .9);
            \end{scope}
            \end{tikzpicture}
        } \\
    \end{tabular}
    \\
    \begin{tabular}{c@{\hskip 0.1cm}c@{\hskip 0.1cm}c@{\hskip 0.1cm}c@{\hskip 0.1cm}c}
         Preechakul \etal &
        \parbox[c]{2cm}{
        \begin{tikzpicture}
            \node[anchor=south west,inner sep=0] (image) at (0,0) {
                \includegraphics*[width=2.0cm]{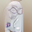} 
                };
            \begin{scope}[x={(image.south east)},y={(image.north west)}]
                \draw[red,thick] (.16, .0) rectangle (.76, .9);
            \end{scope}
            \end{tikzpicture}
        } &
        \parbox[c]{2cm}{
        \begin{tikzpicture}
            \node[anchor=south west,inner sep=0] (image) at (0,0) {
                \includegraphics*[width=2.0cm]{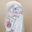} 
                };
            \begin{scope}[x={(image.south east)},y={(image.north west)}]
                \draw[red,thick] (.16, .0) rectangle (.76, .9);
            \end{scope}
            \end{tikzpicture}
        } &
        \parbox[c]{2cm}{
        \begin{tikzpicture}
            \node[anchor=south west,inner sep=0] (image) at (0,0) {
                \includegraphics*[width=2.0cm]{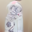} 
                };
            \begin{scope}[x={(image.south east)},y={(image.north west)}]
                \draw[red,thick] (.16, .0) rectangle (.76, .9);
            \end{scope}
            \end{tikzpicture}
        } &
        \parbox[c]{2cm}{
            \begin{tikzpicture}
            \node[anchor=south west,inner sep=0] (image) at (0,0) {
                \includegraphics*[width=2.0cm]{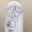} 
                };
            \begin{scope}[x={(image.south east)},y={(image.north west)}]
                \draw[red,thick] (.16, .0) rectangle (.76, .9);
            \end{scope}
            \end{tikzpicture}
        } \\
\end{tabular}
\end{center}

\pagebreak    
\begin{center}
    \begin{tabular}{c@{\hskip 0.1cm}c@{\hskip 0.1cm}c@{\hskip 0.1cm}c@{\hskip 0.1cm}c}
        \\
        \multicolumn{5}{c}{Original} \\
        \multicolumn{5}{c}{
            \begin{tikzpicture}
            \node[anchor=south west,inner sep=0] (original1) at (0,0) {
                \includegraphics*[width=2.0cm]{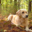} 
                };
            \begin{scope}[x={(original1.south east)},y={(original1.north west)}]
                \draw[red,thick] (.5, .35) rectangle (.93, .78);
            \end{scope}
            \end{tikzpicture}
        } \\
    \end{tabular}
    \\
    \begin{tabular}{c@{\hskip 0.1cm}c@{\hskip 0.1cm}c@{\hskip 0.1cm}c@{\hskip 0.1cm}c}
         Preechakul \etal &
        \parbox[c]{2cm}{
        \begin{tikzpicture}
            \node[anchor=south west,inner sep=0] (image) at (0,0) {
                \includegraphics*[width=2.0cm]{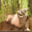} 
                };
            \begin{scope}[x={(image.south east)},y={(image.north west)}]
                \draw[red,thick] (.5, .35) rectangle (.93, .78);
            \end{scope}
            \end{tikzpicture}
        } &
        \parbox[c]{2cm}{
        \begin{tikzpicture}
            \node[anchor=south west,inner sep=0] (image) at (0,0) {
                \includegraphics*[width=2.0cm]{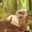} 
                };
            \begin{scope}[x={(image.south east)},y={(image.north west)}]
                \draw[red,thick] (.5, .35) rectangle (.93, .78);
            \end{scope}
            \end{tikzpicture}
        } &
        \parbox[c]{2cm}{
        \begin{tikzpicture}
            \node[anchor=south west,inner sep=0] (image) at (0,0) {
                \includegraphics*[width=2.0cm]{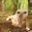} 
                };
            \begin{scope}[x={(image.south east)},y={(image.north west)}]
                \draw[red,thick] (.5, .35) rectangle (.93, .78);
            \end{scope}
            \end{tikzpicture}
        } &
        \parbox[c]{2cm}{
            \begin{tikzpicture}
            \node[anchor=south west,inner sep=0] (image) at (0,0) {
                \includegraphics*[width=2.0cm]{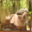} 
                };
            \begin{scope}[x={(image.south east)},y={(image.north west)}]
                \draw[red,thick] (.5, .35) rectangle (.93, .78);
            \end{scope}
            \end{tikzpicture}
        } \\
\end{tabular}
\end{center}

\begin{center}
    \begin{tabular}{c@{\hskip 0.1cm}c@{\hskip 0.1cm}c@{\hskip 0.1cm}c@{\hskip 0.1cm}c}
        \\
        \multicolumn{5}{c}{Original} \\
        \multicolumn{5}{c}{
            \begin{tikzpicture}
            \node[anchor=south west,inner sep=0] (original1) at (0,0) {
                \includegraphics*[width=2.0cm]{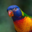} 
                };
            \begin{scope}[x={(original1.south east)},y={(original1.north west)}]
                \draw[red,thick] (.16, .3) rectangle (.8, .92);
            \end{scope}
            \end{tikzpicture}
        } \\
    \end{tabular}
    \\
    \begin{tabular}{c@{\hskip 0.1cm}c@{\hskip 0.1cm}c@{\hskip 0.1cm}c@{\hskip 0.1cm}c}
         Preechakul \etal &
        \parbox[c]{2cm}{
        \begin{tikzpicture}
            \node[anchor=south west,inner sep=0] (image) at (0,0) {
                \includegraphics*[width=2.0cm]{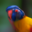} 
                };
            \begin{scope}[x={(image.south east)},y={(image.north west)}]
                \draw[red,thick] (.16, .3) rectangle (.8, .92);
            \end{scope}
            \end{tikzpicture}
        } &
        \parbox[c]{2cm}{
        \begin{tikzpicture}
            \node[anchor=south west,inner sep=0] (image) at (0,0) {
                \includegraphics*[width=2.0cm]{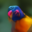} 
                };
            \begin{scope}[x={(image.south east)},y={(image.north west)}]
                \draw[red,thick] (.16, .3) rectangle (.8, .92);
            \end{scope}
            \end{tikzpicture}
        } &
        \parbox[c]{2cm}{
        \begin{tikzpicture}
            \node[anchor=south west,inner sep=0] (image) at (0,0) {
                \includegraphics*[width=2.0cm]{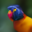} 
                };
            \begin{scope}[x={(image.south east)},y={(image.north west)}]
                \draw[red,thick] (.16, .3) rectangle (.8, .92);
            \end{scope}
            \end{tikzpicture}
        } &
        \parbox[c]{2cm}{
            \begin{tikzpicture}
            \node[anchor=south west,inner sep=0] (image) at (0,0) {
                \includegraphics*[width=2.0cm]{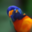} 
                };
            \begin{scope}[x={(image.south east)},y={(image.north west)}]
                \draw[red,thick] (.16, .3) rectangle (.8, .92);
            \end{scope}
            \end{tikzpicture}
        } \\
\end{tabular}
\end{center}